\documentclass[10pt,journal,compsoc]{IEEEtran}

\usepackage{lipsum,booktabs,times,color,graphicx,epstopdf,amssymb,amstext,amsmath,ragged2e}
\usepackage{array,latexsym,bm,diagbox,multicol,multirow,cite}
\DeclareGraphicsExtensions{.pdf,.jpeg,.png,.jpg,.eps}
\usepackage[caption=false,font=footnotesize]{subfig}
\usepackage[T1]{fontenc}% optional T1 font encoding
\usepackage[caption=false]{subfig}
\usepackage{caption,setspace,textcomp,algorithm,algpseudocode,soul}
\usepackage{url}
\soulregister \cite7 % 针对 \cite命令
\usepackage{gensymb}
\usepackage{siunitx}
\newtheorem{theorem}{Theorem}%  meant for continuous numbers
\newtheorem{proposition}[theorem]{Proposition}% 
\newtheorem{definition}{Definition}%

\usepackage{xcolor}         % colors
\usepackage{graphicx}       % figure
\usepackage{wrapfig}        % figure
\usepackage[caption=false,font=footnotesize]{subfig}
\begin{document}
\title{Dynamic Deep Factor Graph for Multi-Agent Reinforcement Learning
}

\author{Yuchen Shi, Shihong~Duan, Cheng~Xu,~\IEEEmembership{Member,~IEEE,} Ran Wang,~\IEEEmembership{Graduate Student Member,~IEEE,} Fangwen~Ye,  Chau~Yuen,~\IEEEmembership{Fellow,~IEEE}
        % <-this % stops a space
\thanks{
  This work was supported in part by the National Natural Science Foundation of China (NSFC) under Grant 62101029, Guangdong Basic and Applied Basic Research Foundation under Grant 2023A1515140071, and in part by the China Scholarship Council Award under Grant 202006465043 and 202306460078. (\textit{Corresponding author:} Cheng Xu)
  }
\IEEEcompsocitemizethanks
{\IEEEcompsocthanksitem Yuchen Shi, Shihong Duan and Fangwen Ye are with School of Computer and Communication Engineering, Shunde Innovation School, University of Science and Technology Beijing (email: shiyuchen199@sina.com; duansh@ustb.edu.cn; yfwen2000@outlook.com).

\IEEEcompsocthanksitem Cheng Xu is with School of Computer and Communication Engineering, Shunde Innovation School, University of Science and Technology Beijing. He is also with School of Electrical and Electronic Engineering, Nanyang Technological University (email: xucheng@ustb.edu.cn).

\IEEEcompsocthanksitem Ran Wang is with School of Computer and Communication Engineering, Shunde Innovation School, University of Science and Technology Beijing. She is also with School of Computer Science and Engineering, Nanyang Technological University (email: wangran423@foxmail.com).

\IEEEcompsocthanksitem Chau Yuen is with School of Electrical and Electronic Engineering, Nanyang Technological University. (email: chau.yuen@ntu.edu.sg).
}% <-this % stops an unwanted space
\thanks{Digital Object Identifier 10.1109/TPAMI.2024.XXXXXXX}
}

\markboth{Journal of \LaTeX\ Class Files,~Vol.~X, No.~X, Month~Year}%
{Shell \MakeLowercase{\textit{et al.}}: Bare Demo of IEEEtran.cls for Journals}

\IEEEpubid{\begin{minipage}{\textwidth}\ \\[30pt] \centering 0162-8828~\copyright~2024 IEEE. Personal use is permitted, but republication/redistribution requires IEEE permission.\\See https://www.ieee.org/publications/rights/index.html for more information.\end{minipage}}

\IEEEtitleabstractindextext{
\begin{abstract}
\justifying
Multi-agent reinforcement learning (MARL) necessitates effective collaboration among agents to accomplish designated tasks. Traditional global value function-based methodologies are encumbered by a curse of dimensionality, exacerbating the computational complexity involved. Classical algorithms for value function decomposition address this challenge by adopting a centralized training with decentralized execution (CTDE) paradigm. However, such fully decomposed approaches often lead to diminished algorithmic performance, succumbing to a phenomenon known as relative overgeneralization. While coordination graphs present a potential solution to value decomposition challenges, they lack the capability to dynamically encapsulate the evolving collaborative relationships between agents, leading to failures in certain task environments.
In response to these challenges, this work introduces a novel value decomposition algorithm, termed \textit{Dynamic Deep Factor Graphs} (DDFG). Unlike traditional coordination graphs, DDFG leverages factor graphs to articulate the decomposition of value functions, offering enhanced flexibility and adaptability to complex value function structures. Central to DDFG is a graph structure generation policy that innovatively generates factor graph structures on-the-fly, effectively addressing the dynamic collaboration requirements among agents. DDFG strikes an optimal balance between the computational overhead associated with aggregating value functions and the performance degradation inherent in their complete decomposition. Through the application of the max-sum algorithm, DDFG efficiently identifies optimal policies. We empirically validate DDFG's efficacy in complex scenarios, including higher-order predator-prey tasks and the StarCraft II Multi-agent Challenge (SMAC), thus underscoring its capability to surmount the limitations faced by existing value decomposition algorithms. DDFG emerges as a robust solution for MARL challenges that demand nuanced understanding and facilitation of dynamic agent collaboration. The implementation of DDFG is made publicly accessible, with the source code available at \url{https://github.com/SICC-Group/DDFG}.
\end{abstract}

\begin{IEEEkeywords}
\justifying
dynamic graph; factor graph; multi-agent reinforcement learning; relative overgeneralization; dynamic collaboration.
\end{IEEEkeywords}
}

\maketitle

\section{Introduction}
Collaborative multi-agent systems have become increasingly relevant across a broad spectrum of real-world scenarios, encompassing areas such as autonomous vehicular navigation \cite{cao2023continuous}, collective robotic decision-making \cite{wang2024}, and distributed sensor networks. In this context, Reinforcement Learning (RL) has demonstrated remarkable efficacy in tackling a diverse range of collaborative multi-agent challenges. Notably, intricate tasks such as the coordination of robotic swarms and the automation of vehicular control are often conceptualized within the framework of Collaborative Multi-Agent Reinforcement Learning (MARL) \cite{oroojlooy2022review}.

Addressing these complex tasks frequently necessitates the disaggregation of either the agents' policy mechanisms or their value functions. Within the domain of policy-based methodologies, Multi-Agent Deep Deterministic Policy Gradient (MADDPG) \cite{lowe2017} has been recognized for its ability to learn distributed policies tailored to continuous action spaces. Concurrently, Counterfactual Multi-Agent (COMA) \cite{foerster2018} effectively tackles the issue of credit assignment by employing counterfactual baselines. On the spectrum of value function-based approaches, Value Decomposition Networks (VDN) \cite{vdn} pioneer the decomposition of the collective action-value function into an aggregation of individual action-value functions. Furthering this paradigm, QMIX \cite{qmix} innovates by conceptualizing the collective action-value function through a monotonic function, thereby enhancing the representational capacity of the system.

However, in the realm of collaborative multi-agent systems, methodologies often grapple with a game-theoretic complication known as \textit{relative overgeneralization}, a phenomenon where the punitive consequences for non-collaboration amongst agents overshadow the rewards for cooperative engagement, culminating in less-than-optimal performance outcomes \cite{panait2006}. To mitigate this issue, QTRAN \cite{qtran} ameliorates the constraints on value function decomposition inherent in QMIX by introducing the Individual-Global-Max (IGM) condition and devising two soft regularization methods aimed at fine-tuning the action selection process to balance between joint and individual value functions. Subsequently, QPLEX \cite{qplex} advances this concept by proposing an enhanced IGM that parallels the original in importance and incorporates the Dueling architecture to further refine the decomposition of the joint value function. Meanwhile, Weighted QMIX (WQMIX) \cite{wqmix} transitions from "monotonic" to "non-monotonic" value functions through the introduction of a weighted operator, addressing the limitations of monotonicity constraints. Despite these advancements, the stated algorithms occasionally falter in identifying the globally optimal solution due to the inherent constraints of value decomposition.

The Coordination Graph (CG) framework \cite{guestrin2002} presents a viable alternative that preserves the benefits of value function decomposition while simultaneously addressing the issue of relative overgeneralization. Within this framework, agents and the synergistic gains from their interactions are represented as vertices and edges, respectively, in a graph structured according to the joint action-observation space. Deep Coordination Graphs (DCG) \cite{dcg} approximate these gain functions using deep neural networks and disseminate the derived value functions across the graph via a message-passing algorithm. The Deep Implicit Coordination Graph (DICG) \cite{li2020} extrapolates upon the CG model by integrating attention mechanisms through a Graph Convolutional Network (GCN), thereby proposing an implicit framework for coordination graphs. Despite their capability to disaggregate the value function, coordination graphs fall short in adequately attributing credit for complex tasks. Additionally, the fixed structure of coordination graphs limits their applicability to diverse real-world scenarios. Attempts to extend CGs to nonlinear scenarios (NL-CG) \cite{kang2022} and to enhance representational capacities through sparse dynamic graph structures (SOPCG \cite{sopcg} and CASEC \cite{casec}) have been made, yet these approaches remain constricted by the inherent limitations of the coordination graph structure, especially when addressing complex coordination relationships.

To surmount these limitations, this paper introduces a novel value decomposition algorithm termed the \textit{Dynamic Deep Factor Graph} (DDFG), which leverages factor graphs \cite{loeliger2004} for the decomposition of the global value function into a sum of local value functions of arbitrary order, thereby offering a more robust characterization capability than coordination graphs. Distinctly, DDFG eschews predefined graph structures in favor of dynamically generated graphs that accurately represent the collaborative relationships among agents, based on real-time observations processed through neural networks. This paper defines the graph generation policy as a quasi-multinomial distribution and provides relevant theoretical proofs. Following the idea of probability modeling from the PPO\cite{schulman2017}, we constructed a networked graph policy over a quasi-multinomial distribution to generate dynamic factor graph structures in various scenarios. The adaptability of the graph structure is further enhanced by the application of the message-passing algorithm \cite{kschischang2001}, optimizing the learning of agents' policies and the representational efficiency of value function decomposition.

The core contributions of our study are delineated as follows:
\begin{enumerate}
\item \textit{Canonical Polyadic (CP) Decomposition-based Factor Graph Value Function Network.} We introduce a novel representation utilizing factor graphs for the decomposition of global value functions into a sum of higher-order local value functions, augmented by tensor-based CP decomposition \cite{phan2021,zhou2019}. This approach, an extension of the matrix low-rank approximation, significantly reduces parameter count and increases update frequency by decomposing higher-order value functions into the outer products of rank-one tensors.

\item \textit{Graph Structure Generation Policy.} We propose a new graph structure generation policy, defined as a quasi-multinomial distribution. It is analogous to agent policies, for generating real-time, variable factor graph structures through probabilistic modeling, akin to the Proximal Policy Optimization (PPO) algorithm. This policy facilitates the depiction of agent collaboration, employing a neural network that processes agents' observations to generate dynamic graph structures.

\item \textit{Dynamic Deep Factor Graph-Based MARL Algorithm.} The DDFG algorithm, leveraging factor graphs over coordination graphs, addresses the pitfall of relative overgeneralization by providing a more versatile decomposition approach and generating dynamic factor graph structures to enhance expressive power. Through the application of factor graph message passing algorithms to various graph structures, DDFG facilitates the learning of agents' policies. Comparative analysis with contemporary MARL methodologies within the experimental section showcases DDFG's efficacy in complex scenarios such as high-order predator-prey and SMAC tasks.
\end{enumerate}

The rest of this paper is organized as follows. Section 2 introduces the definition of factor graphs and Markov decision processes based on factor graphs. Section 3 presents the proposed Dynamic Deep Factor Graph algorithm. Section 4 presents the composition of algorithm related loss functions. The simulation results and corresponding analysis are given in Section 5. In Section 6, we conclude this paper.

\section{Background}

\subsection{FG-POMDP}
A factor graph $G$ is defined by a set of variable nodes $V$, a set of factor nodes $F$, and a set of undirected edges $varepsilon$. We correspond intelligent agents with variable nodes; If the local value function corresponds to a factor node, then a factor graph can represent a decomposition form of the global value function. Define the adjacency matrix of the factor graph at each time t as ${A}_ {t}$ and add it to POMDP, proposing a factor graph based POMDP, namely FG-POMDP. FG-POMDP consists of a tuple $<n,S,A,\mathcal{U},R,P,\{{{O}^{i}}\}_{i=1}^{n},\gamma >$ \cite{oliehoek2016}. Here, $n$ represents the number of agents, $S$ describes the true state of the environment, and ${{\mathcal{U}}^{i}}\in \mathcal{U}$ denotes the set of discrete actions available to agent $i$. At discrete time t, ${{A}_{t}}\in A$ is the dynamic factor graph structure in real-time. We denote it by the adjacency matrix ${{A}_{t}}\in {{\left\{ 0,1 \right\}}^{n\times m}}$, where $m$ denotes the number of factor nodes. The next state ${{s}_{t+1}}\in S$ is obtained from the transfer probability ${{s}_{t+1}}\sim P(\cdot \mid{{s}_{t}},{\bm{u}_{t}})$ with ${{s}_{t+1}}\in S$ and ${\bm{u}_{t}}\in \mathcal{U}:={{\mathcal{U}}^{1}}\times \ldots \times {{\mathcal{U}}^{n}}$ as conditions. Each agent shares the same reward ${{r}_{t}}:=r({{s}_{t}},{\bm{u}_{t}})$ at moment $t$ and $\gamma \in [0,1)$ denotes the discount factor. Due to partial observability, at moment t, the individual observations $o_{t}^{i}\in {{O}^{i}}$ of each agent, the history of observations $o_{t}^{i}$ and actions $u_{t}^{i}$ of agent $i $ are represented as $\tau _{t}^{i}:=\left( o_{0}^{i},u_{0}^{i},o_{ 1}^{i},\ldots,o_{t-1}^{i},u_{t-1}^{i},o_{t}^{i} \right)\in {{\left( {{O}^{i}}\times {{\mathcal{U}}^{i}} \right)}^{t}}\times {{O}^{i}}$. Without loss of generality, the trajectory of the task in this paper is denoted as $\mathcal{T}=\left( {{s}_{0}},\{o_{0}^{i}\}_{i=1}^{n},{{A}_{0}},\bm{u}_{0},{{r}_{0}},\ldots ,{{s}_{T}},\{o_{T}^{i} \}_{i=1}^{n} \right)$, where $T$ is the task length.

Collaborative MARL aims to find the optimal policy ${{\pi }^{}}: S\times \mathcal{U}\to [0,1]$ that selects joint actions ${\bm{u}_{t}}\in \mathcal{U}$ maximizing the expected discounted sum of future rewards, achieved by estimating the optimal Q-value function. The optimal policy ${{\pi }^{}}$ greedily selects actions $\bm{u}\in \mathcal{U}$ that maximize the corresponding optimal Q-value function. However, when facing large joint action spaces, deep neural networks with $\theta$ parameters, such as DQN \cite{mnih2015} and Double-DQN \cite{van2016}, may struggle to approximate the optimal Q-value function ${{Q}_{\theta }}$. To address this issue, various value decomposition algorithms have been proposed to perform Q-learning in MARL efficiently. We describe these value decomposition algorithms(VDN \cite{vdn}, QMIX \cite{qmix}, WQMIX \cite{wqmix}, QTRAN \cite{qtran}, QPLEX \cite{qplex}, DCG \cite{dcg}, SOPCG \cite{sopcg}, CASEC \cite{casec}) in appendix A.

The learned policy cannot depend on the state $s_t$ in a partially observable environment. Therefore, the Q-value function $Q_{\theta}$ is conditioned on the agent's observation-action history ${\bm{\tau }_{t}}:=\{\tau _{t}^{i}\}_{i=1}^{n}$, which can be approximated as $Q_{\theta}:=Q_{\theta}(\bm{u}\mid{\bm{\tau}_{t}})$ \cite{hausknecht2015}. To achieve this, the agent's observation ${\bm{o}_t}:=\left(o_t^1, \ldots, o_t^n \right)$ and previous action ${\bm{u}_{t-1}}$ are fed into an RNN network, such as a GRU, to obtain the hidden state ${\bm{h}_{t}}:={h}_{\psi}(\cdot\mid{\bm{h}_{t-1}},{\bm{o}_{t}},{\bm{u}_{t-1}})$, where ${\bm{h}_{0}}=\bm{0}$. The Q-value function $Q_{\theta}$ is then conditioned on the hidden state ${\bm{h}_t}$, i.e., $Q_{\theta}:=Q_{\theta}(\bm{u}\mid{\bm{h}_t})$.

% \begin{wrapfigure}{r}{6cm}
%   \centering
%   \includegraphics[scale=0.3]{figures/figure1.pdf}
%   \caption{Visualization of the factor graph $Q(\bm{\tau} ,\bm{u})=\sum\limits_{j\in \mathcal{J}}{{{Q}_{j}}({\bm{u}^{j}}\mid\bm{\tau} )}$.}
%   \label{fig:1}
% \end{wrapfigure}

\begin{figure}
  \setlength{\abovecaptionskip}{0cm}
  \setlength{\belowcaptionskip}{-0.5cm}
  \centering
  \includegraphics[scale=0.35]{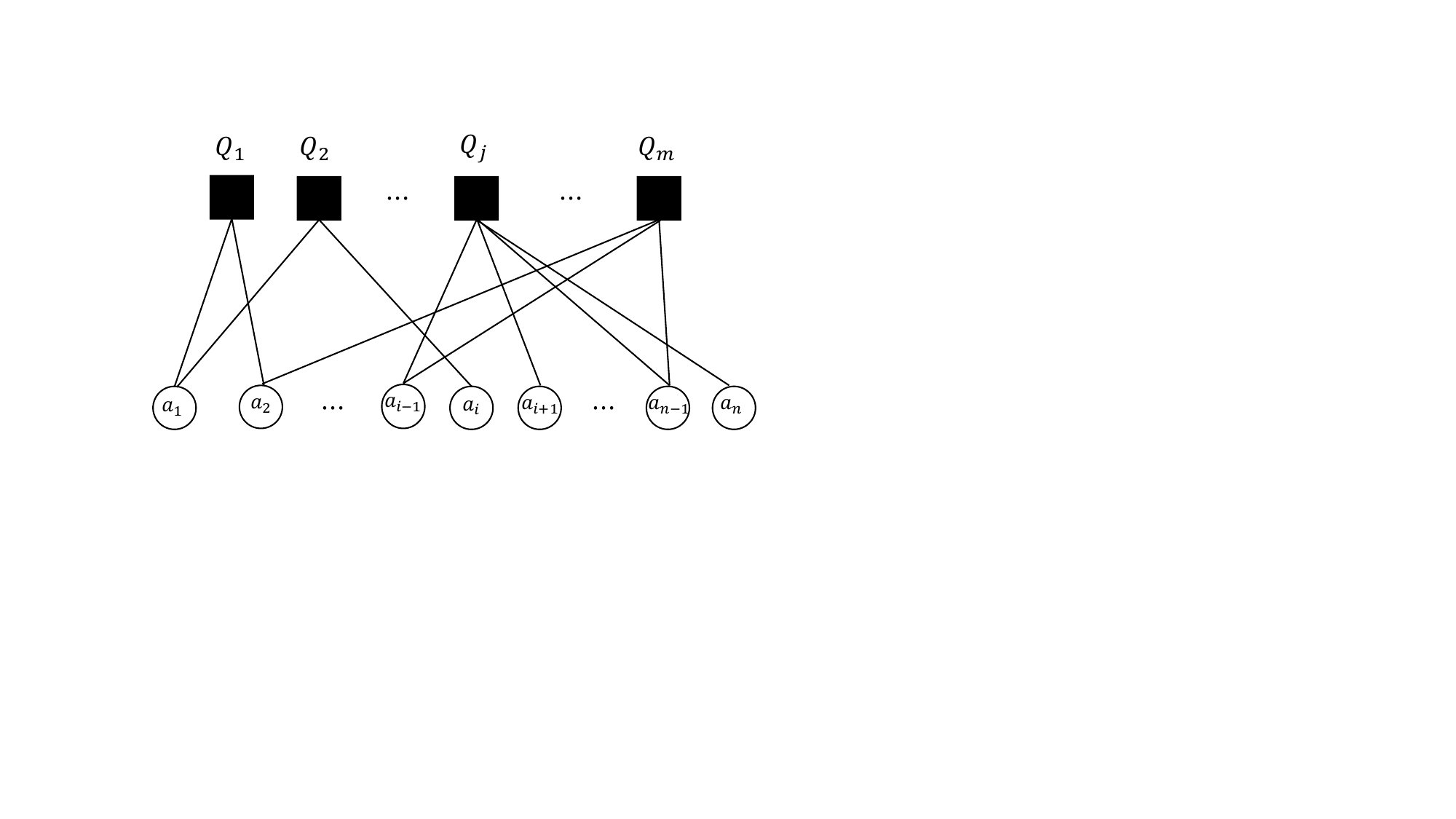}
  \caption{Visualization of the factor graph $Q(\bm{\tau} ,\bm{u})=\sum\limits_{j\in \mathcal{J}}{{{Q}_{j}}({\bm{u}^{j}}\mid\bm{\tau} )}$.}
  \label{fig:1}
\end{figure}

\subsection{Factor Graph}
A factor graph $G=\left\langle V, F,\mathcal{E} \right\rangle $ \cite{loeliger2004} is defined by the set of variable nodes $V$, the set of factor nodes $F$ and the set of undirected edges $\mathcal{E}$. Factor graphs are bipartite graphs represented as factorizations of global functions. Each variable node ${{v}_{i}}\in V$ in the factor graph corresponds to a variable. Similarly, each factor node ${{f}_{j}}\in F$ corresponds to a local function after the global function decomposition and is connected by an edge $\mathcal{E}$ to the variable node ${{v}_{i}}$ and the factor node ${{f}_{j}}$ when and only when ${{v}_{i}}$ is an argument of ${{f}_{j}}$. A factor graph with $n$ variable nodes and $m$ function nodes has a binary adjacency matrix defined as $A\in {{\left\{ 0,1 \right\}}^{n\times m}}$.

In multi-agent reinforcement learning (MARL), agents are regarded as variable nodes, and the value functions of the agents serve as factor nodes. The factor graph $G$ decomposes the global value function $Q(\bm{\tau},\bm{u})$ into a sum of several local value functions\cite{zhang2014}, and the decomposition relations are given by the adjacency matrix $A$. Each variable node ${{v}_{i}}\in V$ represents an agent $i$. Each factor node ${{f}_{j}}\in F$ is equivalent to a local value function ${{Q}_{j}}$, where $n({{Q}_{j}})$ denotes the set of all variable nodes connected to ${{Q}_{j}}$, i.e., ${i,j}\in \mathcal{E}$. ${{Q}_{j}}$ denotes the joint value function of all agents $i\in n({{Q}_{j}})$. The factor graph $G$ (as shown in Figure \ref{fig:1}) represents the joint value function.
\begin{equation}\label{eq:1}
\begin{aligned}
    & Q(\bm{\tau} ,\bm{u})=\sum\limits_{j\in \mathcal{J}}{{{f}_{j}}({{\{{{v}_{i}}\}}_{\{i,j\}\in \mathcal{E}}})} \\
    & =\sum\limits_{j\in \mathcal{J}}{{{f}_{j}}({\bm{u}^{j}}\mid\bm{\tau} )}=\sum\limits_{j\in \mathcal{J}}{{{Q}_{j}}({\bm{u}^{j}}\mid\bm{\tau} )} \\
\end{aligned}
\end{equation}
where $\mathcal{J}$ is the set of factor nodes and ${\bm{u}^{j}}\in \sum\limits_{i\in n({{Q}^{j}})}{{{\mathcal{U}}^{i}}}$ denotes the joint action of all agents.

\section{Method}
\begin{figure*}
  \setlength{\abovecaptionskip}{0cm}
  \setlength{\belowcaptionskip}{-0.5cm}
  \centering
  \includegraphics[scale=0.45]{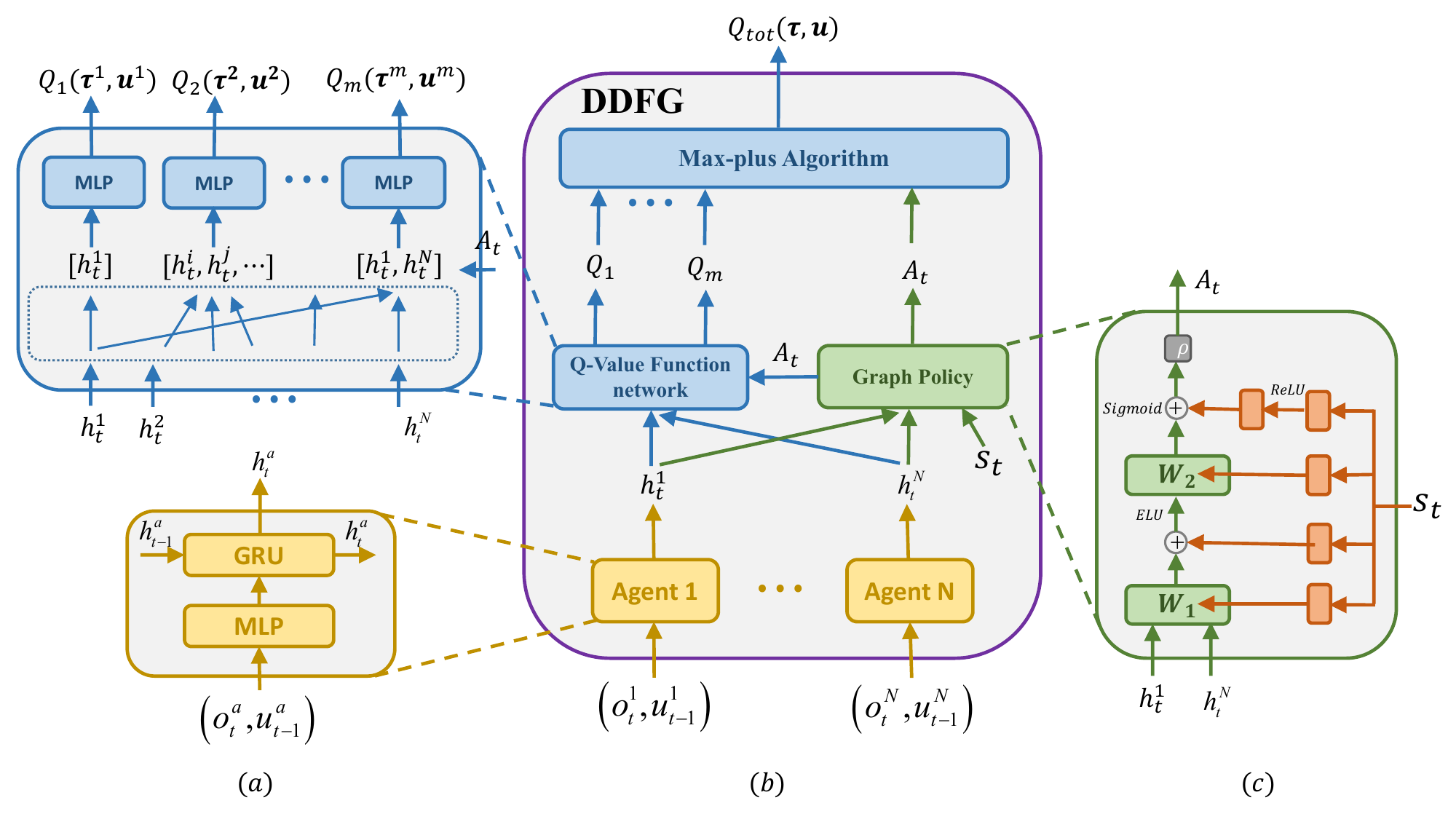}
  \caption{The algorithmic framework of DDFG.
(a) The network structure of Q-value function (Section 3.1). (b) The overall architecture of DDFG (Section 3). (c) The network structure of graph structure generation policy (Section 3.2).}
  \label{fig:2}
\end{figure*}

As depicted in Figure \ref{fig:2}, the network structure of DDFG comprises three main components: \textit{the graph policy}, responsible for generating the graph structure, \textit{the Q-value function network}, and \textit{the max-plus algorithm}. Firstly, we obtain the agent's hidden state $h_{t}^{i}$ through a shared RNN network. Both the graph policy and the Q-value function network share $h_{t}^{i}$, but the RNN network only updates jointly with the Q-value function network. The graph policy takes $h_{t}^{i}$ as input and outputs a real-time factor graph structure ${{A}_{t}}$ based on the global state ${{s}_{t}}$. Meanwhile, the Q-value function network receives $h_{t}^{i}$ and ${{A}_{t}}$ as inputs and generates the local value function ${{Q}_{j}}$. Finally, the max-plus algorithm is utilized to compute the local value function ${{Q}_{j}}$ and obtain the action and global value functions of the agent. 

\subsection{Q-value function network}
Given a known policy for generating graph structures $\rho$ (explained in Section 3.2), at time t, we use ${{A}_{t}}\sim \rho$ to represent the higher-order decomposition of the global value function $Q({\bm{\tau }_{t}},{\bm{u}_{t}})$. Simultaneously, using the adjacency matrix ${{A}_{t}}$, we represent the local value function ${{Q}_{j}}({\bm{\tau }_{t}},\bm{u}_{t}^{j})$ as a network parameterized by ${{\theta }_{j}}$. The joint value function, represented by the factor graph $G$, is given by
\begin{equation}
Q({\bm{\tau }_{t}},{\bm{u}_{t}},{{A}_{t}})=\sum\limits_{j\in \mathcal{J}}{{{f}_{j,{{\theta }_{j}}}}(\bm{u}_{t}^{j}\mid{\bm{\tau }_{t}})}=\sum\limits_{j\in J}{{{Q}_{j}}(\bm{u}_{t}^{j}\mid{\bm{\tau }_{t}};{{\theta }_{j}})}
\end{equation}
where $\mathcal{J}$ is the set of factor nodes, and $\bm{u}_{t}^{j}$ denotes the joint action of all agents at time $t$.

DCG uses a deep network to learn the coordination graph's utility and payoff functions. Similarly, DDFG extends DCG to factor graphs by learning the functions corresponding to the factor nodes in the factor graph through a deep network. We define the "order" of the value function ${{Q}_{j}}$ as $D\left( j \right)$, which represents the number of agent ${j}$'s connections. The order of these value functions can exceed the maximum 2 in the coordination graph, making the value function decomposition network more representable. We present the design principles for the Q-value function network of the DDFG as follows:

\begin{enumerate}
        \item[a)]  
	The local value function ${{Q}_{j}}$ accepts only the local information of the agent $i\in n({{Q}_{j}})$.      
	\item[b)] 
	Adopt a common recurrent neural network sharing parameters across all agents (yellow part in Figure \ref{fig:2}).
	\item[c)] 
	Employing a common fully connected neural network sharing parameters on a local value function ${{Q}_{j}}$ of the same order $D\left( j \right)$ (blue part in Figure \ref{fig:2}).
	\item[d)] 
	Use tensor-based Canonical Polyadic (CP) decomposition for the local value function ${{Q}_{j}}$ of $D(j)\ge 2$.
        \item[e)] 
	Allowing generalization to different factor graph structures.
\end{enumerate} 

According to principle a), the local value function ${{Q}_{j}}(\bm{u}_{t}^{j}\mid{\bm{\tau }_{t}};{{\theta }_{j}})={{Q}_{j}}(\bm{u}_{t}^{j}\mid\bm{\tau}_{t}^{j};{{\theta }_{j}})$, where $\bm{\tau}_{t}^{j}$ denotes the joint observation-action history of all agents $i\in n({{Q}_{j}})$. Based on principle b), all agents share parameters using a recurrent neural network (RNN) with generic architecture, denoted as $h_{t}^{i}={{h}_{\psi }}(\cdot \mid{h}_{t-1}^{i},o_{t}^{i},u_{t-1}^{i})$. This RNN is initialized with $h_{0}^{i}={{h}_{\psi }}(\cdot \mid\bm{0},o_{0}^{i},\bm{0})$. According to principle c), the local value function ${{Q}_{j}}(\bm{u}_{t}^{j}\mid\bm{\tau}_{t}^{j};{{\theta }_{j}})$ can be approximated as ${{Q}_{j}}(\bm{u}_{t}^{j}\mid\bm{h}_{t}^{j};{{\theta }_{D(j)}})$, where all local value functions of order $D\left( j \right)$ share the parameter ${{\theta }_{D(j)}}$, improving operational efficiency. The local value functions are spliced together using the matrix $\bm{h}_{t}^{j}=CONCAT[h_{t}^{i},h_{t}^{i+1},\ldots ],i\in n({{Q}_{j}})$.

The output ${{\mathcal{U}}^{D(j)}}$ of the local value function ${{Q}_{j}}$ grows exponentially with the order $D\left( j \right)$. Since only executed action pairs are updated during Q-learning, the parameters of many outputs remain constant for a long time while the underlying RNN network is continuously updated. This can slow down the training speed and affect the message delivery. To reduce the number of parameters and increase the frequency of their updates, we extend the low-rank approximation of matrices in DCG\cite{dcg} and propose to use the Canonical Polyadic (CP) decomposition of the tensor\cite{phan2021,zhou2019} to approximate the value function ${{Q}_{j}}$. The CP decomposition of ${{Q}_{j}}$ of rank K is defined by:

% \begin{figure*}[htb]
% \begin{equation}\label{eq:2}
% {{Q}_{j}}({\bm{u}^{j}}\mid\bm{h}_{t}^{j};{{\theta }_{D(j)}}) 
% :=\sum\limits_{k=1}^{K}{\left( {{f}_{k}}(\bm{u}_{t}^{j}\mid\bm{h}_{t}^{j};\theta_{D(j)}^{1})\circ {{f}_{k}}(\bm{u}_{t}^{j}\mid\bm{h}_{t}^{j};\theta_{D(j)}^{2})\circ \ldots \circ {{f}_{k}}(\bm{u}_{t}^{j}\mid\bm{h}_{t}^{j};\theta _{D(j)}^{D(j)}) \right)}   
% \end{equation}
% \end{figure*}

\begin{equation}\label{eq:2}
\begin{aligned}
  & {{Q}_{j}}({\bm{u}^{j}}|\bm{h}_{t}^{j};{{\theta }_{D(j)}}):=\sum\limits_{k=1}^{K}{({{f}_{k}}(\bm{u}_{t}^{j}|\bm{h}_{t}^{j};\theta _{D(j)}^{1})} \\ 
 & \circ {{f}_{k}}(\bm{u}_{t}^{j}|\bm{h}_{t}^{j};\theta _{D(j)}^{2})\circ \ldots \circ {{f}_{k}}(\bm{u}_{t}^{j}|\bm{h}_{t}^{j};\theta _{D(j)}^{D(j)})) \\ 
\end{aligned}
\end{equation}

To approximate the value function in our reinforcement learning model, we use a network of local value functions with parameters $\{\theta_{D(j)}^{d}\}_{d=1}^{D(j)}$ and output $D(j)KA$ (as described in Eq.(\ref{eq:2}). The tensor rank is determined by balancing the approximation's accuracy against the parameter learning speed.

However, as the adjacency matrix ${{A}_{t}}$ changes in real time, the value of the global value function becomes unstable. To address this, we fix all local value functions by setting $D(j)=1$, which yields the global value function $Q({\bm{\tau}_{t}},{\bm{u}_{t}})$ plus the VDN decomposition expressed as ${{Q}_{vdn}}$. This results in a new adjacency matrix ${{A}_{t}}^{\prime }=CONCAT[{{A}_{t}},{{I}_{n}}]$. The final global value function is then given by:

\begin{equation}
\begin{aligned}
& {{Q}_{tot}}({\bm{\tau}_{t}},{\bm{u}_{t}},{{A}_{t}};\theta ,\psi )=Q({\bm{\tau }_{t}},{\bm{u}_{t}})+{{Q}_{vdn}} \\
& =\sum\limits_{j\in \mathcal{J}}{{{Q}_{j}}(\bm{u}_{t}^{j}\mid\bm{h}_{t}^{j};{{\theta }_{D(j)}})}+\sum\limits_{i=1}^{n}{{{Q}_{i}}(u_{t}^{i}\mid{h}_{t}^{i};{{\theta }_{D(1)}})} \\
\end{aligned}
\end{equation}

\subsection{Graph structure generation policy}
At each time step, we represent the graph policy as $\rho ({{A}_{t}}\mid{\bm{\tau }_{t}})$. With the graph policy $\rho $, we can obtain the adjacency matrix ${{A}_{t}}\sim \rho $. As described in Section 3.1, the decomposition of a global value function is associated with ${{A}_{t}}$. However, the decomposition of the global value function should not be static, and the collaborative relationship among the agents should be dynamic when facing different environmental states. While the factor graph can decompose the global value function, its decomposition corresponds to a fixed adjacency matrix. Similarly, DCG decomposes the global value function into a fixed, fully connected structure. Therefore, this section proposes a graph policy $\rho $ that can dynamically generate real-time graph structures that represent the changing collaborative relationships among agents.

We propose a graph policy $\rho$ that can dynamically generate real-time graph structures to address this issue. We represent the graph policy $\rho ({{A}_{t}}\mid{\bm{\tau }_{t}})=\rho ({{A}_{t}}\mid{\bm{\tau }_{t}};\varphi )$ using a network parameterized by $\varphi$. The graph policy network shares the same RNN network as the Q-value function network but does not participate in the parameter update. We use $h_{t}^{i}={{h}_{\psi }}(\cdot \mid{h}_{t-1}^{i},o_{t}^{i},u_{t-1}^{i})$ as the input of the graph policy network. To better represent the global relationships of all agents, we use the global states ${{s}_{t}}$ as the input of the hypernetwork to obtain the parameters of the graph policy network. We refer to the network structure design of QMIX and train the weights W of the hypernetwork without absolute value restrictions to obtain more information in ${{s}_{t}}$. 

We employ a two-layer hypernetwork and incorporate a fully connected layer as an action layer to initialize the probability of each edge equally. Subsequently, we apply a softmax activation function to process the output of the graph structured network, constraining the output between 0 and 1, in order to derive the probability of edge connections corresponding to the adjacency matrix $A_t$. Specifically, the graph policy $\rho(A_t\mid \bm{\tau}_t)$ outputs a matrix $P({{A}_{t}})={{\{{{a}_{ij}}\}}^{N\times M}}$, representing the probabilities of edge connections corresponding to the adjacency matrix $A_t$, where $N$ is the number of intelligent agents (variable nodes), $M$ is the number of local value functions (factor nodes), and $a_{ij}$ represents the probability of connection between the i-th intelligent agent and the j-th local value function $(\sum_{i=0}^{N}a_{ij}=1)$. For each local value function $Q_j$, the probability distribution of its connections with all $N$ agents is defined as a multinomial distribution.

\begin{definition}
In the graph policy $\rho ({{A}_{t}}\mid{{\tau }_{t}})$, for each local value function ${{Q}_{j}}$, randomly Variable $X=({{X}_{1}},{{X}_{2}},\ldots ,{{X}_{N}})$ represents the number of connections between ${{Q}_{j}}$ and N agents, where ${{X}_{i}}$ represents the number of times the i-th agent is connected to ${{Q}_{j}}$. $X$ satisfies:
\begin{enumerate}
\item[(1)] ${{X}_{i}}\ge 0(1\le i\le N)$, and ${{X}_{1}}+{{X}_{2}}+\cdots +{{X}_{N}}=m$, $m$ is the total number of connections between ${{Q}_{j}}$ and N agents;
\item[(2)]  Let ${{m}_{1}},{{m}_{2}},\ldots ,{{m}_{N}}$ be any non-negative integer, and ${{m} _{1}}+{{m}_{2}}+\cdots +{{m}_{N}}=m$, then the probability of event $\{{{X}_{1}}={{m}_{1}},{{X}_{2}}={{m}_{2}},\ldots ,{{X}_{N}}={{m}_{N}}\}$ occurring is: 
\begin{equation}
\begin{aligned}
& P\{{{X}_{1}}={{m}_{1}},{{X}_{2}}={{m}_{2}},\ldots ,{{X}_{N}}={{m}_{N}}\} \\
& =\frac{m!}{{{m}_{1}}!{{m}_{2}}!\cdots {{m}_{N}}!}{{p}_{1}}^{{{m}_{1}}}{{p}_{2}}^{{{m}_{2}}}\cdots {{p}_{N}}^{{{m}_{N}}} \\
\end{aligned}
\end{equation}
where, ${{p}_{i}}\ge 0(1\le i\le N),{{p}_{1}}+{{p}_{2}}+\cdots +{{p}_{N}}=1$.
\end{enumerate}
Then $X$ conforms to the multinomial distribution, that is, $X\sim {{P}_{m}}(m:{{p}_{1}},{{p}_{2}},\ldots ,{ {p}_{N}})$.
\end{definition}

\begin{proposition}\label{pro1}
In the graph policy $\rho ({{A}_{t}}\mid{\bm{\tau }_{t}})$, for each local value function ${{Q}_{j}}$, when The number of times it is connected to N agents corresponds to a random variable $X\sim {{P}_{{{D}_{\max }}}}({{D}_{\max }}:{{ p}_{1}},{{p}_{2}},\ldots ,{{p}_{N}})$, and get ${{D}_{\max }}$ as in the algorithm Maximum order.
\end{proposition}

\begin{proposition}\label{pro2}
In the graph policy $\rho ({{A}_{t}}\mid{\bm{\tau }_{t}})$, for each local value function ${{Q}_{j}}$, when The number of times it is connected to N agents corresponds to a random variable $X\sim {{P}_{{{D}_{\max }}}}({{D}_{\max }}:{{ p}_{1}},{{p}_{2}},\ldots ,{{p}_{N}})$, then each ${{Q}_{j}}$ corresponds to "sub-policy" conforms to a quasi-multinomial distribution, that is, $\rho ({{Q}_{j}})\sim {{\tilde{P}}_{{{D}_{\max }}}}( {{D}_{\max }}:{{p}_{1}},{{p}_{2}},\ldots ,{{p}_{N}})$.
\end{proposition}

Detailed proof of Proposition \ref{pro1},\ref{pro2} can be found in appendix C. Through the above definitions and inferences, we represent the graph structure generation policy as a a quasi-multinomial distribution, and train it with a neural network. Meanwhile, it can be concluded that by setting the maximum order ${{D}_{\max }}$ of the graph policy network, the factor graph can represent all collaborative relationships of less than or equal to ${{D}_{\max }}$ agents.

\subsection{Max-plus algorithm}
In the previous section, we obtained the local value functions using the Q-value function network. In this section, we will demonstrate how to use the max-plus algorithm\cite{zhang2014,kok2006} on factor graphs to obtain the final action for each agent.

First, we obtain the adjacency matrix ${{A}_{t}}$ from the graph structure generation policy $\rho $ and concatenate it with ${{I}_{n}}$ to form ${{{A}_{t}}^{\prime}}$. To facilitate the derivation, we consider all factor nodes represented by ${{{A}_{t}}^{\prime}}$ as a single entity. Then, the action selection process aims to solve the following problem(See appendix C for detailed derivation):
\begin{equation}
\begin{aligned}
  & \bm{u}_{t}^{*}=\underset{{\bm{u}_{t}}}{\mathop{\arg \max }}\,{{Q}_{tot}}({\bm{\tau }_{t}},{\bm{u}_{t}},{{A}_{t}};\theta ,\psi ) \\ 
 & \quad =\underset{({\bm{u}_{1}},\ldots ,{\bm{u}_{n}})}{\mathop{\arg \max }}\,(\sum\limits_{j\in {J}'}{{{Q}_{j}}({{\{{{v}_{i}}\}}_{\{i,j\}\in {\mathcal{E}}'}})}) \\ 
\end{aligned}
\end{equation}
where ${\mathcal{J}}'=\mathcal{J}\bigcup \{m+i\}_{i=1}^{n}$,${\mathcal{E}}'=\mathcal{E}\bigcup \{\{i,j\}\mid1\le i\le n,j=j+1(j=1\sim n)\}$.

In illustrating the max-plus algorithm for factor graphs, we will omit moment $t$'s representation. The max-plus algorithm for factor graphs consists of two types of messages. Let $N(x)$ denote the set of neighbors of node $x$. The message sent from the variable node ${{v}_{i}}$ to the factor node ${{Q}_{j}}$ is:
\begin{equation}\label{eq:7}
    {{\mu }_{{{v}_{i}}\to{{Q}_{j}}}}({{v}_{i}})=\sum\limits_{{{Q}_{k}}\in N({{v}_{i}})\backslash \{{{Q}_{j}}\}}{{{\mu }_{{{Q}_{k}}\to{{v}_{i}}}}({{v}_{i}})}+{{c}_{{{v}_{i}}\to{{Q}_{j}}}}
\end{equation}
where $N({{v}_{i}})\backslash \{{{Q}_{j}}\}$ denotes the set of nodes in $N({{v}_{i}})$ except ${{Q}_{j}}$, and ${{c}_{{{v}_{i}}\to{{Q}_{j}}}}$ is the normalization term.

The message sent from the factor node ${{Q}_{p}}$ to the variable node ${{v}_{l}}$ is:
\begin{equation}\label{eq:8}
\begin{aligned}
  &  {{\mu }_{{{Q}_{p}}\to {{v}_{l}}}}({{v}_{l}})=\underset{{\bm{v}_{p}}\backslash {{v}_{l}}}{\mathop{\max }}\,({{Q}_{p}}({\bm{v}_{p}}) \\
  &  +\sum\limits_{{{v}_{k}}\in N({{Q}_{p}})\backslash \{{{v}_{l}}\}}{{{\mu }_{{{v}_{k}}\to {{Q}_{p}}}}({{v}_{k}})})+{{c}_{{{Q}_{p}}\to {{v}_{l}}}} \\
\end{aligned}
\end{equation}
where ${\bm{v}_{p}}\backslash {{v}_{l}}$ denotes the parameters in the local function ${{Q}_{p}}$ except ${{v}_{l}}$, and ${{c}_{{{Q}_{p}}\to {{v}_{l}}}}$ is the normalization term. The agent continuously sends, accepts, and recomputes messages until the values of the messages converge. After convergence, the behavior of the agent is given by the following equation:
\begin{equation}
  {\bm{u}^{*}}=[\underset{{{u}_{i}}}{\mathop{\arg \max }}\,\sum\limits_{{{Q}_{k}}\in N({{v}_{i}})}{{{\mu }_{{{Q}_{k}}\to {{v}_{i}}}}({{v}_{i}})}]_{i=1}^{n}
\end{equation}

However, this convergence exists only in acyclic factor graphs, meaning that the exact solution can only be achieved in such graphs through the max-plus algorithm. When the factor graph contains loops, as depicted in Figure \ref{fig:1}, the general max-plus algorithm cannot guarantee an exact solution due to the uncertain manner nodes receive messages containing messages sent from the node, leading to message explosion. Asynchronous messaging \cite{lan2006} and messaging with damping \cite{som2010} can mitigate this issue.

\section{Loss function}
The value function and graph policy networks use different loss functions and alternate updates. This section will explain the form of loss construction for each network.

\subsection{Loss of the Q-value function network}
We update the Q-value function network using the loss form that is referred to in DQN. We maintain a target network \cite{van2016} and a replay buffer \cite{foerster2017}. We also deposit the adjacency matrix $A$, obtained from the graph policy, into the replay buffer. This way, we obtain the decomposed form of the global value function ${{Q}_{tot}}$ through the adjacency matrix $A$. In contrast, the decomposed local value function ${{Q}_{j}}$ is obtained through the $\theta,\psi$ parameterized Q-value function network. Finally, we use the globally optimal joint action ${\bm{u}^{*}}$ obtained by the max-sum algorithm. To achieve this, we learn $\theta,\psi$ by minimizing the TD error, as follows:
\begin{equation}
\begin{aligned}
 &   {{\mathcal{L}}_{Q}}\left( \bm{\tau} ,A,\bm{u},r;\theta ,\psi  \right):= \\
 &   E\left[ \frac{1}{T}\sum\limits_{t=0}^{T}{{{\left( {{Q}_{tot}}({\bm{\tau }_{t}},{\bm{u}_{t}},{{A}_{t}})-{{y}^{dqn}}\left( {{r}_{t}},{{\tau }_{t+1}},{{A}_{t+1}};\bar{\theta },\bar{\psi } \right) \right)}^{2}}} \right] \\
\end{aligned}
\end{equation}
where $\bm{\tau} =\{{\bm{\tau }_{t}}\}_{t=1}^{T}$ (note: $\bm{\tau} $ in this paper only denotes the set of observation-action histories ${\bm{\tau }_{t}}$, not the trajectory of the whole task), and ${{r}_{t}}$ is the reward for performing action ${\bm{u}_{t}}$ transitions to ${\bm{\tau }_{t+1}}$ in the observation history ${\bm{\tau }_{t}}$.${{y}^{dqn}}\left( {{r}_{t}},{\bm{\tau }_{t+1}},{{A}_{t+1}};\bar{\theta },\bar{\psi } \right)={{r}_{t}}+\gamma {{Q}_{tot}}({\bm{\tau }_{t+1}},u_{t+1}^{*},{{A}_{t+1}};\bar{\theta },\bar{\psi })$,$\bm{u}_{t+1}^{*}=[\underset{u_{t+1}^{i}}{\mathop{\arg \max }}\,\sum\limits_{{{Q}_{k}}\in N({{v}_{i}})}{{{\mu }_{{{Q}_{k}}\to{{v}_{i}}}}({{v}_{i}})}]_{i=1}^{n}$,$\bar{\theta },\bar{\psi }$ is the parameter copied periodically from $\theta ,\psi$.

\subsection{Loss of graph policy network}
MARL aims to maximize the expected discounted sum of future rewards. Therefore, the natural objective of the graph policy network is to find the optimal graph policy $\rho ({{A}_{t}}\mid{\bm{\tau }_{t}};\varphi )$ that achieves this goal. However, if we use a DQN-style loss function, the gradient cannot be returned to the graph policy network, making training impossible. To address this issue, we adopt the policy gradient approach \cite{schulman2017} and treat the graph policy $\rho ({{A}_{t}}\mid{\bm{\tau }_{t}};\varphi )$ as the action policy $\pi \left( {\bm{u}_{t}}\mid{\bm{\tau }_{t}} \right)$, and use the policy gradient approach to design the graph policy network's loss function.

The loss function for the graph policy network is:
\begin{equation}
  {{\mathcal{L}}_{G}}\left( \bm{\tau} ,A,\bm{u},r;\varphi  \right)={{\mathcal{L}}_{PG}}-{{\lambda }_{\mathcal{H}}}{{\mathcal{L}}_{entropy}}
\end{equation}
where ${{\lambda }_{\mathcal{H}}}$ is the weight constant of ${{\mathcal{L}}_{entropy}}$. Our objective is to maximize the expected discounted return, for which we define the maximization objective function as follows:
\begin{equation}
    \underset{\varphi }{\mathop{maximize}}\,\ \ J(\theta ,\psi ,\varphi )=\underset{\varphi }{\mathop{maximize}}\,\ \ {{v}_{{{\Pi }_{\theta ,\psi ,\varphi }}}}\left( {{s}_{0}} \right)
\end{equation} 
where ${{\Pi }_{\theta ,\psi ,\varphi }}={{\Pi }_{\theta ,\psi ,\varphi }}\left( \bm{u},A\mid s \right)={{\rho }_{\varphi }}\left( A\mid s \right){{\pi }_{\theta ,\psi }}\left( \bm{u}\mid s,A \right)$ denotes the joint policy of all agent policies and graph policies, ${{v}_{{{\Pi }_{\theta ,\psi ,\varphi }}}}\left( {{s}_{0}} \right)$ is the policy ${{\Pi }_{\theta ,\psi ,\varphi }}$ under ${{s}_{0}}$ as a function of the state value of ${{s}_{0}}$. The derivation of the objective function $J(\cdot)$ yields (see appendix C for the corresponding derivation):
\begin{equation}\label{eq:3}
\begin{aligned}
  & {{\nabla }_{\varphi }}J(\theta ,\psi ,\varphi )= \\
  & \underset{\left( {\bm{\tau }_{t}},{\bm{u}_{t}},{{A}_{t}} \right)\sim \mathcal{T}}{\mathop{E}}\,\left[ {{Q}_{tot}}\left( {\bm{\tau }_{t}},{\bm{u}_{t}},{{A}_{t}} \right)\nabla \ln {{\rho }_{\varphi }}\left( {{A}_{t}}|{\bm{\tau }_{t}} \right) \right] \\ 
\end{aligned}
\end{equation}

We refer to the PPO \cite{schulman2017}, which uses importance sampling for graph policies to improve training efficiency:
\begin{equation}\label{eq:x}
\begin{aligned}
  & {{\nabla }_{\varphi }}J= \\
  & \underset{\left( {\bm{\tau }_{t}},{\bm{u}_{t}},{{A}_{t}} \right)\sim {{\mathcal{T}}_{old}}}{\mathop{E}}\,[ \frac{{{\rho }_{\varphi }}\left( {{A}_{t}}\mid{\bm{\tau }_{t}} \right)}{{{\rho }_{{{\varphi }_{old}}}}\left( {{A}_{t}}\mid{\bm{\tau }_{t}} \right)}{{Q}_{tot}}\left( {\bm{\tau }_{t}},{\bm{u}_{t}},{{A}_{t}} \right) \\
  & \nabla \ln {{\rho }_{\varphi }}\left( {{A}_{t}}\mid{\bm{\tau }_{t}} \right) ] \\
\end{aligned}
\end{equation}

In order to get a more accurate graph policy $\rho$, we use the local value function ${{Q}_{j}}$ instead of the global value function ${{Q}_{tot}}$ for calculation. The graph policy $\rho$ can be decomposed into "sub-policy" in which all local value functions are connected to the agent. Comparing the value function used in the actor-critic algorithm to evaluate the quality of the policy, in DDFG the global value function ${{Q}_{tot}}$ can evaluate the graph policy $\rho$ Then it is natural to use the local value function to evaluate the quality of the "sub-policy". Essentially, this is a credit allocation idea. The global value function ${{Q}_{tot}}$ is the sum of the local value functions ${{Q}_{j}}$, The larger ${Q}_{j}$, the greater its contribution to ${{Q}_{tot}}$ as ${{Q}_{j}}$ whole, which means that the corresponding local The greater the contribution of the value function ${{Q}_{j}}$ to the "sub-policy" connected to the agent. Therefore, the "sub-policy" should be accurately evaluated by ${{Q}_{j}}$, then ${{\nabla }_{\varphi }}J$ becomes:
\begin{equation}\label{eq:5}
\begin{aligned}
  & {{\nabla }_{\varphi }}J= \\ 
 & \underset{\left( {\bm{\tau }_{t}},{\bm{u}_{t}},{{A}_{t}} \right)\sim{{\mathcal{T}}_{old}}}{\mathop{E}}\,[\sum\limits_{j\in \mathcal{J}}{\left( \frac{{{P}_{\varphi ,j}}\left( {{m}_{j}} \right)}{{{P}_{{{\varphi }_{old}},j}}\left( {{m}_{j}} \right)}{{Q}_{j}}\left( \bm{\tau}_{t}^{j},\bm{u}_{t}^{j},{{A}_{t}} \right) \right)} \\ 
 & \nabla \ln {{\rho }_{\varphi }}\left( {{A}_{t}}\mid {\bm{\tau }_{t}} \right)] \\ 
\end{aligned}
\end{equation}
where ${{P}_{\varphi ,j}}\left( {{m}_{j}} \right)=\sum\limits_{\{{{m}_{j,1}},\ldots ,{{m}_{j,N}}\}\in {{H}_{m}}}{p_{j,1}^{{{m}_{j,1}}}\cdots p_{j,i}^{{{m}_{j,i}}}\cdots p_{j,N}^{{{m}_{j,N}}}}$, ${{p}_{j,i}}$ represents the probability that the local value function ${{Q}_{j}}$ is connected to agent i, ${{m}_{j,i}}$ represents the number of times the local value function ${{Q}_{j}}$ is connected to agent i, ${{m}_{j}}={{\{{{m}_{ j,i}}\}}_{i=1-N}}$.

\begin{proposition}\label{pro3}
In the graph policy $\rho ({{A}_{t}}\mid{\bm{\tau }_{t}})$, for each local value function ${{Q}_{j}}$, when Its corresponding "sub-policy" $\rho ({{Q}_{j}})\sim {{\tilde{P}}_{{{D}_{\max }}}}({{D} _{\max }}:{{p}_{1}},{{p}_{2}},\ldots ,{{p}_{N}})$, then Eq.(\ref{eq:5}) can be derived from Eq.(\ref{eq:x}).
\end{proposition}

Detailed proof of Proposition \ref{pro3} can be found in appendix C.

Generalized advantage estimation (GAE) \cite{schulman2015} is adopted to estimate the advantage function $\mathcal{A}$ and use the advantage function ${{\mathcal{A}}_{_{j}}}$ instead of the local value function ${{Q}_{j}}$, thus reducing the variance:
\begin{equation}
    \hat{\mathcal{A}}_{j}^{GAE}\left( \bm{\tau}_{t}^{j},\bm{u}_{t}^{j},{{A}_{t}} \right):=\sum\limits_{t=0}^{\infty }{{{\left( \gamma {{\lambda }_{GAE}} \right)}^{l}}\delta _{t+l}^{{{Q}_{j}}}}
\end{equation}
where ${{\lambda }_{GAE}}$ is the discount factor in GAE, $\delta _{t}^{{{Q}_{j}}}={{Q}_{j}}(\bm{\tau} _{t}^{j},\bm{u}_{t}^{j},A)-{{V}_{j}}(\bm{\tau}_{t}^{j})$. We construct the relationship between ${V}_{tot}$ and ${V}_{j}$ using the same ${A}_{t}$ and train ${V}_{tot}$ with TD-error (see appendix B for detailed descriptions).

We use the clip function to constrain Importance sampling, and derive ${{\nabla }_{\varphi }}{{\mathcal{L}}_{PG}}$:
\begin{equation}\label{eq:6}
\begin{aligned}
  & {{\nabla }_{\varphi }}{{\mathcal{L}}_{PG}}= \\ 
 & =-{{E}_{\left( {\bm{\tau }_{t}},{\bm{u}_{t}},{{A}_{t}} \right)\sim \mathcal{D}}}[\sum\limits_{j}{\min \left( r_{t}^{j}\left( \varphi  \right){{{\hat{\mathcal{A}}}}_{j}},{{f}_{clip}}(r_{t}^{j}\left( \varphi  \right),\epsilon ){{{\hat{\mathcal{A}}}}_{j}} \right)} \\ 
 & \nabla \ln {{\rho }_{\varphi }}\left( {{A}_{t}}|{\bm{\tau }_{t}} \right)] \\ 
\end{aligned}
\end{equation}
where $r_{t}^{j}\left( \varphi  \right)={{{P}_{\varphi ,j}}\left( {{m}_{j}} \right)}/{{{P}_{{{\varphi }_{old}},j}}\left( {{m}_{j}} \right)}\;$, ${{f}_{clip}}(r_{t}^{j}\left( \varphi  \right),\epsilon )=clip\left( r_{t}^{j}\left( \varphi  \right),1-\epsilon ,1+\epsilon  \right)$, $\mathcal{D}$ is the replay buffer of the graph policy network and is independent of the replay buffer of the Q-value function network.

${{{\mathcal{L}}}_{entropy}}$ is the loss function of the policy entropy. We add the policy entropy to the loss to improve the exploration ability of the graph policy network:
\begin{equation}
\begin{aligned}
& {{\mathcal{L}}_{entropy}}=-{{E}_{{\bm{\tau }_{t}}\sim \mathcal{D}}}[\mathcal{H}\left( \rho (\cdot \mid{\bm{\tau }_{t}};\varphi ) \right)] \\
& =-{{E}_{{\bm{\tau }_{t}}\sim \mathcal{D}}}[\sum\limits_{i}{\sum\limits_{j}{{{\rho }_{ij}}(\cdot \mid{\bm{\tau }_{t}};\varphi )\log \left( {{\rho }_{ij}}(\cdot \mid{\bm{\tau }_{t}};\varphi ) \right)}}] \\
\end{aligned}
\end{equation}
where ${{\rho }_{ij}}(\cdot \mid{{\tau }_{t}};\varphi )$ denotes the probability of connecting agent i with the local function ${{Q}_{j}}$.

\section{Experiment}
This section delineates a comparative analysis of the Dynamic Deep Factor Graph (DDFG) algorithm against a suite of state-of-the-art algorithms including Value Decomposition Networks (VDN) \cite{vdn}, QMIX \cite{qmix}, Combined-Weighted QMIX (CW-QMIX)/Optimistically Weighted QMIX (OW-QMIX) \cite{wqmix}, Multi-Agent Deep Deterministic Policy Gradient (MADDPG) \cite{lowe2017}, QTRAN \cite{qtran}, QPLEX \cite{qplex}, Deep Coordination Graphs (DCG) \cite{dcg}, Sparse Optimistic Policy Coordination Graph (SOPCG) \cite{sopcg}, and Coordination Among Sparsely Communicating Entities (CASEC) \cite{casec}. The implementation of DDFG, constructed using PyTorch, is publicly available alongside baseline experiment codes in the designated repository.

To elucidate the efficacy of DDFG, we evaluate it across two distinct scenarios, detailed further in appendix D with regards to hyperparameter configuration and additional experiments: (1) an advanced Higher-Order Predator-Prey model \cite{dcg}, and (2) the StarCraft II Multi-agent Challenge (SMAC) \cite{samvelyan2019}.

\begin{figure}
  \setlength{\abovecaptionskip}{0cm}
  \setlength{\belowcaptionskip}{-0.5cm}
  \centering
  \includegraphics[scale=0.35]{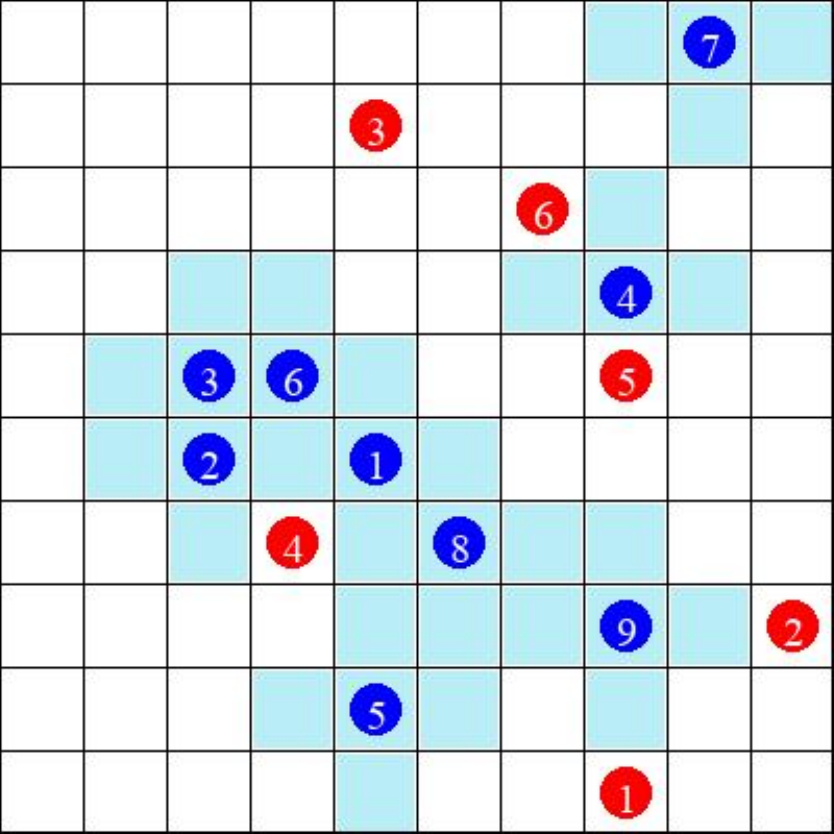}
  \caption{HO-Predator-Prey environment. Predators are marked in blue, prey is marked in red, and the blue grid represents the range of movement of the predator.}
  \label{fig:hopp}
\end{figure}

\begin{figure*}
  \setlength{\abovecaptionskip}{0cm}
  \setlength{\belowcaptionskip}{-0.5cm}
  \centering
  \subfloat{
	\begin{minipage}[t]{0.5\linewidth}
		\centering
		\includegraphics[width=3.3in,height=3in]{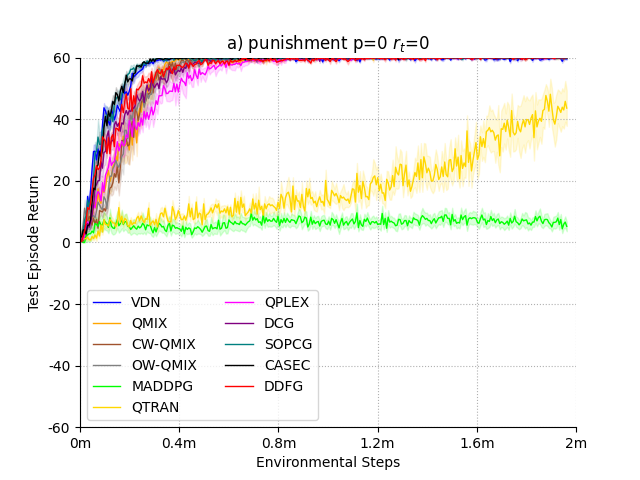}
        \end{minipage}
        \label{fig:4a}
  }%
  \subfloat{
	\begin{minipage}[t]{0.5\linewidth}
		\centering
		\includegraphics[width=3.3in,height=3in]{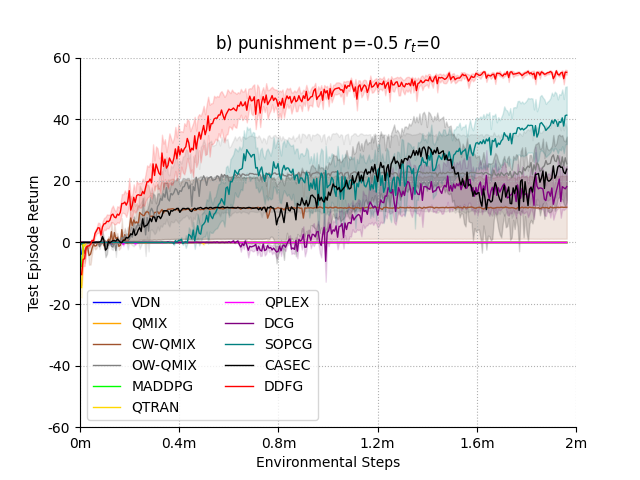}
        \end{minipage}
        \label{fig:4b}
  }%

 \subfloat{
	\begin{minipage}[t]{0.5\linewidth}
		\centering
		\includegraphics[width=3.3in,height=3in]{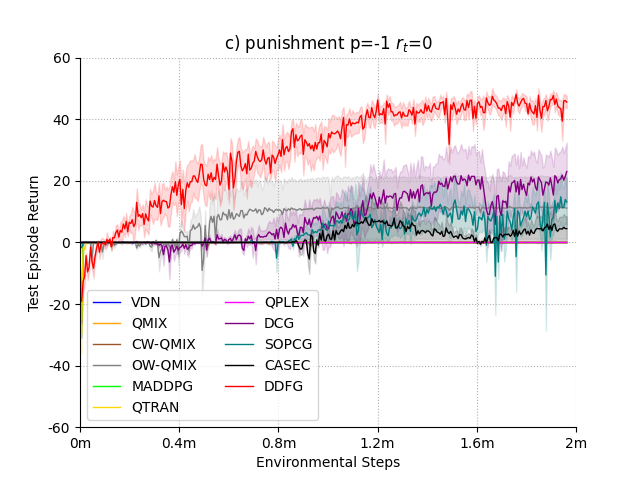}
	\end{minipage}
        \label{fig:4c}
  }%
  \subfloat{
	\begin{minipage}[t]{0.5\linewidth}
		\centering
		\includegraphics[width=3.3in,height=3in]{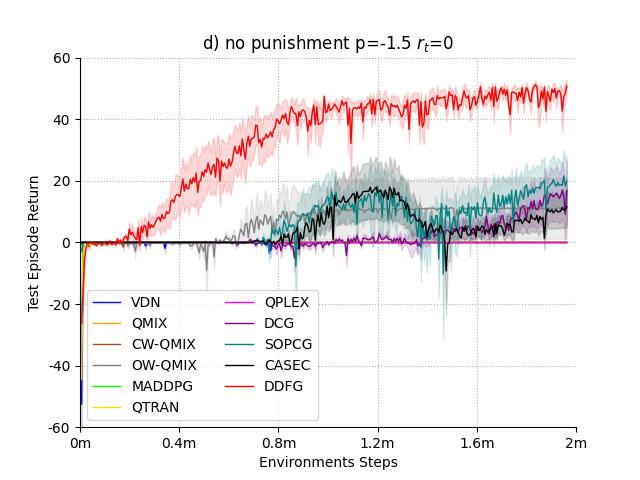}
        \end{minipage}
        \label{fig:4d}
  }
  \caption{Median test return for the Higher-order Predator-Prey task with different penalties p(0,-0.5,-1,-1.5), comparing DDFG and baselines.}
  \label{fig:4}
\end{figure*}

\subsection{Higher-order Predator-Prey}
The Predator-Prey environment, as utilized in DCG, was augmented to engender a Higher-Order Predator-Prey (HO-Predator-Prey) scenario, exhibiting increased complexity.

\textbf{Original Predator-Prey Scenario:} This model positions the prey on a grid, where agents may execute capture actions. Successive captures by two or more agents on the prey, positioned in adjacent squares (top, bottom, left, or right), result in a collective reward of $r$. Conversely, an unsuccessful solo capture action inflicts a sub-reward penalty of $p$.

\textbf{Higher-order Predator-Prey (HO-Predator-Prey):} The HO-Predator-Prey scenario (as shown in Figure \ref{fig:hopp}) extends the adjacency to include eight squares around the prey: Upper, Lower, Left, Right, Upper Left, Upper Right, Lower Left, and Lower Right, necessitating $X$ or more simultaneous capture actions for success, with $X$ set to 3. Furthermore, the decomposition of the DDFG value function is constrained to a highest order of 3, linking each local value function with no more than three agents. 

Our experimental framework evaluated the influence of varying penalty values p (0, -0.5, -1, -1.5), with the outcomes depicted in Figure \ref{fig:4}. In scenarios devoid of penalties (Figure \ref{fig:4a}), a broad array of algorithms including Value Decomposition Networks (VDN), QMIX, Combined-Weighted QMIX (CW-QMIX), Optimistically Weighted QMIX (OW-QMIX), QPLEX, Deep Coordination Graphs (DCG), Sparse Optimistic Policy Coordination Graph (SOPCG), Coordination Among Sparsely Communicating Entities (CASEC), and Dynamic Deep Factor Graph (DDFG) converged to the optimal solution. In contrast, QTRAN and Multi-Agent Deep Deterministic Policy Gradient (MADDPG) only managed to achieve partial success. Notably, DDFG exhibited a reduced rate of convergence, attributable to its ongoing optimization of graphical policies concurrent with value function fitting, which inherently diminishes convergence velocity. Upon the introduction of penalties (Figures \ref{fig:4b}-\ref{fig:4d}), all baseline algorithms except for CW-QMIX, OW-QMIX, DCG, SOPCG, CASEC, and DDFG succumbed to failure, unable to circumvent the pitfall of relative overgeneralization. As penalties intensified, CW-QMIX, OW-QMIX, DCG, SOPCG, and CASEC progressively struggled to attain the optimal solution, with complete failures recorded in certain trials.

When penalty $p$ was set to -0.5, CW-QMIX exhibited partial success; however, as the penalty increased to -1 or beyond, it failed outright. OW-QMIX, despite partial successes across varying penalty levels, encountered difficulties in accurate action learning as penalties escalated. This is in stark contrast to the outright failure of QMIX, where CW-QMIX/OW-QMIX's partial successes can be attributed to the Weighted QMIX operator's non-monotonic function mapping, thus boosting algorithm performance. Despite DCG's ability to learn successfully across all scenarios, its average reward and convergence rate fell short of DDFG's performance. This discrepancy is linked to DCG's static coordination graph structure, which hinders learning of collaborative policies involving more than two agents due to redundant information in message passing and an inability to adapt to complex collaborative decisions among three or more agents. Conversely, SOPCG and CASEC, despite their advancement in dynamic sparse graph structures, were hampered by the intrinsic limitations of the coordination graph's expressive capacity, thus failing to achieve comprehensive success. DDFG, however, demonstrated proficiency in learning optimal policies across all scenarios by dynamically modeling cooperation among predators through graph policy networks and employing factor graphs for collaborative decision-making among multiple agents, facilitated by the max-plus algorithm.

\begin{figure*}
  \setlength{\abovecaptionskip}{0cm}
  \setlength{\belowcaptionskip}{-0.5cm}
  \centering
  \subfloat{
	\begin{minipage}[t]{0.5\linewidth}
		\centering
		\includegraphics[width=3.3in,height=3in]{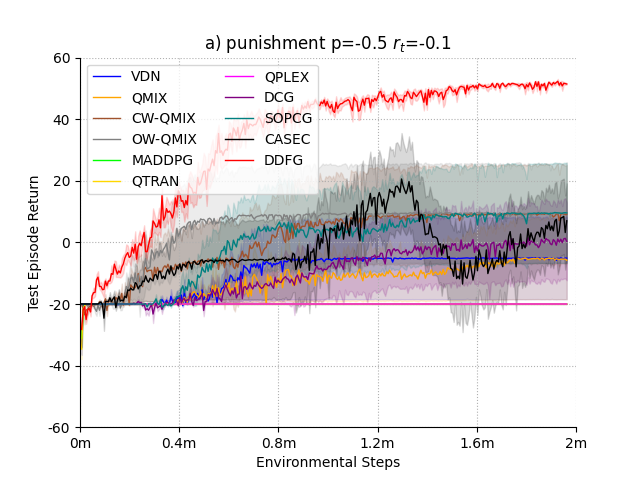}
        \end{minipage}
  }%
 \subfloat{
	\begin{minipage}[t]{0.5\linewidth}
		\centering
		\includegraphics[width=3.3in,height=3in]{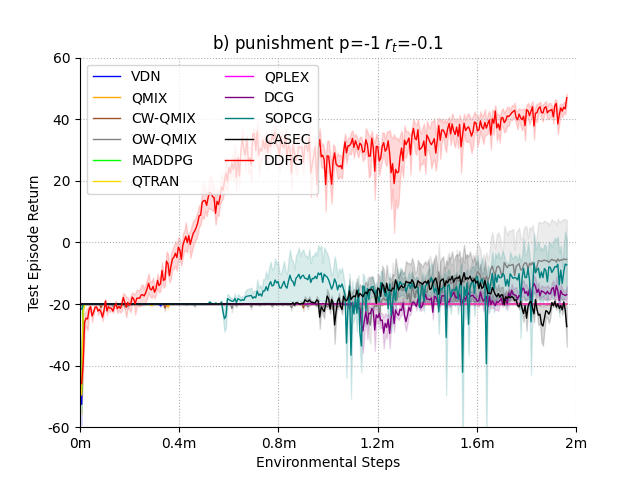}
	\end{minipage}
  }
  \caption{Median test return for the Higher-order Predator-Prey task with ${r}_{t}$, comparing DDFG and baselines.}
  \label{fig:5}
\end{figure*}

Further experimentation, incorporating a time-step penalty ${r}_{t}=-0.1$ while maintaining constant penalty values p (-0.5, -1), is presented in Figure \ref{fig:5}. The imposition of a temporal penalty ${r}_{t}$, equating to a negative reward at each time step when agents remain stationary, ostensibly complicates convergence by augmenting negative rewards. Nonetheless, this approach simultaneously fosters exploration. In scenarios with $p = -0.5$, the inclusion of ${r}_{t}$ enabled algorithms such as VDN, QMIX, CW-QMIX/OW-QMIX, SOPCG, and CASEC to enhance exploration, thereby overcoming relative overgeneralization to achieve optimal solution convergence in certain instances. Yet, in a majority of cases, these algorithms still faced outright failure. With $p=-1$, the augmentation of ${r}_{t}$ failed to facilitate success for these algorithms. Particularly for SOPCG and CASEC, the substantial increase in ${r}_{t}$ markedly impeded algorithmic success probability. The dynamic sparse coordination graph structure exhibited no discernible advantage in the HO-Predator-Prey context, with CASEC's performance even trailing behind DCG. For DCG, the negative temporal reward ${r}_{t}$ adversely affected convergence, rendering outcomes in scenarios with $p=-1$ less favorable compared to those without ${r}_{t}$. Conversely, DDFG consistently converged to the optimal policy irrespective of ${r}_{t}$'s presence, though the introduction of ${r}_{t}$ inevitably slowed DDFG's convergence by accruing more negative rewards.

\begin{figure*}
  \setlength{\abovecaptionskip}{0cm}
  \setlength{\belowcaptionskip}{-0.5cm}
  \centering
  \includegraphics[scale=0.45]{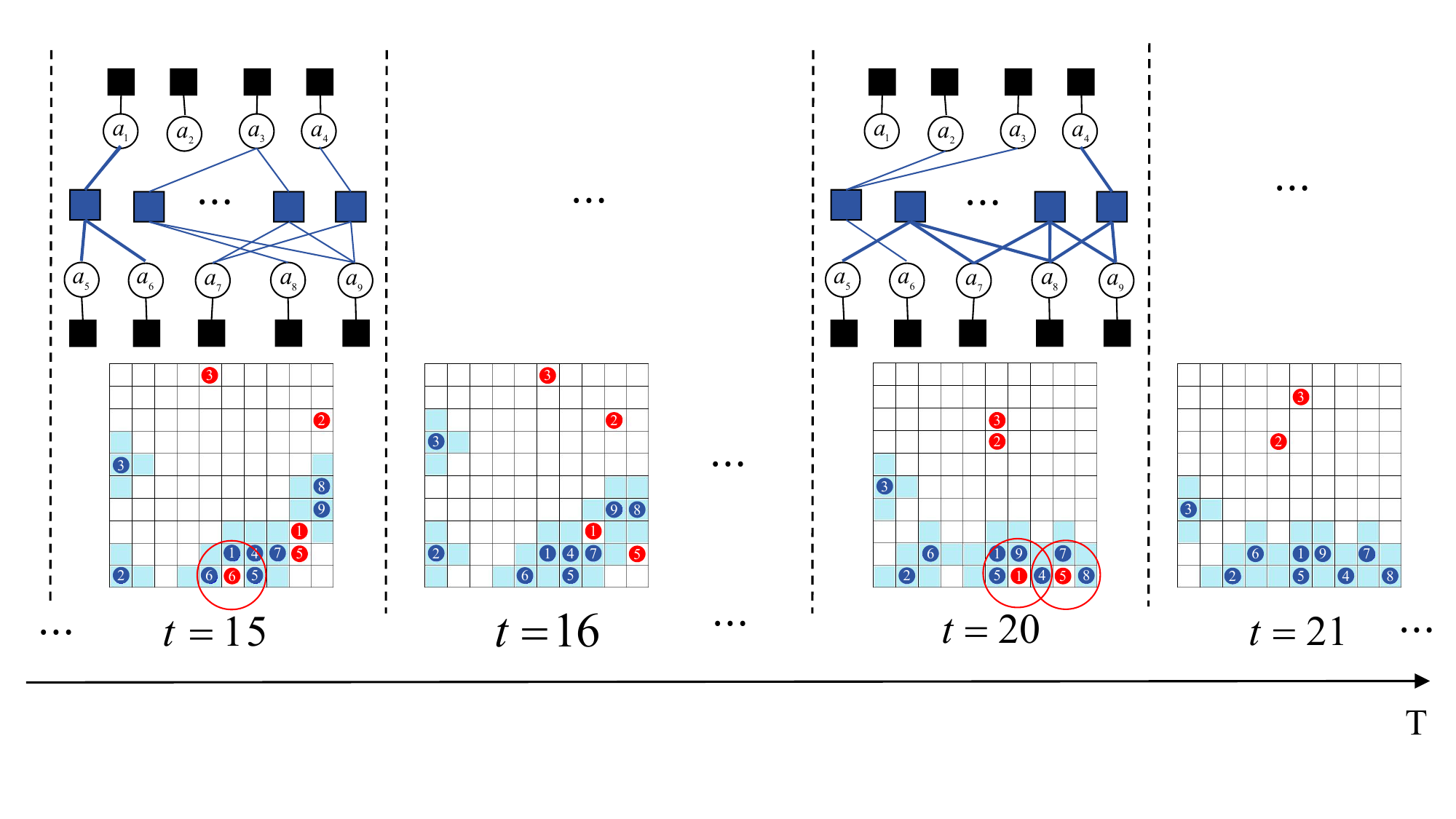}
  \caption{The dynamic changes in the graph structure of DDFG in the Higher-order Predator-Prey task (performing the "capture" action).}
  \label{fig:dym}
\end{figure*}

Finally, this paper visualizes the dynamic structure of factor graphs in HO-Predator-Prey tasks, as shown in Figure \ref{fig:dym}. We tested the DDFG after two million steps of training in an experimental scenario with a penalty of $p=-1.5$ and $r=0$. In an episode shown in Figure \ref{fig:dym}, 9 agents successfully captured 6 prey within 50 time steps (ending 150 time steps earlier). This paper selects scenarios where agents cooperate to capture prey at two moments, $t=15$ and $t=20$, and visualizes the factor graph structure at these two moments.

At $t=15$, agents 1, 4, 5, and 6 collectively encompass prey 6 and successfully captured it. The actions executed at $t=15$ reveal that agents 1, 5, and 6 initiated the 'capture' action leading to the successful capture of prey 6. Insight from the factor graph structure at $t=15$ demonstrates that, the dynamic graph policy generated joint value function nodes for agents 1, 5, and 6, managing to orchestrate their capture actions effectively.

At $t=20$, agents 1, 4, 5, and 9 encircle prey 1, while agents 4, 7, and 8 concurrently surround prey 5, leading to the capture of both prey. In reference to the list of actions executed by the agents, agents 4, 5, 7, 8, and 9 are all noted for executing the 'capture' action. This means that agents 4, 5, and 9 are responsible for capturing prey 1, whereas agents 4, 7, 8 are accountable for the capture of prey 5. Notably, agent 1 is not involved in the capture, while agent 4 contributes to both capture actions. This pattern is reflected in the factor graph structure at time $t=20$. This reflects that the graph structure generated by the dynamic graph policy will directly guide the coordinated actions between agents. Moreover, even though the dynamic graph policy does not successfully generate joint nodes for agents 4, 5, 9 or 4, 7, 8, the capture action is still successfully completed. This highlights the significance of the message passing algorithm. Starting from the factor nodes, Eq.(\ref{eq:8}) and (\ref{eq:7}) are alternately called to propagate and aggregate the value function information between the factor nodes and variable nodes, which ultimately leads to the successful capture of two preys by five agents.

In summary, the disparity in the factor graph structures between $t=15$ and $t=20$ as indicated in Figure \ref{fig:dym} underscores the ability of the dynamic graph policy introduced in this paper to generate factor graph structures that adapt to the real-time changes. Meanwhile, the above discussion confirms that the dynamic graph policy has the ability to generate real-time factor graph structures, thereby guiding multi-agent collaboration. Additionally, the scenario at $t=20$ highlights that even without the generation of an 'absolutely correct' joint value function node, the message passing algorithm can ensure successful coordinated actions among agents, hence proving the resilience of the dynamic graph policy.

\begin{figure*}[t]\centering
 %  \subfloat[]{\includegraphics[width=3.3in]{figures/3s5z.png}}%
 %  %
 % \subfloat[]{\includegraphics[width=3.3in]{figures/5m_vs_6m.png}}%
 %
  \subfloat{
	\begin{minipage}[t]{0.33\linewidth}
		\centering
		\includegraphics[width=2.4in]{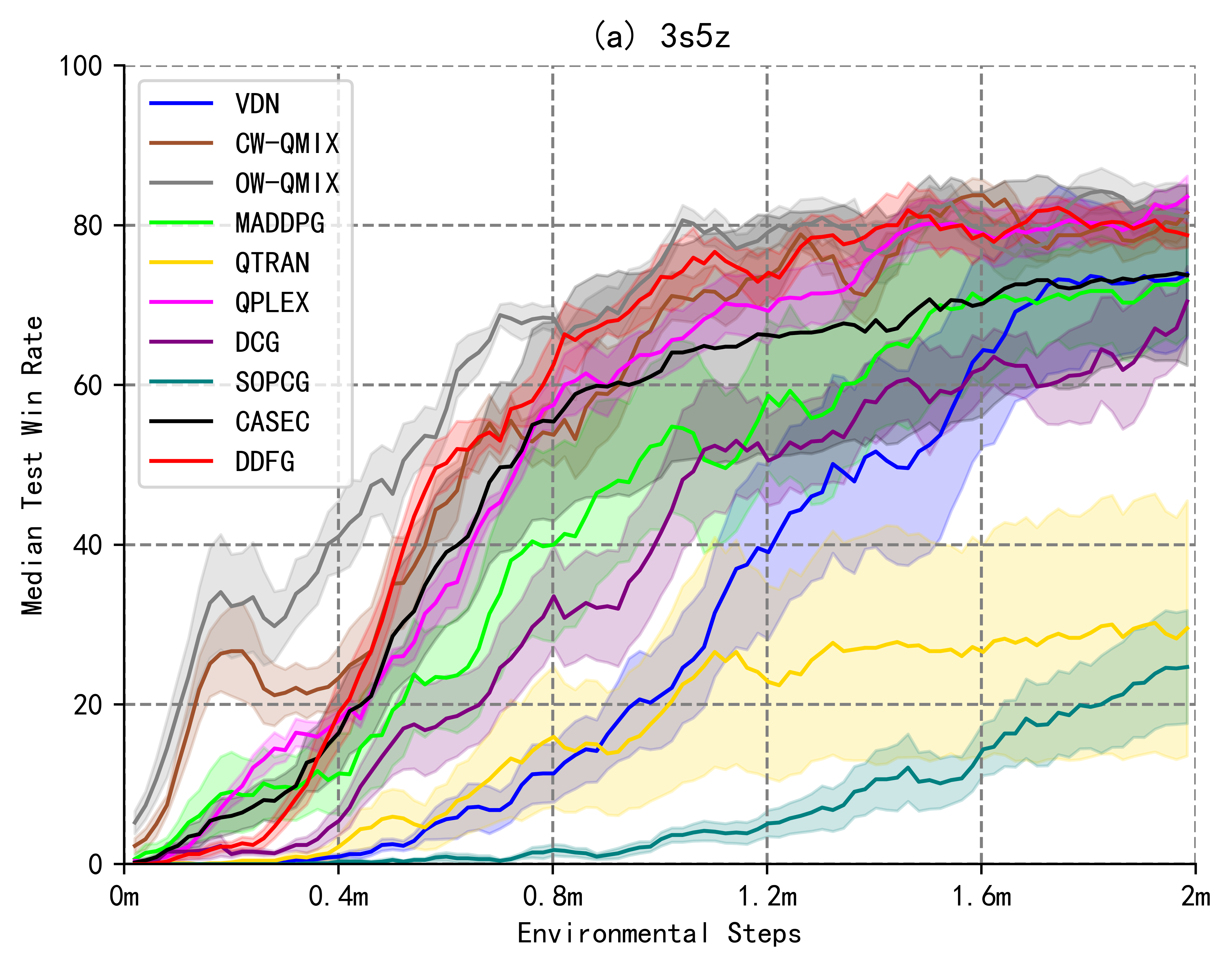}
        \end{minipage}
  }%
 \subfloat{
	\begin{minipage}[t]{0.33\linewidth}
		\centering
		\includegraphics[width=2.4in]{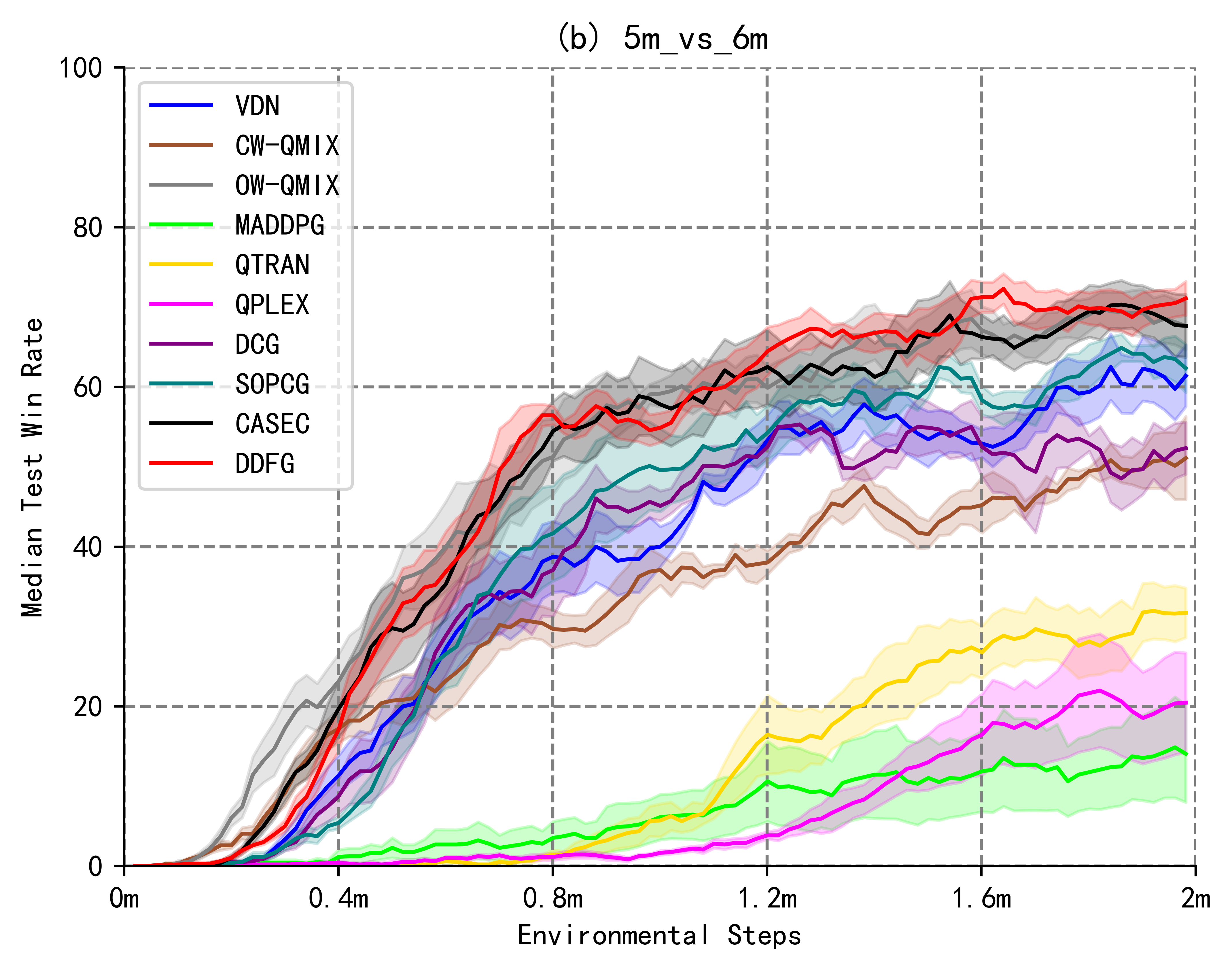}
	\end{minipage}
  }%
 \subfloat{
	\begin{minipage}[t]{0.33\linewidth}
		\centering
		\includegraphics[width=2.4in]{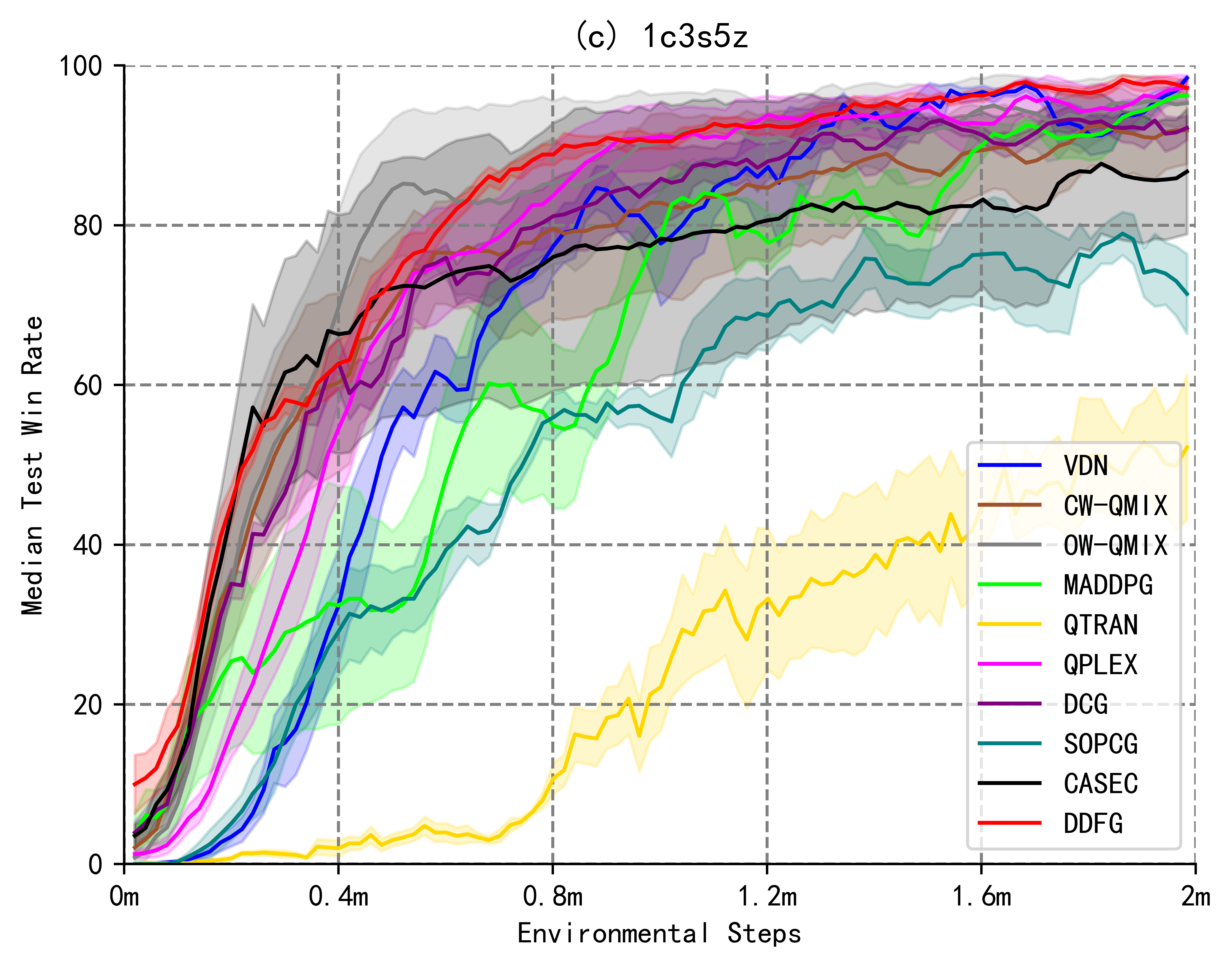}
        \end{minipage}
  }%

 \subfloat{
	\begin{minipage}[t]{0.33\linewidth}
		\centering
		\includegraphics[width=2.4in]{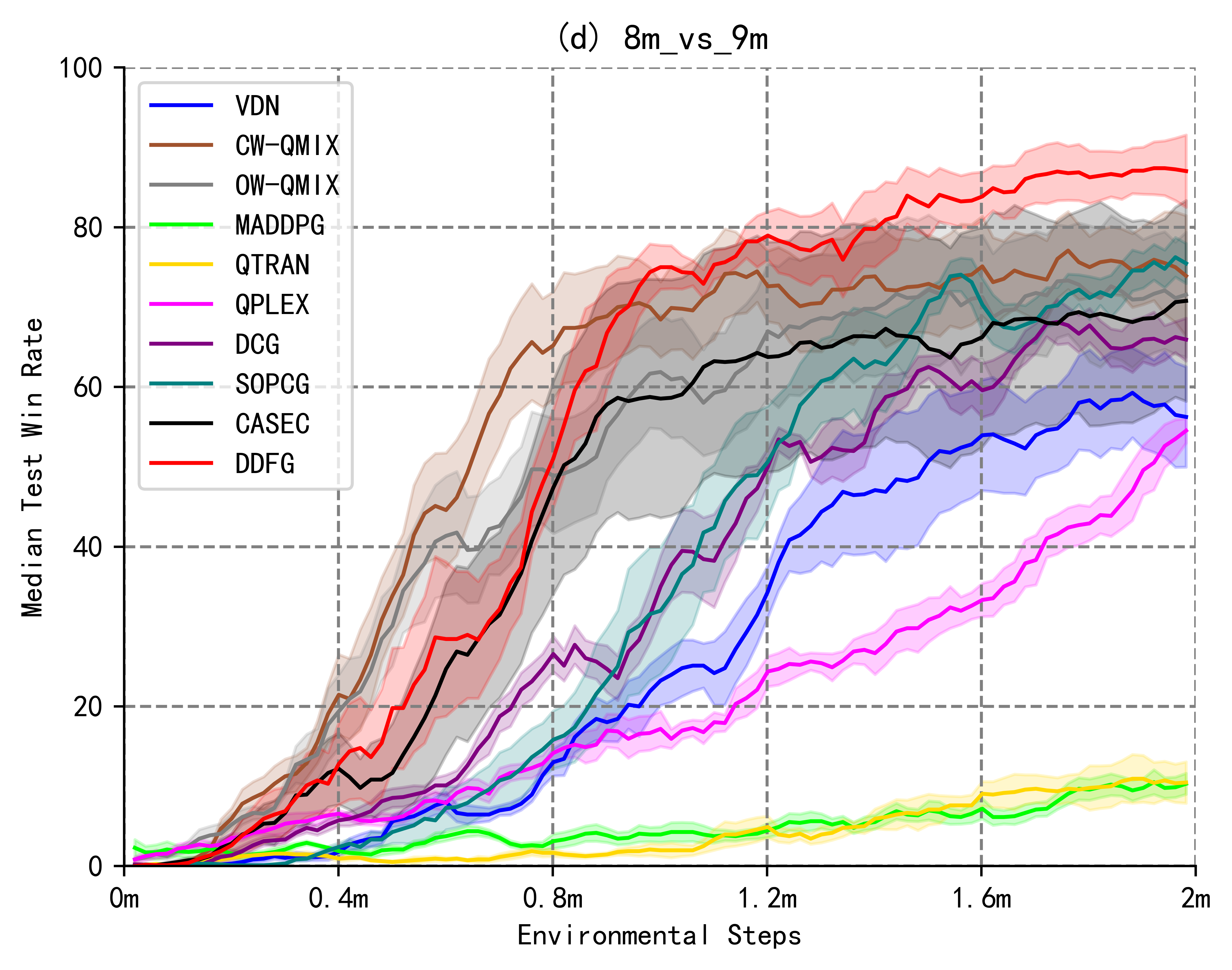}
	\end{minipage}
  }%
  \subfloat{
	\begin{minipage}[t]{0.33\linewidth}
		\centering
		\includegraphics[width=2.4in]{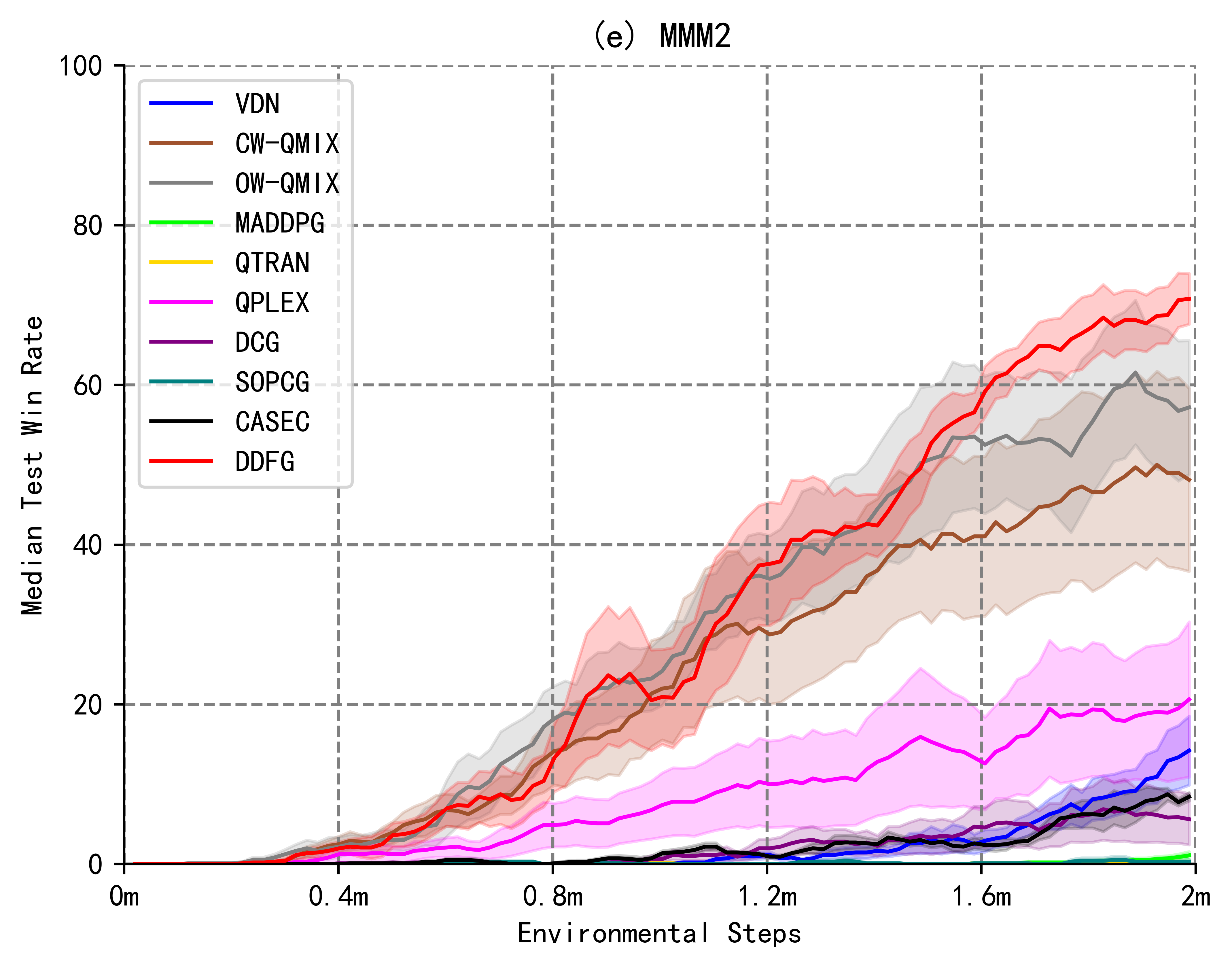}
	\end{minipage}
  }

 %  \subfloat{
	% \begin{minipage}[t]{0.5\linewidth}
	% 	\includegraphics[width=2in]{figures/MMM2.png}
	% \end{minipage}
 %  }
  %\includegraphics[scale=0.4]{figure4.1.pdf}
  \caption{Median test win rate $\%$ for the SMAC task, comparing DDFG and baselines.}
  \label{fig:6}
\end{figure*}

\subsection{SMAC}
Transitioning to the SMAC domain, we confront a series of more daunting experiments within the StarCraft Multi-Intelligent Challenge Benchmark, evaluating empirical performance across various environments through exploration and FIFO buffers. For our purposes, we have selected a total of five maps:: (a) 3s5z (Easy), (b) $5m \_ vs \_ 6m$ (Hard), (c) 1c3s5z (hard), (d) $8m\_vs\_9m$(hard), (e) MMM2 (superhard), with the opponent AI set to hard in both instances.

The experimental results, showcased in Figure \ref{fig:6}, indicate that in the 3s5z and 1c3s5z scenarios, DDFG's performance aligns with that of the leading OW-QMIX/CW-QMIX algorithms, surpassing all other baselines. In the $5m\_vs\_6m$ map, DDFG matches the performance of OW-QMIX and CASEC, outperforming the remainder of the baselines. In contrast, on the $8m\_vs\_9m$ and MMM2 maps, DDFG demonstrates superior performance relative to all baseline models. Initial learning velocity for DDFG is slower due to two primary factors: the SMAC scenarios’ immunity to relative overgeneralization, preventing baseline algorithms from stagnating at local optima, and the temporal requirements for DDFG’s graph policy to identify the optimal structural decomposition. This temporal lag results in DDFG exhibiting a slower initial learning curve when compared to some baselines during the experiment's initial phase.

The experiments validate DDFG's scalability and efficacy in complex tasks like SMAC, even in the absence of relative overgeneralization. Notably, DDFG demonstrates superior performance to DCG in all scenarios by generating dynamic graph structures that facilitate adaptive agent collaboration against adversaries. DDFG's ability to control varied agents for joint decision-making against enemies secures a higher victory rate. SOPCG and CASEC exhibit diminished success rates on the MMM2 map, while maintaining average performance on the $8m\_vs\_9m$ scenario. This discrepancy underscores the variability and stability concerns associated with dynamically sparse graph structures in fluctuating contexts, as evidenced by the divergent success rates across different maps. DDFG's employment of high-order value function networks and the max-plus algorithm for global policy optimization culminates in elevated success rates across varied SMAC maps.

\begin{figure*}
  \setlength{\abovecaptionskip}{0cm}
  \setlength{\belowcaptionskip}{-0.5cm}
  \centering
  \includegraphics[scale=0.45]{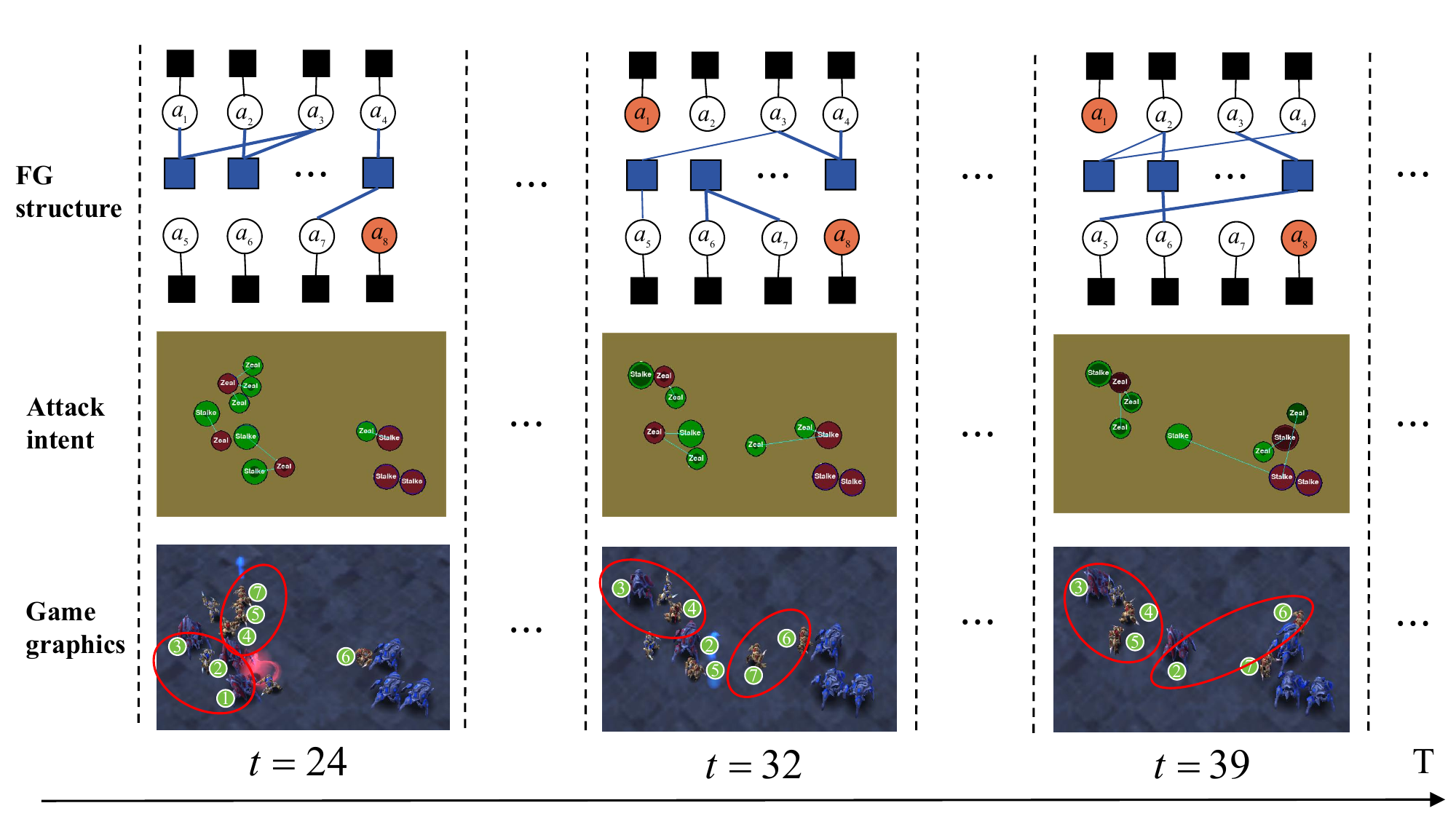}
  \caption{In the StarCraft II task (3s5z map), the dynamic changes in the graph structure of DDFG. The first row shows the dynamic FG structure at different time moments t, where the red agent nodes represent dead agents. The second row indicates the intent to attack from our agents, with the green nodes being our agents and the red ones being the enemies. The third row shows the corresponding game screen, where the red circle marks collaboration between agents, and the green numbers are agent IDs (corresponding to the FG structure in the first row).}
  \label{fig:dym_smac}
\end{figure*}

Figure \ref{fig:dym_smac} presents the visualization of the factor-graph structure in the StarCraft II experiment. This paper uses the 3s5z scenario and tests the DDFG after two million steps of training. In an episode displayed in the figure, our team controls 3 Stalkers and 5 Zealots against the enemy's identical setup of 3 Stalkers and 5 Zealots and eventually secures the victory.

At $t=24$, agents 4, 5, and 7 collaboratively attack a Zealot. The graph-structured policy successfully learns the cooperation between 4 and 7, and manipulates 5 through separate factor nodes to accomplish the collective cooperation of the three agents. At the same time, the graph-structured policy constructs cooperation between agents 1, 3 and 2, 3. The message passing is conducted through the maximum sum algorithm on the factor graph, ultimately achieving the collaborative attack on the enemy by agents 1 and 3.

At $t=32$, agents 3, 4 collaboratively attack a Stalker, while agents 6 and 7 jointly attack a  Zealot. The graph-structured policy successfully learns both cooperations simultaneously. At $t=39$, agents 2 and 6 attack a Stalker, while agents 3, 4, and 5 collaborate to attack a Zealot. The graph-structured policy learns the cooperation of 2 and 6. As for the collaboration of agents 3, 4, and 5, the policy only learns the cooperation between 3 and 5, but also manipulates 4 through a separate factor node to ultimately complete the attack.

In summary, Figure \ref{fig:dym_smac} displays different factor graph structures at three different moments, validating the effectiveness of the graph-structured policy. In most situations, the graph-structured policy can directly create corresponding factor nodes to accomplish collaborative attacks. In other cases, the policy can pass messages among different agents by executing the maximum sum algorithm on the factor graph. 

\begin{figure*}
  \setlength{\abovecaptionskip}{0cm}
  \setlength{\belowcaptionskip}{-0.5cm}
  \centering
  \subfloat{
	\begin{minipage}[t]{0.5\linewidth}
		\centering
		\includegraphics[width=2.3in,height=2in]{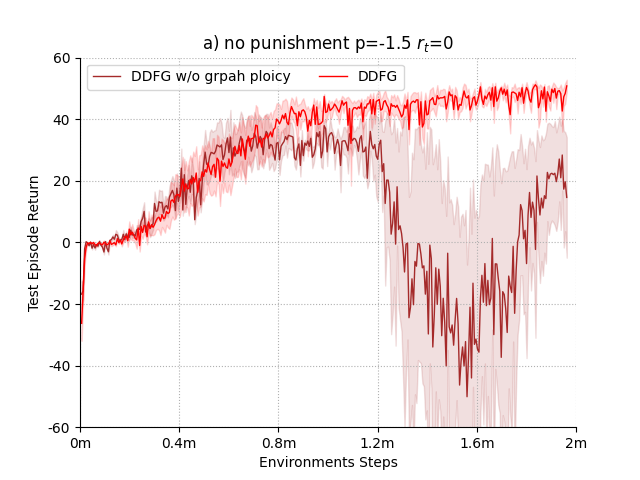}
        \end{minipage}
  }%
 \subfloat{
	\begin{minipage}[t]{0.5\linewidth}
		\centering
		\includegraphics[width=2.3in,height=2in]{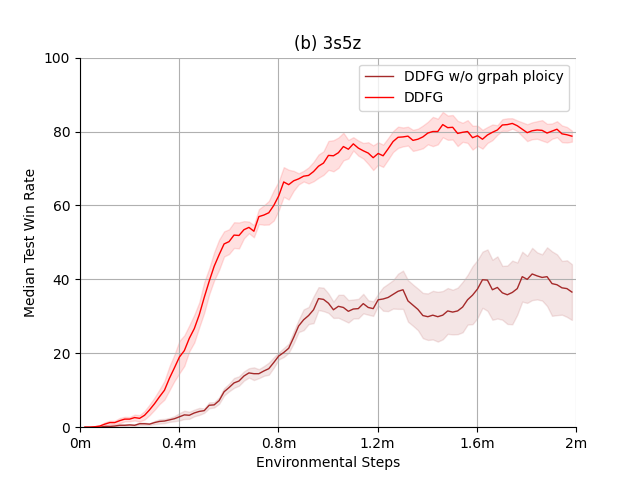}
	\end{minipage}
  }
  \caption{The ablation experiment of graph generation policy in HO-Predator-Prey and SMAC environments.}
  \label{fig:9}
\end{figure*}

\subsection{Ablation study}
The Dynamic Deep Factor Graph (DDFG) algorithm is distinguished by its integration of a value function module and a graph generation policy. As delineated in appendix B, a notable inference is that DDFG, when constrained such that every pair of agents is connected to a factor node, effectively reduces to the Deep Coordination Graph (DCG) framework. This scenario can be interpreted as DDFG operating under a static graph structure. DCG employs a fully connected graph structure, and as evidenced in Figures \ref{fig:4}, \ref{fig:5}, and \ref{fig:6}, it achieves a measurable degree of success/win rate across various scenarios. This observation underscores the intrinsic value of the function module within the algorithm. This section aims to elucidate the contributions of graph generation policies through the deployment of ablation studies.

First, we initiate a comparison between dynamic and fixed graph policies. Observations from Figures \ref{fig:4}, \ref{fig:5}, and \ref{fig:6} reveal that DDFG consistently outperforms DCG across all tested environments. This disparity suggests that static graph strategies are inadequate in encapsulating the dynamic collaborative relationships among agents within the environment. Conversely, the graph generation policy introduced in this work adeptly captures these collaborative dynamics, thereby enhancing overall algorithmic performance. Moreover, DDFG consistently demonstrates a superior learning velocity compared to DCG, implying that the iterative alternation between updating the dynamic graph policy and the value function network contributes to expedited learning processes.

Second, the study contrasts the dynamic graph policy with a random graph policy, with results depicted in Figure \ref{fig:9}. The random graph policy is characterized by a uniform probability distribution governing the connectivity between the value function and agents, effectively equating the local value function to a uniform connection probability among agents. Selecting specific scenarios/maps within the HO-Predator-Prey and SMAC frameworks for experimentation, Figure \ref{fig:9} delineates a pronounced divergence in performance between the dynamic and random graph policies. This divergence validates the dynamic graph policy's capability to adaptively learn the evolution of collaborative relationships among agents across different temporal junctures, a feat unachievable by the random graph policy. Furthermore, in the context of HO-Predator-Prey experiments, the application of a random graph policy within the DDFG framework engendered a negative feedback loop between the learning mechanisms of the random graph and the value function. This phenomenon highlights the detrimental effects of a random graph policy on the value function network within challenging environments, thereby emphasizing the criticality of accurately learning collaborative relationships.

\section{Conclusions$\And $Future Work}

This paper has introduced the Dynamic Deep Factor Graph (DDFG) algorithm, leveraging factor graph decomposition for global value functions, significantly enhancing the algorithm's capability to counteract relative overgeneralization. Employing tensor-based CP decomposition, DDFG adeptly navigates large action spaces with heightened efficiency. Furthermore, through the development of a graph structure generation policy utilizing a hypernetwork, DDFG dynamically generates adjacency matrices, capturing the fluid collaborative dynamics among agents. Employing the max-sum algorithm, the optimal policy for agents is ascertained. DDFG's effectiveness is empirically validated in complex higher-order predator-prey tasks and numerous challenging scenarios within the SMAC II framework. Compared to DCG, DDFG strikes a more refined balance between centralized and decentralized methodologies, explicitly learning dynamic agent collaboration and offering a more adaptable decomposition policy. Future endeavors will explore the elimination of the highest order limitation within DDFG to enhance the decomposition method's flexibility while preserving algorithmic learning efficiency.

% \section{Acknowledgements}
% This work is supported in part by the National Natural Science Foundation of China under Grant 62101029, in part by Guangdong Basic and Applied Basic Research Foundation under Grant 2023A1515140071, and in part by the China Scholarship Council Award under Grant 202006465043 and 202306460078.

% \section{Author Contributions}
% Yuchen Shi and Shihong Duan’s main contribution is Methodology, Writing- Original draft preparation. Cheng Xu’s main contribution is Conceptualization, Methodology, Writing- Reviewing, Editing and Supervision. Ran Wang and Fangwen Ye's main contribution is Investigation and Validation. Chau Yuen’s main contribution is Writing- Reviewing.

% \section{Competing Interests}
% All authors declare no financial or non-financial competing interests.

% \section{Data Availability}
% The data that support the findings of this study are available at \url{https://github.com/SICC-Group/DDFG}, and also from the corresponding author upon reasonable request.

% \section{Code Availability}
% The code that supports the findings of this study is available The source code is available at \url{https://github.com/SICC-Group/DDFG}. Issues and technique supports are available from the corresponding author upon reasonable request.

% \begin{appendices}
% \end{appendices}

% \newpage
% %%%%%%%%%%%%%%%%%%%%%%%%%%%%%%%%%%%%%%%%%%%%%%%%%%%%%%%%%%%%

% \appendix

% {\Large \centering \bf{Supplementary materials for “ Dynamic Deep Factor Graph for Multi-agent Reinforcement Learning”}}

\bibliographystyle{IEEEtran}
\bibliography{references}

\begin{thebibliography}{10}
\expandafter\ifx\csname url\endcsname\relax
  \def\url#1{\texttt{#1}}\fi
\expandafter\ifx\csname urlprefix\endcsname\relax\def\urlprefix{URL }\fi
\expandafter\ifx\csname href\endcsname\relax
  \def\href#1#2{#2} \def\path#1{#1}\fi

\bibitem{cao2023continuous}
Z.~Cao, K.~Jiang, W.~Zhou, S.~Xu, H.~Peng, D.~Yang, Continuous improvement of self-driving cars using dynamic confidence-aware reinforcement learning, Nature Machine Intelligence 5~(2) (2023) 145--158.

\bibitem{wang2024}
H.~Wang, J.~Wang, Enhancing multi-uav air combat decision making via hierarchical reinforcement learning, Scientific Reports 14~(1) (2024) 4458.

\bibitem{oroojlooy2022review}
A.~Oroojlooy, D.~Hajinezhad, A review of cooperative multi-agent deep reinforcement learning, Applied Intelligence (2022) 1--46.

\bibitem{lowe2017}
R.~Lowe, Y.~I. Wu, A.~Tamar, J.~Harb, O.~Pieter~Abbeel, I.~Mordatch, Multi-agent actor-critic for mixed cooperative-competitive environments, Advances in neural information processing systems 30.

\bibitem{foerster2018}
J.~Foerster, G.~Farquhar, T.~Afouras, N.~Nardelli, S.~Whiteson, Counterfactual multi-agent policy gradients, in: Proceedings of the AAAI conference on artificial intelligence, Vol.~32, 2018.

\bibitem{vdn}
P.~Sunehag, G.~Lever, A.~Gruslys, W.~M. Czarnecki, V.~Zambaldi, M.~Jaderberg, M.~Lanctot, N.~Sonnerat, J.~Z. Leibo, K.~Tuyls, et~al., Value-decomposition networks for cooperative multi-agent learning based on team reward, in: Proceedings of the 17th International Conference on Autonomous Agents and MultiAgent Systems, 2018, pp. 2085--2087.

\bibitem{qmix}
T.~Rashid, M.~Samvelyan, C.~S. De~Witt, G.~Farquhar, J.~Foerster, S.~Whiteson, Monotonic value function factorisation for deep multi-agent reinforcement learning, The Journal of Machine Learning Research 21~(1) (2020) 7234--7284.

\bibitem{panait2006}
L.~Panait, S.~Luke, R.~P. Wiegand, Biasing coevolutionary search for optimal multiagent behaviors, IEEE Transactions on Evolutionary Computation 10~(6) (2006) 629--645.

\bibitem{qtran}
K.~Son, D.~Kim, W.~J. Kang, D.~E. Hostallero, Y.~Yi, Qtran: Learning to factorize with transformation for cooperative multi-agent reinforcement learning, in: International conference on machine learning, PMLR, 2019, pp. 5887--5896.

\bibitem{qplex}
J.~Wang, Z.~Ren, T.~Liu, Y.~Yu, C.~Zhang, Qplex: Duplex dueling multi-agent q-learning, in: International Conference on Learning Representations, 2020.

\bibitem{wqmix}
T.~Rashid, G.~Farquhar, B.~Peng, S.~Whiteson, Weighted qmix: Expanding monotonic value function factorisation for deep multi-agent reinforcement learning, Advances in neural information processing systems 33 (2020) 10199--10210.

\bibitem{guestrin2002}
C.~Guestrin, M.~Lagoudakis, R.~Parr, Coordinated reinforcement learning, in: ICML, Vol.~2, Citeseer, 2002, pp. 227--234.

\bibitem{dcg}
W.~B{\"o}hmer, V.~Kurin, S.~Whiteson, Deep coordination graphs, in: International Conference on Machine Learning, PMLR, 2020, pp. 980--991.

\bibitem{li2020}
S.~Li, J.~K. Gupta, P.~Morales, R.~Allen, M.~J. Kochenderfer, Deep implicit coordination graphs for multi-agent reinforcement learning, in: Proceedings of the 20th International Conference on Autonomous Agents and MultiAgent Systems, 2021, pp. 764--772.

\bibitem{kang2022}
Y.~Kang, T.~Wang, Q.~Yang, X.~Wu, C.~Zhang, Non-linear coordination graphs, Advances in Neural Information Processing Systems 35 (2022) 25655--25666.

\bibitem{sopcg}
Q.~Yang, W.~Dong, Z.~Ren, J.~Wang, T.~Wang, C.~Zhang, Self-organized polynomial-time coordination graphs, in: International Conference on Machine Learning, PMLR, 2022, pp. 24963--24979.

\bibitem{casec}
T.~Wang, L.~Zeng, W.~Dong, Q.~Yang, Y.~Yu, C.~Zhang, Context-aware sparse deep coordination graphs, in: International Conference on Learning Representations, 2021.

\bibitem{loeliger2004}
H.-A. Loeliger, An introduction to factor graphs, IEEE Signal Processing Magazine 21~(1) (2004) 28--41.

\bibitem{schulman2017}
J.~Schulman, F.~Wolski, P.~Dhariwal, A.~Radford, O.~Klimov, Proximal policy optimization algorithms, arXiv preprint arXiv:1707.06347.

\bibitem{kschischang2001}
F.~R. Kschischang, B.~J. Frey, H.-A. Loeliger, Factor graphs and the sum-product algorithm, IEEE Transactions on information theory 47~(2) (2001) 498--519.

\bibitem{phan2021}
A.-H. Phan, P.~Tichavsk{\`y}, K.~Sobolev, K.~Sozykin, D.~Ermilov, A.~Cichocki, Canonical polyadic tensor decomposition with low-rank factor matrices, in: ICASSP 2021-2021 IEEE International Conference on Acoustics, Speech and Signal Processing (ICASSP), IEEE, 2021, pp. 4690--4694.

\bibitem{zhou2019}
M.~Zhou, Y.~Chen, Y.~Wen, Y.~Yang, Y.~Su, W.~Zhang, D.~Zhang, J.~Wang, Factorized q-learning for large-scale multi-agent systems, in: Proceedings of the first international conference on distributed artificial intelligence, 2019, pp. 1--7.

\bibitem{oliehoek2016}
F.~A. Oliehoek, C.~Amato, A concise introduction to decentralized POMDPs, Springer, 2016.

\bibitem{mnih2015}
V.~Mnih, K.~Kavukcuoglu, D.~Silver, A.~A. Rusu, J.~Veness, M.~G. Bellemare, A.~Graves, M.~Riedmiller, A.~K. Fidjeland, G.~Ostrovski, et~al., Human-level control through deep reinforcement learning, nature 518~(7540) (2015) 529--533.

\bibitem{van2016}
H.~Van~Hasselt, A.~Guez, D.~Silver, Deep reinforcement learning with double q-learning, in: Proceedings of the AAAI conference on artificial intelligence, Vol.~30, 2016.

\bibitem{hausknecht2015}
M.~Hausknecht, P.~Stone, Deep recurrent q-learning for partially observable mdps, in: 2015 aaai fall symposium series, 2015.

\bibitem{zhang2014}
Z.~Zhang, D.~Zhao, Clique-based cooperative multiagent reinforcement learning using factor graphs, IEEE/CAA Journal of Automatica Sinica 1~(3) (2014) 248--256.

\bibitem{kok2006}
J.~R. Kok, N.~Vlassis, Using the max-plus algorithm for multiagent decision making in coordination graphs, in: RoboCup 2005: Robot Soccer World Cup IX 9, Springer, 2006, pp. 1--12.

\bibitem{lan2006}
X.~Lan, S.~Roth, D.~Huttenlocher, M.~J. Black, Efficient belief propagation with learned higher-order markov random fields, in: Computer Vision--ECCV 2006: 9th European Conference on Computer Vision, Graz, Austria, May 7-13, 2006. Proceedings, Part II 9, Springer, 2006, pp. 269--282.

\bibitem{som2010}
P.~Som, A.~Chockalingam, Damped belief propagation based near-optimal equalization of severely delay-spread uwb mimo-isi channels, in: 2010 IEEE International Conference on Communications, IEEE, 2010, pp. 1--5.

\bibitem{foerster2017}
J.~Foerster, N.~Nardelli, G.~Farquhar, T.~Afouras, P.~H. Torr, P.~Kohli, S.~Whiteson, Stabilising experience replay for deep multi-agent reinforcement learning, in: International conference on machine learning, PMLR, 2017, pp. 1146--1155.

\bibitem{schulman2015}
J.~Schulman, P.~Moritz, S.~Levine, M.~Jordan, P.~Abbeel, High-dimensional continuous control using generalized advantage estimation, arXiv preprint arXiv:1506.02438.

\bibitem{samvelyan2019}
M.~Samvelyan, T.~Rashid, C.~Schroeder~de Witt, G.~Farquhar, N.~Nardelli, T.~G. Rudner, C.-M. Hung, P.~H. Torr, J.~Foerster, S.~Whiteson, The starcraft multi-agent challenge, in: Proceedings of the 18th International Conference on Autonomous Agents and MultiAgent Systems, 2019, pp. 2186--2188.

\end{thebibliography}


\begin{thebibliography}{1}
\expandafter\ifx\csname url\endcsname\relax
  \def\url#1{\texttt{#1}}\fi
\expandafter\ifx\csname urlprefix\endcsname\relax\def\urlprefix{URL }\fi
\expandafter\ifx\csname href\endcsname\relax
  \def\href#1#2{#2} \def\path#1{#1}\fi

\bibitem{kraemer2016}
L.~Kraemer, B.~Banerjee, Multi-agent reinforcement learning as a rehearsal for decentralized planning, Neurocomputing 190 (2016) 82--94.

\bibitem{qtran}
K.~Son, D.~Kim, W.~J. Kang, D.~E. Hostallero, Y.~Yi, Qtran: Learning to factorize with transformation for cooperative multi-agent reinforcement learning, in: International conference on machine learning, PMLR, 2019, pp. 5887--5896.

\bibitem{duelingdqn}
Z.~Wang, T.~Schaul, M.~Hessel, H.~Hasselt, M.~Lanctot, N.~Freitas, Dueling network architectures for deep reinforcement learning, in: International conference on machine learning, PMLR, 2016, pp. 1995--2003.

\bibitem{wang2020rode}
T.~Wang, T.~Gupta, A.~Mahajan, B.~Peng, S.~Whiteson, C.~Zhang, Rode: Learning roles to decompose multi-agent tasks, in: International Conference on Learning Representations, 2020.

\end{thebibliography}

\end{document}

\title{Dynamic Deep Factor Graph for Multi-Agent Reinforcement Learning\\
--- Supplementary Material ---
}

\author{Yuchen Shi, Shihong~Duan, Cheng~Xu,~\IEEEmembership{Member,~IEEE,} Ran Wang,~\IEEEmembership{Graduate Student Member,~IEEE,} Fangwen~Ye,  Chau~Yuen,~\IEEEmembership{Fellow,~IEEE}
        % <-this % stops a space
\thanks{
  This work was supported in part by the National Natural Science Foundation of China (NSFC) under Grant 62101029, Guangdong Basic and Applied Basic Research Foundation under Grant 2023A1515140071, and in part by the China Scholarship Council Award under Grant 202006465043 and 202306460078. (\textit{Corresponding author:} Cheng Xu)
  }
\IEEEcompsocitemizethanks
{\IEEEcompsocthanksitem Yuchen Shi, Shihong Duan and Fangwen Ye are with School of Computer and Communication Engineering, Shunde Innovation School, University of Science and Technology Beijing (email: shiyuchen199@sina.com; duansh@ustb.edu.cn; yfwen2000@outlook.com).

\IEEEcompsocthanksitem Cheng Xu is with School of Computer and Communication Engineering, Shunde Innovation School, University of Science and Technology Beijing. He is also with School of Electrical and Electronic Engineering, Nanyang Technological University (email: xucheng@ustb.edu.cn).

\IEEEcompsocthanksitem Ran Wang is with School of Computer and Communication Engineering, Shunde Innovation School, University of Science and Technology Beijing. She is also with School of Computer Science and Engineering, Nanyang Technological University (email: wangran423@foxmail.com).

\IEEEcompsocthanksitem Chau Yuen is with School of Electrical and Electronic Engineering, Nanyang Technological University. (email: chau.yuen@ntu.edu.sg).
}% <-this % stops an unwanted space
\thanks{Digital Object Identifier 10.1109/TPAMI.2024.XXXXXXX}
}

\onecolumn

\markboth{Journal of \LaTeX\ Class Files,~Vol.~X, No.~X, Month~Year}%
{Shell \MakeLowercase{\textit{et al.}}: Bare Demo of IEEEtran.cls for Journals}

\maketitle

In this supplementary material, we give additional related work discussion, detailed derivations and proof of the equations and propositions mentioned in the main text. Supplementary experiment settings and the code reproduction steps are also presented.

\appendices

\section{Related Work}
\subsection{VDN}
The main assumption made and utilized by the VDN is that the joint action-value function of the system can be decomposed into a sum of the value functions of the single agent:
\begin{equation}
Q(({{h}^{1}},{{h}^{2}},\ldots ,{{h}^{N}}),({{u}^{1}},{{u}^{2}},\ldots ,{{u}^{N}}))\approx \sum\limits_{i=1}^{N}{{{{\tilde{Q}}}_{i}}({{h}^{i}},{{u}^{i}})}
\end{equation}
where ${{{\tilde{Q}}}_{i}}$ depends only on the local observation of each agent. We learn ${{\tilde{Q}}_{i}}$ by back-propagating the rule in Q-learning, which learns the joint reward by summation. The ${{\tilde{Q}}_{i}}$ is learned implicitly, not from any reward specific to agent i. The algorithm does not constrain ${{\tilde{Q}}_{i}}$ to be an action-value function for any particular reward. The agents in the VDN algorithm can be deployed independently because the greedy policy of each agent ${{u}^{i}}=\arg {{\max }_{{{u}^{i}}}} {{{\tilde{Q}}}_{i}}({{h}^{i}},{{u}^{i}})$ for the local value function ${{\tilde{Q}}_{i}}$ is equivalent to choosing the joint action to maximize $\sum\limits_{i=1}^{N}{{{\tilde{Q}}}_{i}}$.

\subsection{QMIX}
QMIX implements two improvements over VDN: 1) the inclusion of global information to assist in the training process and 2) the use of a hybrid network to merge local value functions of single agents.

QMIX, as a representative value decomposition method, follows the centralized training, distributed execution (CTDE) paradigm\cite{kraemer2016}. The core part of QMIX is the hybrid network, which is responsible for credit assignments. In QMIX, each agent a has a separate Q-network ${{Q}_{i}}({{\tau }^{i}},{{u}^{i}})$. The output of the single Q-network is passed through the hybrid network, which implicitly assigns credits to each agent and generates an approximation of the global Q-value ${{Q}_{tot}}(\bm{\tau} ,\bm{u},s;\theta )$. The ${{Q}_{tot}}$ is computed as follows:
\begin{equation}
    {{Q}_{tot}}(\bm{\tau} ,\bm{u},s;\theta )={{W}_{2}}ELU\left( {{W}_{1}}\bm{Q}+{{b}_{1}} \right)+{{b}_{2}}
\end{equation}
where $\bm{Q}={{\left[ {{Q}_{1}}({{\tau }_{1}},{{u}_{1}}),{{Q}_{2}}({{\tau }_{2}},{{u}_{2}}),\ldots ,{{Q}_{n}}({{\tau }_{n}},{{u}_{n}}) \right]}^{T}}\in {{\mathbb{R}}^{n\times 1}}$ is the output of the independent Q-network. ${{W}_{1}}\in \mathbb{R}_{+}^{m\times n},{{W}_{2}}\in \mathbb{R}_{+}^{1\times m},{{b}_{1}}\in \mathbb{R}_{{}}^{m\times 1},{{b}_{2}}\in \mathbb{R}$ is the weight generated by the HyperNetworks. Since the elements of ${{W}_{1}}$ and ${{W}_{2}}$ are non-negative, QMIX satisfies the following condition:
\begin{equation}
    \frac{\partial {{Q}_{tot}}}{\partial {{Q}_{i}}}\ge 0,\forall i\in \{1,2,\cdots,n\}
\end{equation}

This property guarantees the individual-global maximum (IGM) principle\cite{qtran}, i.e., that the optimal action of each agent is jointly constituted as an optimal joint action of ${{Q}_{tot}}$:
\begin{equation}
\underset{\bm{u}}{\mathop{\arg \max }}\,{{Q}_{tot}}(\bm{\tau} ,\bm{u},s)=\left( 
\begin{aligned}
  & \underset{{{u}^{1}}}{\mathop{\arg \max }}\,{{Q}_{1}}({{\tau }^{1}},{{u}^{1}}) \\ 
 & \quad \quad \quad \ \,\vdots  \\ 
 & \underset{{{u}^{n}}}{\mathop{\arg \max }}\,{{Q}_{n}}({{\tau }^{n}},{{u}^{n}}) \\ 
\end{aligned} 
\right)
\end{equation}
In this way, the agent can choose the best local action based only on its local observation-action history, while the joint action is the best action for the whole system.

The system is updated by minimizing the square of the TD loss on ${{Q}_{tot}}$, according to the following equation:
\begin{equation}
    \mathcal{L}(\theta )={{({{y}^{dqn}}-{{Q}_{tot}}(\bm{\tau} ,\bm{u},s;\theta ))}^{2}}
\end{equation}
where ${{y}^{dqn}}=r+\gamma {{\max }_{{\bm{u}'}}}{{Q}_{tot}}(\bm{\tau }',\bm{u}',{s}';{{\theta }^{-}})$ is the TD objective.

\subsection{WQMIX}
Although QMIX improves the form of value function decomposition of VDN, the monotonicity constraint of QMIX makes its function expressiveness somewhat limited.WQMIX improves the value function expressiveness of QMIX by defining the QMIX operator and weighting the QMIX operator according to the importance of each action in the joint action space, which in turn maps it to the "non-monotonic" functions.

First, WQMIX defines the function space where the global value function ${{Q}_{tot}}$ resides:
\begin{equation}
    {{Q}^{mix}}=\{{{Q}_{tot}}\mid{{Q}_{tot}}(s,\bm{u})={{f}_{s}}({{Q}_{1}}(s,{{u}^{1}}),\ldots ,{{Q}_{n}}(s,{{u}^{n}})),\frac{\partial {{f}_{s}}}{\partial {{Q}_{i}}}\ge 0,{{Q}_{i}}(s,{{u}^{i}})\in \mathbb{R}\}
\end{equation}

Constraining ${{Q}^{mix}}$ in ${{Q}_{tot}}$ can be seen as solving the following optimization problem:
\begin{equation}
    \arg {{\min }_{q\in {{Q}^{mix}}}}{{\sum\nolimits_{\bm{u}\in U}{({{\mathcal{T}}^{*}}{{Q}_{tot}}(s,\bm{u})-q(s,\bm{u}))}}^{2}},\forall s\in S
\end{equation}
where ${{\mathcal{T}}^{*}}$ is the Bellman optimality operator.

Meanwhile, define the corresponding projection operator ${{\Pi }_{Qmix}}$:
\begin{equation}
  {{\Pi }_{Qmix}}Q:=\arg {{\min }_{q\in {{Q}^{mix}}}}\sum\nolimits_{\bm{u}\in U}{{{(Q(s,\bm{u})-q(s,\bm{u}))}^{2}}}
\end{equation}

Define the QMIX operator as a composite of the Bellman optimality operator and the projection operator: $\mathcal{T}_{Qmix}^{*}={{\Pi }_{Qmix}}{{\mathcal{T}}^{*}}$.

The QMIX operator $\mathcal{T}_{Qmix}^{*}$ itself has many problems, such as $\mathcal{T}_{Qmix}^{*}$ is not a compression mapping, and the joint actions obtained by maximizing ${{Q}_{tot}}$ in QMIX are erroneous in some scenarios (underestimating the value of some joint actions).Based on these drawbacks, WQMIX proposes the Weighted QMIX operator.

First, define the new projection operator ${{\Pi }_{w}}$:
\begin{equation}
    {{\Pi }_{w}}Q:=\arg {{\min }_{q\in {{Q}^{mix}}}}\sum\nolimits_{\bm{u}\in U}{w(s,\bm{u}){{(Q(s,\bm{u})-q(s,\bm{u}))}^{2}}}
\end{equation}
where the weight function $w:S\times U\to (0,1]$, when $w(s,\bm{u})\equiv 1$, ${{\Pi }_{w}}={{\Pi }_{Qmix}}$.

And the weight calculation needs to use ${{Q}^{*}}$, so it is necessary to additionally learn ${{{\hat{Q}}}^{*}}$ to fit ${{Q}^{*}}$, and ${{Q}^{*}}$ is updated by the following operator:
\begin{equation}
    \mathcal{T}_{w}^{*}{{\hat{Q}}^{*}}(s,u):=E[r+\gamma {{\hat{Q}}^{*}}({s}',\arg {{\max }_{{{u}'}}}{{Q}_{tot}}({s}',{u}'))]
\end{equation}

Then the weighted QMIX operator is defined as: $\mathcal{T}_{WQMIX}^{*}{{\hat{Q}}^{*}}:={{\Pi }_{w}}\mathcal{T}_{w}^{*}{{\hat{Q}}^{*}}$.

Eventually WQMIX trains both ${{Q}_{tot}}$ and ${{\hat{Q}}^{*}}$, updating the network by minimizing the square of the TD loss of the two Q functions:
\begin{equation}
    \mathcal{L}(\theta )=w(s,\bm{u}){{({{y}^{dqn}}-{{Q}_{tot}}(\bm{\tau} ,\bm{u},s;{{\theta }_{1}}))}^{2}}+{{({{y}^{dqn}}-{{\hat{Q}}^{*}}(\bm{\tau} ,\bm{u},s;{{\theta }_{2}}))}^{2}}
\end{equation}
where ${{y}^{dqn}}=r+\gamma {{\hat{Q}}^{*}}(\bm{\tau }',\arg {{\max }_{{\bm{u}'}}}{{Q}_{tot}}(\bm{\tau }',\bm{u}',{s}';{{\theta }_{1}}),{s}';{{\theta }_{2}})$ is the TD target.

And WQMIX obtains two WQMIX algorithms, Centrally-Weighted QMIX(CW-QMIX) and Optimistically-Weighted QMIX(OW-QMIX), by designing different weight functions $w(s,\bm{u})$.

Weighting functions in CW-QMIX:
\begin{equation}
    w(s,u)=\left\{ \begin{aligned}
    & 1\quad {{y}^{dqn}}>{{{\hat{Q}}}^{*}}(\bm{\tau} ,{{\bm{\hat{u}}}^{*}},s)\ or\ \bm{u}={{\bm{\hat{u}}}^{*}} \\ 
    & \alpha \quad otherwise \\ 
    \end{aligned} \right.
\end{equation}
where ${\bm{\hat{u}}^{*}}=\arg {{\max }_{\bm{u}}}{{Q}_{tot}}(\bm{\tau} ,\bm{u},s)$.

Weighting functions in OW-QMIX:
\begin{equation}
w(s,\bm{u})=\left\{ \begin{aligned}
  & 1\quad {{Q}_{tot}}(\bm{\tau} ,\bm{u},s)<{{y}^{dqn}} \\ 
 & \alpha \quad otherwise \\ 
\end{aligned} \right.
\end{equation}

\subsection{QTRAN}
QTRAN improves the decomposition form of the value function in VDN, QMIX. QTRAN proposes the IGM condition and proves that the linear decomposition in VDN and the monotonic decomposition in QMIX are sufficient but not necessary for the IGM. This leads to the decomposition of value functions in VDN and QMIX in a way that is too constrained, making it impossible to solve non-monotonicity tasks. QTRAN proposes to satisfy the sufficient and necessary conditions for IGM and proves its correctness.

Theorem proposed by QTRAN:
A global value function ${{Q}_{tot}}\left( \bm{\tau} ,\bm{u} \right)$ can be decomposed into $\left[ {{Q}_{i}}\left( {{\tau }^{i}},{{u}^{i}} \right) \right]_{i=1}^{N}$, if it satisfies:
\begin{equation}
\sum\limits_{i=1}^{N}{{{Q}_{i}}\left( {{\tau }^{i}},{{u}^{i}} \right)-{{Q}_{tot}}\left( \bm{\tau} ,\bm{u} \right)+{{V}_{tot}}\left( \bm{\tau}  \right)}=\left\{ \begin{aligned}
  & 0\quad \bm{u}=\bm{\bar{u}}\  \\ 
 & \ge 0\ \bm{u}\ne \bm{\bar{u}} \\ 
\end{aligned} \right.
\end{equation}
where ${{V}_{tot}}\left( \bm{\tau}  \right)=\underset{\bm{u}}{\mathop{\max }}\,{{Q}_{tot}}\left( \bm{\tau} ,\bm{u} \right)-\sum\limits_{i=1}^{N}{{{Q}_{i}}\left( {{\tau }^{i}},{{{\bar{u}}}^{i}} \right)}$.

QTRAN learns ${{Q}_{i}},{{Q}_{tot}},{{V}_{tot}}$ through the network separately and designs the loss function based on ${{Q}_{tot}},{{V}_{tot}}$:
\begin{equation}
    \mathcal{L}\left( \bm{\tau} ,\bm{u},r,\bm{\tau }';\theta  \right)={{\mathcal{L}}_{td}}+{{\lambda }_{opt}}{{\mathcal{L}}_{opt}}+{{\lambda }_{nopt}}{{\mathcal{L}}_{nopt}}
\end{equation}

${{\mathcal{L}}_{td}}$ is used to fit ${{Q}_{tot}}$, ${{\mathcal{L}}_{opt}}$ and ${{\mathcal{L}}_{nopt}}$ is used to fit ${{V}_{tot}}$. ${\lambda }_{opt}$ and ${\lambda }_{nopt}$ are the weight constants for two losses. ${{{\mathcal{L}}}_{td}},{{\mathcal{L}}_{opt}},{{\mathcal{L}}_{nopt}}$ are defined as follows:
\begin{equation}
\begin{aligned}
  & {{\mathcal{L}}_{td}}\left( ;\theta  \right)={{({{y}^{dqn}}-{{Q}_{tot}}(\bm{\tau} ,\bm{u}))}^{2}},{{y}^{dqn}}=r+\gamma {{Q}_{tot}}(\bm{\tau }',\bm{\bar{u}}';{{\theta }^{\_}}) \\ 
 & {{\mathcal{L}}_{opt}}\left( ;\theta  \right)={{({{{{Q}'}}_{tot}}(\bm{\tau} ,\bm{\bar{u}})-{{{\hat{Q}}}_{tot}}(\bm{\tau} ,\bm{\bar{u}})+{{V}_{tot}}\left( \bm{\tau}  \right))}^{2}} \\ 
 & {{\mathcal{L}}_{nopt}}\left( ;\theta  \right)={{(\min \left[ {{{{Q}'}}_{tot}}(\bm{\tau} ,\bm{u})-{{{\hat{Q}}}_{tot}}(\bm{\tau} ,\bm{u})+{{V}_{tot}}\left( \bm{\tau}  \right),0 \right])}^{2}} \\ 
\end{aligned}
\end{equation}
where ${{\theta }^{\_}}$ is the target network parameter. ${{\mathcal{L}}_{opt}}$ and ${{\mathcal{L}}_{nopt}}$ are used to ensure that the above theorem holds.

\subsection{QPLEX}
QPLEX proposes an improvement to the IGM. By drawing on the f dueling
decomposition structure $Q=V+A$ proposed by Dueling DQN\cite{duelingdqn}, QPLEX formalizes the IGM as an advantage-based IGM. 

Advantage-based IGM: For the joint action-value function ${{Q}_{tot}}:\mathcal{T}\times U\to \mathbb{R}$ and the individual action-value function $[{{Q}_{i}}:\mathcal{T}\times U\to \mathbb{R}]_{i=1}^{n}$, where $\forall \tau \in \mathcal{T},\forall u\in U,\forall i\in N$,

\quad \textbf{Joint Dueling} \quad ${{Q}_{tot}}(\bm{\tau} ,\bm{u})={{V}_{tot}}(\bm{\tau} )+{{\mathcal{A}}_{tot}}(\bm{\tau} ,\bm{u})\ and\ {{V}_{tot}}(\bm{\tau} )=\underset{{{u}'}}{\mathop{\max }}\,{{Q}_{tot}}(\bm{\tau} ,\bm{u}')$

\quad \textbf{Individual Dueling} \quad ${{Q}_{i}}({{\tau }^{i}},{{u}^{i}})={{V}_{i}}({{\tau }^{i}})+{{\mathcal{A}}_{i}}({{\tau }^{i}},{{u}^{i}})\ and\ {{V}_{tot}}({{\tau }^{i}})=\underset{{{u}^{i}}^{\prime }}{\mathop{\max }}\,{{Q}_{i}}({{\tau }^{i}},{{u}^{i}}^{\prime })$

such that the following holds:
\begin{equation}
\underset{\bm{u}\in U}{\mathop{\arg \max }}\,{{\mathcal{A}}_{tot}}(\bm{\tau} ,\bm{u})\ =\left( \underset{{{u}^{1}}\in U}{\mathop{\arg \max }}\,{{\mathcal{A}}_{i}}({{\tau }^{1}},{{u}^{1}}),\ldots ,\underset{{{u}^{n}}\in U}{\mathop{\arg \max }}\,{{\mathcal{A}}_{n}}({{\tau }^{n}},{{u}^{n}}) \right)
\end{equation}

Then, $[{{Q}_{i}}]_{i=1}^{n}$ satisfies the Advantage-based IGM of ${{Q}_{tot}}$ (which is equivalent to the IGM).

The structure of QPLEX consists of two parts:  (i) an Individual Action-Value Function for each agent, and (ii) a Duplex Dueling
component.

An Individual Action-Value Function consists of an RNN Q-network for each agent, where the network inputs are the hidden state $h_{t-1}^{i}$ and action $u_{t-1}^{i}$ of the previous moment, and the local observation $o_{t}^{i}$ of the current moment. The output is a localized value function ${{Q}_{i}}({{\tau }^{i}},{{u}^{i}})$.

The Duplex Dueling component connects local and joint action-valued functions through two modules: (i) a Transformation network module (ii) a Dueling Mixing network module.

The Transformation network module uses centralized information to transform the local duel structure $[{{V}_{i}}({{\tau }^{i}}),{{\mathcal{A}}_{i}}({{\tau }^{i}},{{u}^{i}})]_{i=1}^{n}$ to $[{{V}_{i}}(\bm{\tau} ),{{\mathcal{A}}_{i}}(\bm{\tau} ,{{u}^{i}})]_{i=1}^{n}$ for any agent i, i.e., ${{Q}_{i}}(\bm{\tau} ,{{u}^{i}})={{w}_{i}}(\bm{\tau} ){{Q}_{i}}({{\tau }^{i}},{{u}^{i}})+{{b}_{i}}(\bm{\tau} )$, thus:
\begin{equation}
{{V}_{i}}(\bm{\tau} )={{w}_{i}}(\bm{\tau} ){{V}_{i}}({{\tau }^{i}})+{{b}_{i}}(\bm{\tau} )\quad and\quad {{\mathcal{A}}_{i}}(\bm{\tau} ,{{u}^{i}})={{w}_{i}}(\bm{\tau} ){{\mathcal{A}}_{i}}({{\tau }^{i}},{{u}^{i}})
\end{equation}
where ${{w}_{i}}(\bm{\tau} )>0$, this positive linear transformation maintains the consistency of greedy action selection.

The Dueling Mixing network module takes as input the output of the Transformation network module $[{{V}_{i}},{{\mathcal{A}}_{i}}]_{i=1}^{n}$ and outputs the global value function ${{Q}_{tot}}$.

Advantage-based IGM conditions do not impose constraints on ${{V}_{tot}}$, so QPLEX uses a simple sum structure for ${{V}_{tot}}$: ${{V}_{tot}}(\tau )=\sum\limits_{i=1}^{n}{{{V}_{i}}(\tau )}$.

To satisfy the Advantage-based IGM condition, QPLEX calculates the joint advantage function as follows:
\begin{equation}
    {{\mathcal{A}}_{tot}}(\bm{\tau} ,\bm{u})=\sum\limits_{i=1}^{n}{{{\lambda }_{i}}(\bm{\tau} ,\bm{u})}{{\mathcal{A}}_{i}}(\bm{\tau} ,{{u}^{i}}),\ where\ {{\lambda }_{i}}(\bm{\tau} ,\bm{u})>0
\end{equation}

The positive linear transformation guided by ${{\lambda }_{i}}$ will continue to maintain consistency in greedy action selection. To be able to efficiently learn ${{\lambda }_{i}}$ with joint histories and actions, QPLEX uses an extensible multi-head attention module:
\begin{equation}
  {{\lambda }_{i}}(\bm{\tau} ,\bm{u})=\sum\limits_{k=1}^{K}{{{\lambda }_{i,k}}}(\bm{\tau} ,\bm{u}){{\phi }_{i,k}}(\bm{\tau} ){{v}_{k}}(\bm{\tau} )
\end{equation}
where K is the number of attention heads, ${{\lambda }_{i,k}}(\bm{\tau} ,\bm{u})$ and ${{\phi }_{i,k}}(\bm{\tau} )$ are the attention weights activated by the sigmoid function, and ${{v}_{k}}(\bm{\tau} )>0$ is the positive key for each head.

Eventually, the joint action-value function ${{Q}_{tot}}$ is expressed as follows:
\begin{equation}
    {{Q}_{tot}}(\bm{\tau} ,\bm{u})={{V}_{tot}}(\bm{\tau} )+{{\mathcal{A}}_{tot}}(\bm{\tau} ,\bm{u})=\sum\limits_{i=1}^{n}{{{Q}_{i}}(\bm{\tau} ,{{u}^{i}})}+\sum\limits_{i=1}^{n}{({{\lambda }_{i}}(\bm{\tau} ,\bm{u})-1)}{{\mathcal{A}}_{i}}(\bm{\tau} ,{{u}^{i}})
\end{equation}

\subsection{DCG}
DCG is a MARL algorithm that combines coordination graphs and deep neural networks. In DCG, a coordination graph $CG$ decomposes the joint value function into the sum of the utility function $f^i$ and the cost function $f^{ij}$ as follows:
\begin{equation}
Q^{CG}({s_t},\bm{u})=Q^{CG}({\bm{\tau}_t},\bm{u})=\frac{1}{\lvert V \rvert}\sum\limits_{{v^i}\in V}{f^i({u^i}\mid{\bm{h}_t})+\frac{1}{\lvert \varepsilon \rvert}\sum\limits_{{i,j}\in \varepsilon}{f^{ij}({u^i},{u^j}\mid{\bm{h}_t})}}
\end{equation}

DCG uses the max-plus algorithm of the coordination graph to deliver messages. At each time step $t$, every node $i$ sends a message $\mu_{t}^{ij}({{u}^{j}})\in \mathbb{R}$ to all neighboring nodes ${i,j}\in \varepsilon$. The message can be decentralized and computed using the following equation:

\begin{equation}
\mu_{t}^{ij}({{u}^{j}}) \leftarrow \underset{{{a}^{i}}}{\mathop{\max }}\left\{\frac{1}{\lvert V \rvert}{{f}^{i}}({{u}^{i}}\mid{\bm{h}_{t}})+\frac{1}{\lvert \varepsilon \rvert}{{f}^{ij}}({{u}^{i}},{{u}^{j}}\mid{\bm{h}_{t}})+\sum\limits_{{k,i}\in \varepsilon }{\mu _{t}^{ki}({{u}^{i}})}-\mu _{t}^{ji}({{u}^{i}}) \right\}
\end{equation}

This message computation and passing must be repeated several times until convergence so that each agent $i$ can find the optimal action $u_{}^{i}$ to obtain the maximized estimated joint action value ${{Q}^{CG}}({\bm{\tau }_{t}},{\bm{u}_{*}})$. The optimal action $u_{*}^{i}$ can be computed using the following equation:

\begin{equation}
u_{}^{i}:=\underset{{{u}^{i}}}{\mathop{\arg \max }}\left\{\frac{1}{\lvert V \rvert}{{f}^{i}}({{u}^{i}}\mid{\bm{h}_{t}})+\sum\limits_{{k,i}\in \varepsilon }{\mu _{t}^{ki}({{u}^{i}})} \right\}
\end{equation}

At each time step $t$, DCG approximates the utility function ${{f}^{i}}$ and the cost function ${{f}^{ij}}$ using a network parameterized by $\theta$, $\phi$, and $\psi$. The joint action value is denoted as follows:

\begin{equation}
Q_{\theta \phi \psi }^{DCG}({\bm{\tau }_{t}},\bm{u}):=\frac{1}{\lvert V \rvert}f_{\theta }^{v}({{u}^{i}}\mid{h}_{t}^{i})+\frac{1}{2 \lvert \varepsilon \rvert}\sum\limits_{{i,j}\in \varepsilon }(f_{\phi }^{e}({{u}^{i}},{{u}^{j}}\mid{h}_{t}^{i},h_{t}^{j})+f_{\phi }^{e}({{u}^{j}},{{u}^{i}}\mid{h}_{t}^{j},h_{t}^{i}))
\end{equation}

Finally, the DCG updates the parameters of the DCG network end-to-end according to the DQN-style loss function:
\begin{equation}
    {{\mathcal{L}}_{DQN}}:=E\left[ \frac{1}{T}\sum\limits_{t=0}^{T-1}{{{\left( Q_{\theta \phi \psi }^{DCG}({\bm{u}_{t}}\mid{\bm{\tau }_{t}})-\left( {{r}_{t}}+\gamma \max Q_{\bar{\theta }\bar{\phi }\bar{\psi }}^{DCG}(\cdot \mid{\bm{\tau }_{t+1}}) \right) \right)}^{2}}} \right]
\end{equation}
where $\bar{\theta },\bar{\phi },\bar{\psi }$ is the parameter copied periodically from $\theta ,\phi ,\psi$.

\subsection{SOPCG}
SOPCG uses a class of polynomial-time coordination graphs to construct a dynamic and state-dependent topology. SOPCG first models coordination relations upon the graph-based value factorization specified by DCG, where the joint value function of the multi-agent system is factorized into the summation of individual utility functions ${f_{i}}$ and pairwise payoff functions ${f_{ij}}$.
\begin{equation}
    Q({{\tau }_{t}},u;G)=\sum\limits_{i\in [n]}{{{f}^{i}}({{u}^{i}}\mid\tau _{t}^{i})+}\sum\limits_{(i,j)\in G}{{{f}^{ij}}({{u}^{i}},{{u}^{j}}\mid\tau _{t}^{i},\tau _{t}^{j})}
\end{equation}
where the coordination graph $G$ is represented by a set of undirected edges. With this second-order value decomposition, the hardness of greedy action selection is highly related to the graph topology.

Polynomial-Time Coordination Graph Class: A graph class $G$ is a Polynomial-Time Coordination Graph Class if there exists an algorithm that can solve any induced DCOP of any coordination graph $G\in \mathcal{G}$ in $Poly(n,A)$ running time.

The set of undirected acyclic graphs, denoted as ${{G}{uac}}$, forms a polynomial-time tractable graph class. However, within an environment containing $n$ agents, an undirected acyclic graph can have at most $n-1$ edges, limiting the expressive capacity of functions. To mitigate this issue, SPOCG permits the dynamic evolution of the graph's topology through transitions in the environmental state. Across different environmental states, the joint value can be decomposed by selecting coordinating graphs from a predefined graph class $G\subseteq {{G}{uac}}$. This graph class choice augments the limited representational capacity while preserving the accuracy of greedy action selection. 

SOPCG introduce an imaginary coordinator agent whose action space refers to the selection of graph topologies, aiming to select a proper graph for minimizing the suboptimality of performance within restricted coordination. The graph topology $G$ can be regarded as an input of joint value function $Q({{\tau }_{t}},u;G)$. The objective of the coordinator agent is to maximize the joint value function over the specific graph class. SOPCG handle the imaginary coordinator as a usual agent in the multi-agent Q-learning framework and rewrite the joint action as ${\bm{u}^{cg}}=({{u}^{1}},\cdots ,{{u}^{n}},G)$

Formally, at time step t, greedy action selection indicates the following joint action:
\begin{equation}
\bm{u}_{t}^{cg}\leftarrow \underset{({{u}^{1}},\cdots ,{{u}^{n}},G)}{\mathop{\arg \max }}\,Q({\bm{\tau }_{t}},{{u}^{1}},\cdots {{u}^{n}};G)
\end{equation}

Hence the action of coordinator agent ${{G}_{t}}$ is:
\begin{equation}
{{G}_{t}}\leftarrow \underset{G\in \mathcal{G}}{\mathop{\arg \max }}\,\left( \underset{u}{\mathop{\max }}\,Q({\bm{\tau }_{t}},\bm{u};G) \right)
\end{equation}
After determining the graph topology ${{G}_{t}}$, the agents can choose their individual actions to jointly maximize the value function $Q({{\tau }_{t}},u;G)$ upon the selected topology.

With the imaginary coordinator, SOPCG can reformulate the Bellman optimality equation and maximize the future value over the coordinator agent’s action:
\begin{equation}
    {{Q}^{*}}(\tau ,u;G)=\underset{{{\tau }'}}{\mathop{E}}\,\left[ r+\underset{{{G}'}}{\mathop{\max }}\,\underset{{{u}'}}{\mathop{\max }}\,{{Q}^{*}}({\tau }',{u}';{G}') \right]
\end{equation}

Since the choice of the graph is a part of the agents' actions, $Q(\tau ,u;G)$ can be updated using temporal difference learning:
\begin{equation}
    {{\mathcal{L}}_{cg}}(\theta )=\underset{(\tau ,u,G,r,{\tau }')\sim D}{\mathop{E}}\,\left[ {{({{y}_{cg}}-Q(\tau ,u;G;\theta ))}^{2}} \right]
\end{equation}

\subsection{CASEC}
CASEC utilizes the variance of paired payoff functions as the criterion for selecting edges. A sparse graph is employed when choosing greedy joint actions for execution and updating the Q-function. Simultaneously, to mitigate the impact of estimation errors on sparse topological learning, CASEC further incorporates a network structure based on action representation for utility and payoff learning.

If the actions of agent $j$ significantly influence the expected utility for agent $i$, then agent $i$ needs to coordinate its action selection with agent $j$. For a fixed action $u^i$, $Var_{u^j}[q_{ij}(u^i, u^j \mid \tau_t^i, \tau_t^j)]$ can quantify the impact of agent $j$ on the expected payoff. This motivates the use of variance of payoff functions as the criterion for edge selection in CASEC:
\begin{equation}
    \zeta _{ij}^{{{q}_{\operatorname{var}}}}=\underset{{{u}^{i}}}{\mathop{\max }}\,Va{{r}_{{{u}^{j}}}}[{{q}_{ij}}({{u}^{i}},{{u}^{j}}\mid\tau _{t}^{i},\tau _{t}^{j})]
\end{equation}
The maximization operator guarantees that the most affected action is considered. When $\zeta _{ij}^{{{q}_{\operatorname{var}}}}$ is large, the expected utility of agent $i$ fluctuates dramatically with the action of agent $j$, and they need to coordinate their actions. Therefore, to construct sparse coordination graphs, we can set a sparseness controlling constant $\lambda \in (0,1)$ and select $\lambda \lvert \mathcal{V} \rvert(\lvert \mathcal{V} \rvert-1)$ edges with the largest $\zeta _{ij}^{{{q}_{\operatorname{var}}}}$ values.

CASEC consists of two main components–learning value functions and selecting greedy actions. 

In CASEC, agents learn a shared utility function ${{q}_{\theta }}(\cdot \mid\tau _{t}^{i})$, parameterized by $\theta$, and a shared pairwise
payoff function ${{q}_{\varphi }}(\cdot \mid\tau _{t}^{i},\tau _{t}^{j})$, parameterized by $\varphi$. The global Q value function is estimated as:
\begin{equation}
    {{Q}_{tot}}({\bm{\tau }_{t}},\bm{u})=\frac{1}{\lvert \mathcal{V} \rvert}\sum\limits_{i}{{{q}_{\theta }}({{u}^{i}}\mid\tau _{t}^{i})+}\frac{1}{\lvert \mathcal{V} \rvert(\lvert \mathcal{V} \rvert-1)}\sum\limits_{i\ne j}{{{q}_{\varphi }}({{u}^{i}},{{u}^{j}}\mid\tau _{t}^{i},\tau _{t}^{j})}
\end{equation}

which is updated by the TD loss:
\begin{equation}
    {{\mathcal{L}}_{TD}}(\theta ,\varphi )={{E}_{D}}\left[ {{(r+\gamma {{{\overset{\scriptscriptstyle\frown}{Q}}}_{tot}}(\bm{\tau }',Max-Sum({{q}_{\theta }},{{q}_{\varphi }}))-{{Q}_{tot}}(\bm{\tau} ,\bm{u}))}^{2}} \right]
\end{equation}
$Max-Sum(\cdot ,\cdot )$ is the greedy joint action selected by Max-Sum, ${{\overset{\scriptscriptstyle\frown}{Q}}_{tot}}$ is a target network with parameters $\hat{\theta },\hat{\varphi }$ periodically copied from ${{Q}_{tot}}$. Meanwhile, we also minimize a sparseness loss:
\begin{equation}
    \mathcal{L}_{sparse}^{{{q}_{\operatorname{var}}}}(\varphi )=\frac{1}{\lvert \mathcal{V} \rvert(\lvert \mathcal{V} \rvert-1)\lvert A \rvert}\sum\limits_{i\ne j}{\sum\limits_{{{u}^{i}}}{Va{{r}_{{{u}^{j}}}}[{{q}_{ij}}({{u}^{i}},{{u}^{j}}\mid\tau _{t}^{i},\tau _{t}^{j})]}}
\end{equation}
which is a regularization on $\zeta _{ij}^{{{q}_{\operatorname{var}}}}$. Introducing a scaling factor ${{\lambda }_{sparse}}\in (0,1]$ and minimizing ${{\mathcal{L}}_{TD}}(\theta ,\varphi )+{{\lambda }_{sparse}}\mathcal{L}_{sparse}^{{{q}_{\operatorname{var}}}}(\varphi )$ builds in inductive biases which favor minimized coordination graphs that would not sacrifice global returns.

In CASEC, the issue of estimation error becomes particularly pronounced as the construction of the coordination graph relies on estimates of ${{q}_{ij}}$. As the construction of the coordination graph also impacts the learning of ${{q}_{ij}}$, a negative feedback loop is generated. This loop leads to instability in learning. To address this problem and stabilize training, CASEC proposes the following strategies: 1) Periodically constructing the graph using a target payoff function to maintain a consistent graph construction approach; 2) Accelerating the training of the payoff function between updates of the target network, by learning action representations to reduce estimation error.

For 2), CASEC proposes a conditional approach on action representations to enhance sample efficiency. We train an action encoder $f_{\phi}(u)$, where the input is a one-hot encoding of an action $u$, and the output is its representation $z_a$. The action representation $z_a$, along with the current local observation, is used to predict rewards and observations in the next time step. The prediction loss is minimized to update the action encoder $f_{\phi}(u)$. At the beginning of learning, the action encoder is trained with a small number of samples \cite{wang2020rode} and remains fixed for the rest of the training process.

Using action representations, the utility and payoff functions can now be estimated as:
\begin{equation}
\begin{aligned}
  & {{q}_{\theta }}({{u}^{i}}\mid\tau _{t}^{i})={{h}_{\theta }}{{(\tau _{t}^{i})}^{T}}{{z}_{{{u}^{i}}}}; \\ 
 & {{q}_{\varphi }}({{u}^{i}},{{u}^{j}}\mid\tau _{t}^{i},\tau _{t}^{j})={{h}_{\theta }}{{(\tau _{t}^{i},\tau _{t}^{j})}^{T}}[{{z}_{{{u}^{i}}}},{{z}_{{{u}^{j}}}}] \\ 
\end{aligned}
\end{equation}

\section{Additional notes from DDFG}
\begin{figure}
  \centering
  \includegraphics[scale=0.38]{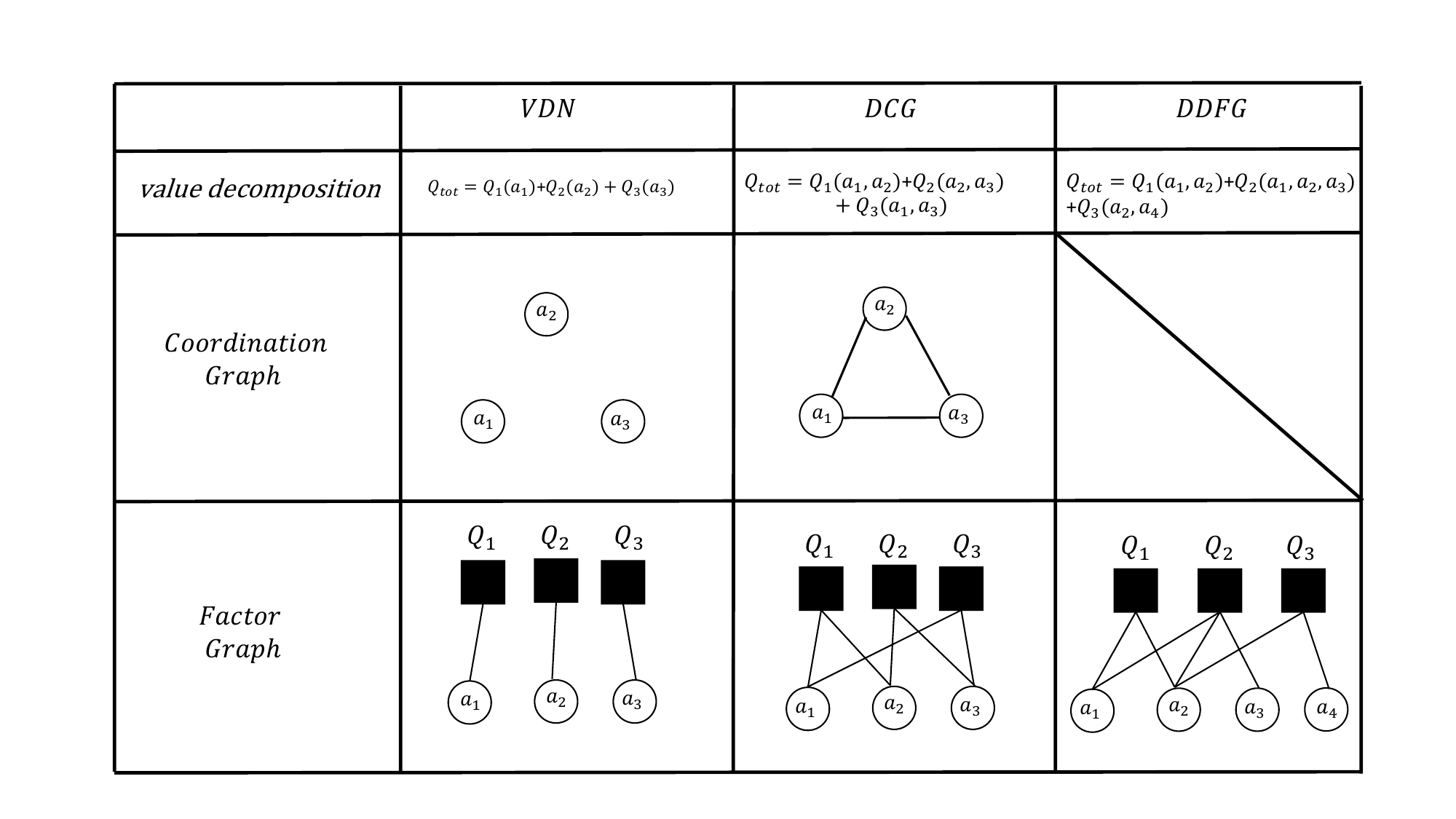}
  \caption{The graph structure representation of VDN, DCG and DDFG, using both coordination graph and factor graph.}
  \label{fig:3}
\end{figure}

As depicted in Figure \ref{fig:3}, when the set of edges $\mathcal{E}$ is empty (i.e., ${{A}_{t}}^{\prime}={{I}_{n}}$), the algorithm becomes VDN. On the other hand, when every two agents are connected to a factor node (i.e., $\mathcal{E}:=\{\{i,j\},\{i+k,j\}\mid1\le i\le n,1\le k\le n-i,j=j+1(j=1\sim n(n-1)/2)\}$), the algorithm degenerates to DCG.

In Section IV-B of the main text, we propose to construct the advantage function ${{\mathcal{A}}_{{j}}}$ by GAE instead of the value function ${{Q}_{j}}$. This is precisely because the ${{Q}_{j}}$ function itself has too much variance, which is not conducive to the learning of graph policies. And the advantage function ${{\mathcal{A}}_{{j}}}$ can better evaluate the goodness of graph policies. The design of the advantage function ${{\mathcal{A}}_{{j}}}$ is as follows:
\begin{equation}
    \hat{\mathcal{A}}_{j}^{GAE}\left( \bm{\tau}_{t}^{j},\bm{u}_{t}^{j},{{A}_{t}} \right):=\sum\limits_{t=0}^{\infty }{{{\left( \gamma {{\lambda }_{GAE}} \right)}^{l}}\left( {{Q}_{j}}\left( \bm{\tau}_{t}^{j},\bm{u}_{t}^{j},{{A}_{t}} \right)-{{V}_{j}}\left( \bm{\tau}_{t}^{j} \right) \right)}
\end{equation}
where ${{\lambda }_{GAE}}$ is the discount factor in GAE.

We construct the network of ${{V}_{j}}\left( \bm{\tau}_{t}^{j} \right)$ by a network structure similar to that of Section III-A. We also use the adjacency matrix ${{A}_{t}}$, we represent ${{V}_{j}}({\bm{\tau }_{t}},\bm{u}_{t}^{j})$ as a network parameterized by ${{\phi }_{j}}$. The joint value function ${{V}_{tot}}$, represented by the factor graph $G$:
\begin{equation}
{V}_{tot}({\bm{\tau }_{t}},{{A}_{t}})=\sum\limits_{j\in J}{{{V}_{j}}({\bm{\tau }_{t}};{{\phi }_{j}})}
\end{equation}
where $\mathcal{J}$ is the set of factor node numbers.

The network of ${{V}_{tot}}$ also share $h_{t}^{i}$ with Q-value function networks, using $h_{t}^{i}$ as an input to the network:
\begin{equation}
{V}_{tot}({\bm{\tau }_{t}},{{A}_{t}})=\sum\limits_{j\in J}{{V}_{j}}(\bm{u}_{t}^{j}\mid\bm{h}_{t}^{j};{{\phi }_{D(j)}})
\end{equation}

Unlike the Q-value function network, the network of ${{V}_{tot}}$ does not need to go through the calculation of the max-plus algorithm and updates the network parameters directly with the TD-error:
\begin{equation}
   {{\mathcal{L}}_{V}}\left( \bm{\tau} ,A,r;\phi  \right):=E\left[ \frac{1}{T}\sum\limits_{t=0}^{T}{{{\left( {{r}_{t}}+\gamma {{V}_{tot}}({\bm{\tau }_{t+1}},{{A}_{t+1}};\bar{\phi })-{{V}_{tot}}({\bm{\tau }_{t}},{{A}_{t}}) \right)}^{2}}} \right]
\end{equation}
where $\bm{\tau} =\{{\bm{\tau }_{t}}\}_{t=1}^{T}$ and ${{r}_{t}}$ is the reward for performing action ${\bm{u}_{t}}$ transitions to ${\bm{\tau }_{t+1}}$ in the observation history ${\bm{\tau }_{t}}$, $\bar{\phi}$ is the parameter copied periodically from $\phi$.

\section{Supplementary Proof}
\subsection{Derivation of Equation (6)}
To facilitate the derivation, we consider all factor nodes represented by ${{{A}_{t}}^{\prime}}$ as a single entity. Then, the action selection process aims to solve the following problem:
\begin{equation}
\begin{aligned}
  & \bm{u}_{t}^{*}=\underset{{\bm{u}_{t}}}{\mathop{\arg \max }}\,{{Q}_{tot}}({\bm{\tau }_{t}},{\bm{u}_{t}},{{A}_{t}};\theta ,\psi ) \\ 
 & \quad =\underset{(u_{t}^{1},\ldots ,u_{t}^{n})}{\mathop{\arg \max }}\,\left( \sum\limits_{j\in J}{{{Q}_{j}}(\bm{u}_{t}^{j}\mid\bm{h}_{t}^{j};{{\theta }_{D(j)}})}+\sum\limits_{i=1}^{n}{{{Q}_{i}}(u_{t}^{i}\mid{h}_{t}^{i};{{\theta }_{D(1)}})} \right) \\ 
 & \quad =\underset{(u_{t}^{1},\ldots ,u_{t}^{n})}{\mathop{\arg \max }}\,\left( \sum\limits_{j\in {J}'}{{{Q}_{j}}(\bm{u}_{t}^{j}\mid\bm{h}_{t}^{j};{{\theta }_{D(j)}})} \right) \\ 
 & \quad =\underset{(u_{t}^{1},\ldots ,u_{t}^{n})}{\mathop{\arg \max }}\,\left( \sum\limits_{j\in {J}'}{{{Q}_{j}}({{\{{{v}_{i}}\}}_{\{i,j\}\in {\mathcal{E}}'}})} \right) \\ 
\end{aligned}
\end{equation}
where ${\mathcal{J}}'=\mathcal{J}\bigcup \{m+i\}_{i=1}^{n}$,${\mathcal{E}}'=\mathcal{E}\bigcup \{\{i,j\}\mid1\le i\le n,j=j+1(j=1\sim n)\}$.

\subsection{Derivation of Equation (13)}
The derivation of Eq. (13) in Section IV-B of the main text is as follows:

Since the graph policy network parameters are $\varphi$, we first derive the derivatives of $J(\theta ,\psi ,\varphi )$ with respect to $\varphi$. Note that the subsequent gradients are all found for $\varphi$, and the graph policies are determined by $\varphi$. For the sake of brevity of the derivation, we will ${{\nabla }_{\varphi }},{{\pi }_{\theta ,\psi }},{{v}_{{{\Pi }_{\theta ,\psi ,\varphi }}}},{{Q}_{{{\Pi }_{\theta ,\psi ,\varphi }}}}$ are abbreviated as $\nabla ,\pi ,{{v}_{\Pi }},{{Q}_{\Pi }}$ respectively.
\begin{footnotesize}
\begin{equation}
\begin{aligned}
  & {{\nabla }_{\varphi }}J(\theta ,\psi ,\varphi )={{\nabla }_{\varphi }}{{v}_{{{\Pi }_{\theta ,\psi ,\varphi }}}}\left( {{s}_{0}} \right)={{\nabla }_{\varphi }}\left[ \sum\limits_{{\bm{u}_{0}},{{A}_{0}}}{{{\Pi }_{\theta ,\psi ,\varphi }}\left( {\bm{u}_{0}},{{A}_{0}}\mid{{s}_{0}} \right){{Q}_{\Pi }}\left( {{s}_{0}},{\bm{u}_{0}},{{A}_{0}} \right)} \right] \\ 
 & ={{\nabla }_{\varphi }}\left[ \sum\limits_{{\bm{u}_{0}}}{{{\pi }_{\theta ,\psi }}\left( {\bm{u}_{0}}\mid{{s}_{0}},{{A}_{0}} \right)\sum\limits_{{{A}_{0}}}{{{\rho }_{\varphi }}\left( {{A}_{0}}\mid{{s}_{0}} \right){{Q}_{\Pi }}\left( {{s}_{0}},{\bm{u}_{0}},{{A}_{0}} \right)}} \right] \\ 
 & =\sum\limits_{{\bm{u}_{0}}}{\left[ \begin{aligned}
  & \pi \left( {\bm{u}_{0}}\mid{{s}_{0}},{{A}_{0}} \right)\nabla \sum\limits_{{{A}_{0}}}{{{\rho }_{\varphi }}\left( {{A}_{0}}\mid{{s}_{0}} \right){{Q}_{\Pi }}\left( {{s}_{0}},{\bm{u}_{0}},{{A}_{0}} \right)} \\ 
 & +\pi \left( {\bm{u}_{0}}\mid{{s}_{0}},{{A}_{0}} \right)\sum\limits_{{{A}_{0}}}{{{\rho }_{\varphi }}\left( {{A}_{0}}\mid{{s}_{0}} \right)\nabla {{Q}_{\Pi }}\left( {{s}_{0}},{\bm{u}_{0}},{{A}_{0}} \right)} \\ 
\end{aligned} \right]} \\
 & =\sum\limits_{{\bm{u}_{0}}}{\left[ \begin{aligned}
  & \pi \left( {\bm{u}_{0}}\mid{{s}_{0}},{{A}_{0}} \right)\sum\limits_{{{A}_{0}}}{\nabla {{\rho }_{\varphi }}\left( {{A}_{0}}\mid{{s}_{0}} \right){{Q}_{\Pi }}\left( {{s}_{0}},{\bm{u}_{0}},{{A}_{0}} \right)} \\ 
 & +\pi \left( {\bm{u}_{0}}\mid{{s}_{0}},{{A}_{0}} \right)\sum\limits_{{{A}_{0}}}{{{\rho }_{\varphi }}\left( {{A}_{0}}\mid{{s}_{0}} \right)}\nabla \sum\limits_{{{s}_{1}},{{r}_{1}}}{p({{s}_{1}},{{r}_{1}}\mid{{s}_{0}},{\bm{u}_{0}},{{A}_{0}})\left( {{r}_{1}}+\gamma {{v}_{\Pi }}\left( {{s}_{1}} \right) \right)} \\ 
\end{aligned} \right]} \\
 & =\sum\limits_{{\bm{u}_{0}}}{\pi \left( {\bm{u}_{0}}\mid{{s}_{0}},{{A}_{0}} \right)\sum\limits_{{{A}_{0}}}{\nabla {{\rho }_{\varphi }}\left( {{A}_{0}}\mid{{s}_{0}} \right){{Q}_{\Pi }}\left( {{s}_{0}},{\bm{u}_{0}},{{A}_{0}} \right)}} \\
 & \quad+\sum\limits_{{\bm{u}_{0}}}{\pi \left( {\bm{u}_{0}}\mid{{s}_{0}},{{A}_{0}} \right)\sum\limits_{{{A}_{0}}}{{{\rho }_{\varphi }}\left( {{A}_{0}}\mid{{s}_{0}} \right)\sum\limits_{{{s}_{1}}}{p({{s}_{1}}\mid{{s}_{0}},{\bm{u}_{0}},{{A}_{0}})\cdot \gamma \nabla {{v}_{\Pi }}\left( {{s}_{1}} \right)}}} \\ 
 & =\sum\limits_{{\bm{u}_{0}}}{\pi \left( {\bm{u}_{0}}\mid{{s}_{0}},{{A}_{0}} \right)\sum\limits_{{{A}_{0}}}{\nabla {{\rho }_{\varphi }}\left( {{A}_{0}}\mid{{s}_{0}} \right){{Q}_{\Pi }}\left( {{s}_{0}},{\bm{u}_{0}},{{A}_{0}} \right)}} \\ 
 & \quad +\sum\limits_{{\bm{u}_{0}}}{\pi \left( {\bm{u}_{0}}\mid{{s}_{0}},{{A}_{0}} \right)}\sum\limits_{{{A}_{0}}}{{{\rho }_{\varphi }}\left( {{A}_{0}}\mid{{s}_{0}} \right)}\sum\limits_{{{s}_{1}}}{p({{s}_{1}}\mid{{s}_{0}},{\bm{u}_{0}})\cdot \gamma }\sum\limits_{{\bm{u}_{1}}}{\pi \left( {\bm{u}_{1}}\mid{{s}_{1}},{{A}_{1}} \right)} \\ 
 & \quad \ \ \,\sum\limits_{{{A}_{1}}}{\nabla {{\rho }_{\varphi }}\left( {{A}_{1}}\mid{{s}_{1}} \right){{Q}_{\Pi }}\left( {{s}_{1}},{\bm{u}_{1}},{{A}_{1}} \right)} \\
 & \quad \quad +\sum\limits_{{\bm{u}_{0}}}{\pi \left( {\bm{u}_{0}}\mid{{s}_{0}},{{A}_{0}} \right)}\sum\limits_{{{A}_{0}}}{{{\rho }_{\varphi }}\left( {{A}_{0}}\mid{{s}_{0}} \right)}\sum\limits_{{{s}_{1}}}{p({{s}_{1}}\mid{{s}_{0}},{\bm{u}_{0}})\cdot \gamma }\sum\limits_{{\bm{u}_{1}}}{\pi \left( {\bm{u}_{1}}\mid{{s}_{1}},{{A}_{1}} \right)} \\ 
  & \quad \quad \ \ \,\sum\limits_{{{A}_{1}}}{\nabla {{\rho }_{\varphi }}\left( {{A}_{1}}\mid{{s}_{1}} \right)\sum\limits_{{{s}_{2}}}{p({{s}_{2}}\mid{{s}_{1}},{\bm{u}_{1}},{{A}_{1}})\cdot \gamma \nabla {{v}_{\Pi }}\left( {{s}_{1}} \right)}} \\ 
 & =\sum\limits_{{\bm{u}_{0}}}{\pi \left( {\bm{u}_{0}}\mid{{s}_{0}},{{A}_{0}} \right)\sum\limits_{{{A}_{0}}}{\nabla {{\rho }_{\varphi }}\left( {{A}_{0}}\mid{{s}_{0}} \right){{Q}_{\Pi }}\left( {{s}_{0}},{\bm{u}_{0}},{{A}_{0}} \right)}} \\ 
 & \quad +\sum\limits_{{\bm{u}_{0}}}{\pi \left( {\bm{u}_{0}}\mid{{s}_{0}},{{A}_{0}} \right)}\sum\limits_{{{A}_{0}}}{{{\rho }_{\varphi }}\left( {{A}_{0}}\mid{{s}_{0}} \right)}\sum\limits_{{{s}_{1}}}{p({{s}_{1}}\mid{{s}_{0}},{\bm{u}_{0}})\cdot \gamma }\sum\limits_{{\bm{u}_{1}}}{\pi \left( {\bm{u}_{1}}\mid{{s}_{1}},{{A}_{1}} \right)} \\ 
 & \quad \ \ \,\sum\limits_{{{A}_{1}}}{\nabla {{\rho }_{\varphi }}\left( {{A}_{1}}\mid{{s}_{1}} \right){{Q}_{\Pi }}\left( {{s}_{1}},{\bm{u}_{1}},{{A}_{1}} \right)}+\cdots  \\
\end{aligned}
\end{equation}
\end{footnotesize}

\begin{footnotesize}
\begin{equation}
\begin{aligned} 
 & =\sum\limits_{{{s}_{0}}}{\Pr \left( {{s}_{0}}\to {{s}_{0}},0,\Pi  \right)}\sum\limits_{{\bm{u}_{0}}}{\pi \left( {\bm{u}_{0}}\mid{{s}_{0}},{{A}_{0}} \right)\sum\limits_{{{A}_{0}}}{\nabla {{\rho }_{\varphi }}\left( {{A}_{0}}\mid{{s}_{0}} \right){{\gamma }^{0}}{{Q}_{\Pi }}\left( {{s}_{0}},{\bm{u}_{0}},{{A}_{0}} \right)}} \\ 
 & \quad+\sum\limits_{{{s}_{1}}}{\Pr \left( {{s}_{0}}\to {{s}_{1}},1,\Pi  \right)\sum\limits_{{\bm{u}_{1}}}{\pi \left( {\bm{u}_{1}}\mid{{s}_{1}},{{A}_{1}} \right)\sum\limits_{{{A}_{1}}}{\nabla {{\rho }_{\varphi }}\left( {{A}_{1}}\mid{{s}_{1}} \right){{\gamma }^{1}}{{Q}_{\Pi }}\left( {{s}_{1}},{\bm{u}_{1}},{{A}_{1}} \right)}}}+\cdots  \\ 
 & =\sum\limits_{{{s}_{0}}}{\Pr \left( {{s}_{0}}\to {{s}_{0}},0,\Pi  \right)}\sum\limits_{{\bm{u}_{0}}}{\pi \left( {\bm{u}_{0}}\mid{{s}_{0}},{{A}_{0}} \right)\sum\limits_{{{A}_{0}}}{{{\rho }_{\varphi }}\left( {{A}_{0}}\mid{{s}_{0}} \right)\left[ {{\gamma }^{0}}{{Q}_{\Pi }}\left( {{s}_{0}},{\bm{u}_{0}},{{A}_{0}} \right)\nabla \ln {{\rho }_{\varphi }}\left( {{A}_{0}}\mid{{s}_{0}} \right) \right]}} \\ 
 & \quad+\sum\limits_{{{s}_{1}}}{\Pr \left( {{s}_{0}}\to {{s}_{1}},1,\Pi  \right)\sum\limits_{{\bm{u}_{1}}}{\pi \left( {\bm{u}_{1}}\mid{{s}_{1}},{{A}_{1}} \right)\sum\limits_{{{A}_{1}}}{{{\rho }_{\varphi }}\left( {{A}_{1}}\mid{{s}_{1}} \right)\left[ {{\gamma }^{1}}{{Q}_{\Pi }}\left( {{s}_{1}},{\bm{u}_{1}},{{A}_{1}} \right)\nabla \ln {{\rho }_{\varphi }}\left( {{A}_{1}}\mid{{s}_{1}} \right) \right]}}} \\
 & \quad+\cdots  \\ 
 & =\sum\limits_{t=0}^{\infty }{\sum\limits_{{{s}_{t}}}{\Pr \left( {{s}_{0}}\to {{s}_{t}},t,\Pi  \right)\sum\limits_{{\bm{u}_{t}}}{\pi \left( {\bm{u}_{t}}\mid{{s}_{t}},{{A}_{t}} \right)\sum\limits_{{{A}_{t}}}{{{\rho }_{\varphi }}\left( {{A}_{t}}\mid{{s}_{t}} \right)\left[ {{\gamma }^{t}}{{Q}_{\Pi }}\left( {{s}_{t}},{\bm{u}_{t}},{{A}_{t}} \right)\nabla \ln {{\rho }_{\varphi }}\left( {{A}_{t}}\mid{{s}_{t}} \right) \right]}}}} \\ 
\end{aligned}
\end{equation}
\end{footnotesize}
where $\gamma $ is the discount factor, $\Pr \left( {{s}_{0}}\to {{s}_{0}},0,\Pi \right)=1$, 
$\Pr \left( {{s}_{0}}\to {{s}_{1}},1,\Pi  \right)=\sum\nolimits_{{\bm{u}_{0}}}{\Pi \left( {\bm{u}_{0}}\mid{{s}_{0}} \right)p\left( {{s}_{1}}\mid{{s}_{0}},{\bm{u}_{0}},{{A}_{0}} \right)}=\sum\nolimits_{{{a}_{0}}}{\pi \left( {\bm{u}_{0}}\mid{{s}_{0}} \right)\sum\nolimits_{{{A}_{0}}}{{{\rho }_{\varphi }}\left( {{A}_{0}}\mid{{s}_{0}} \right)p\left( {{s}_{1}}\mid{{s}_{0}},{\bm{u}_{0}},{{A}_{0}} \right)}}$.

Then, we sum over t moments in $\nabla J$:
\begin{equation}
\begin{aligned}
 & {{\nabla }_{\varphi }}J(\theta ,\psi ,\varphi ) \\ 
 & =\sum\limits_{t=0}^{\infty }{\sum\limits_{{{s}_{t}}}{\Pr \left( {{s}_{0}}\to {{s}_{t}},t,\Pi  \right)\sum\limits_{{\bm{u}_{t}}}{\pi \left( {\bm{u}_{t}}\mid{{s}_{t}},{{A}_{t}} \right)}}} \\ 
 & \quad \sum\limits_{{{A}_{t}}}{{{\rho }_{\varphi }}\left( {{A}_{t}}\mid{{s}_{t}} \right)\left[ {{\gamma }^{t}}{{Q}_{\Pi }}\left( {{s}_{t}},{\bm{u}_{t}},{{A}_{t}} \right)\nabla \ln {{\rho }_{\varphi }}\left( {{A}_{t}}\mid{{s}_{t}} \right) \right]} \\
 & =\sum\limits_{t=0}^{\infty }{\sum\limits_{{{s}_{t}}}{{{\gamma }^{t}}\Pr \left( {{s}_{0}}\to {{s}_{t}},t,\Pi  \right)\sum\limits_{{\bm{u}_{t}}}{\pi \left( {\bm{u}_{t}}\mid{{s}_{t}},{{A}_{t}} \right)}}} \\ 
 & \quad \sum\limits_{{{A}_{t}}}{{{\rho }_{\varphi }}\left( {{A}_{t}}\mid{{s}_{t}} \right)\left[ {{Q}_{\Pi }}\left( {{s}_{t}},{\bm{u}_{t}},{{A}_{t}} \right)\nabla \ln {{\rho }_{\varphi }}\left( {{A}_{t}}\mid{{s}_{t}} \right) \right]} \\
 & =\sum\limits_{x\in S}{\sum\limits_{t=0}^{\infty }{{{\gamma }^{t}}\Pr \left( {{s}_{0}}\to x,t,\Pi  \right)\sum\limits_{\bm{u}}{\pi \left( \bm{u}\mid x,A \right)}}} \\ 
 & \quad \sum\limits_{A}{{{\rho }_{\varphi }}\left( A\mid x \right)\left[ {{Q}_{\Pi }}\left( x,\bm{u},A \right)\nabla \ln {{\rho }_{\varphi }}\left( A\mid x \right) \right]} \\
 & =\sum\limits_{x\in S}{{{d}^{\Pi }}\left( x \right)\sum\limits_{\bm{u}}{\pi \left( \bm{u}\mid x,A \right)\sum\limits_{A}{{{\rho }_{\varphi }}\left( A\mid x \right)\left[ {{Q}_{\Pi }}\left( x,\bm{u},A \right)\nabla \ln {{\rho }_{\varphi }}\left( A\mid x \right) \right]}}} \\ 
\end{aligned}
\end{equation}
where ${{d}^{\Pi }}\left( x \right)=\sum\limits_{t=0}^{\infty }{{{\gamma }^{t}}\Pr \left( {{s}_{0}}\to x,t,\Pi \right)}$ is the discounted state distribution.

Finally, write $\nabla J$ in the expected form:
\begin{equation}
\begin{aligned}
  & {{\nabla }_{\varphi }}J(\theta ,\psi ,\varphi ) \\ 
 & =\sum\limits_{x\in S}{{{d}^{\Pi }}\left( x \right)\sum\limits_{\bm{u}}{\pi \left( \bm{u}\mid x,A \right)\sum\limits_{A}{{{\rho }_{\varphi }}\left( A\mid x \right)\left[ {{Q}_{\Pi }}\left( x,\bm{u},A \right)\nabla \ln {{\rho }_{\varphi }}\left( A\mid x \right) \right]}}} \\ 
 & =\frac{1}{1-\gamma }\sum\limits_{x\in S}{\left( 1-\gamma  \right){{d}^{\Pi }}\left( x \right)\sum\limits_{\bm{u}}{\pi \left( \bm{u}\mid x,A \right)\sum\limits_{A}{{{\rho }_{\varphi }}\left( A\mid x \right)\left[ {{Q}_{\Pi }}\left( x,\bm{u},A \right)\nabla \ln {{\rho }_{\varphi }}\left( A\mid x \right) \right]}}} \\ 
 & =\frac{1}{1-\gamma }\sum\limits_{x\in S}{{{D}^{\Pi }}\left( x \right)\sum\limits_{\bm{u}}{\pi \left( \bm{u}\mid x,A \right)\sum\limits_{A}{{{\rho }_{\varphi }}\left( A\mid x \right)\left[ {{Q}_{\Pi }}\left( x,\bm{u},A \right)\nabla \ln {{\rho }_{\varphi }}\left( A\mid x \right) \right]}}} \\ 
 & =\frac{1}{1-\gamma }\underset{\begin{smallmatrix} 
 s\sim {{D}^{\Pi }} \\ 
 A\sim {{\rho }_{\varphi }} 
 \\ 
 \bm{u}\sim \pi  
\end{smallmatrix}}{\mathop{E}}\,\left[ {{Q}_{\Pi }}\left( s,\bm{u},A \right)\nabla \ln {{\rho }_{\varphi }}\left( A\mid s \right) \right] \\ 
 & \propto \underset{\begin{smallmatrix} 
 s\sim {{D}^{\Pi }} \\ 
 A\sim {{\rho }_{\varphi }} 
 \\ 
 \bm{u}\sim \pi  
\end{smallmatrix}}{\mathop{E}}\,\left[ {{Q}_{\Pi }}\left( s,\bm{u},A \right)\nabla \ln {{\rho }_{\varphi }}\left( A\mid s \right) \right] \\ 
\end{aligned}
\end{equation}
where ${{D}^{\Pi }}\left( x \right)$ is the standard distribution and $\sum\limits_{x\in S}{{{D}^{\Pi }}\left( x \right)}=1$.
\begin{equation}
\begin{aligned}
  & \sum\limits_{x\in S}{{{D}^{\Pi }}\left( x \right)}=\left( 1-\gamma  \right)\sum\limits_{x\in S}{{{d}^{\Pi }}\left( x \right)} \\ 
 & =\left( 1-\gamma  \right)\sum\limits_{x\in S}{\sum\limits_{t=0}^{\infty }{{{\gamma }^{t}}\Pr \left( {{s}_{0}}\to x,t,\Pi  \right)}} \\ 
 & =\left( 1-\gamma  \right)\sum\limits_{t=0}^{\infty }{{{\gamma }^{t}}}\sum\limits_{x\in S}{\Pr \left( {{s}_{0}}\to x,t,\Pi  \right)} \\ 
 & =\left( 1-\gamma  \right)\sum\limits_{t=0}^{\infty }{{{\gamma }^{t}}} \\ 
 & =1 \\ 
\end{aligned}
\end{equation}

We usually ignore the scale factor so that the final form of $\nabla J$ is:
\begin{equation}
{{\nabla }_{\varphi }}J(\theta ,\psi ,\varphi )=\underset{\left( s,\bm{u},A \right)\sim \mathcal{T}}{\mathop{E}}\,\left[ {{Q}_{\Pi }}\left( s,\bm{u},A \right)\nabla \ln {{\rho }_{\varphi }}\left( A\mid s \right) \right]
\end{equation}
where $\mathcal{T}=\left( {{s}_{0}},\{o_{0}^{i}\}_{i=1}^{n},{{A}_{0}},u_{o}^{i},{{r}_{0}},\ldots ,{{s}_{T}},\{o_{T}^{i}\}_{i=1}^{n} \right)$.

\subsection{Proof of Proposition 1}
Because $X\sim {{P}_{{{D}_{\max }}}}({{D}_{\max }}:{{p}_{1}},{{p}_ {2}},\ldots ,{{p}_{N}})$, there are ${{m}_{1}}+{{m}_{2}}+\cdots +{{m}_ {N}}={{D}_{\max }}$. When $\max {{m}_{i}}=1$, ${{Q}_{j}}$ connects exactly ${{D}_{\max }}$ agents, and its order is ${{D}_{\max }}$; when $\max {{m}_{i}}>1$, there must be at least one agent i and ${{Q}_{j}} $ is connected more than once, then the number of connected agents ${{Q}_{j}}$ is less than ${{D}_{\max }}$, and its order is less than ${{D}_{\max }}$. So for each ${{Q}_{j}}$, its maximum order is ${{D}_{\max }}$, that is, ${{D}_{\max }}$ is The maximum order in the algorithm.

\subsection{Proof of Proposition 2}
Through Proposition 1, it can be seen that the "sub-policy" corresponding to ${{Q}_{j}}$ needs to be able to generate a collaborative relationship of $D\le {{D}_{\max }}$ agents. When the order ${{Q}_{j}}$ generated by "sub-policy" is ${{D}_{j}}$, ${{Q}_{j}}$ will be the same as ${{ D}_{j}}$ agents are connected, and the set of ${{D}_{j}}$ agents is defined as $G\in {{H}_{{{D}_{j} }}}$, ${{H}_{{{D}_{j}}}}=\{\{{{i}_{1}},\ldots ,{{i}_{d}} ,\ldots ,{{i}_{{{D}_{j}}}}\}\mid{{i}_{d}}\in \{1,2,\ldots ,N\}\}$. Then the probability of the "sub-policy" corresponding to ${{Q}_{j}}$ is the probability of ${{Q}_{j}}$ being connected to all agents in $G$:
\begin{equation}\label{eq:1}
\tilde{P}\{G\}=\sum\limits_{\{{{m}_{1}},\ldots ,{{m}_{N}}\}\in {{H}_{m}}}{P\{{{X}_{1}}={{m}_{1}},{{X}_{2}}={{m}_{2}},\ldots ,{{X}_{N}}={{m}_{N}}\}}
\end{equation}
where ${{H}_{m}}=\{\{{{m}_{1}},\ldots ,{{m}_{N}}\}\mid\sum\nolimits_{{{i}_{d}}\in G}{{{m}_{{{i}_{d}}}}={{D}_{\max }}},\forall {{i}_{d}}\in G,{{m}_{{{i}_{d}}}}\ge 1\}$.

Then, it is necessary to prove that $\tilde{P}\{G\}$ is the probability mass function of the class multinomial distribution ${{\tilde{P}}_{{{D}_{\max }}}}({{D}_{\max }}:{{p}_{1}},{{p}_{2}},\ldots ,{{p}_{N}})$.
\begin{enumerate}
\item[(1)] $\tilde{P}\{G\}\ge 0$

Because $P\{{{X}_{1}}={{m}_{1}},{{X}_{2}}={{m}_{2}},\ldots ,{ {X}_{N}}={{m}_{N}}\}$ is the multinomial distribution ${{P}_{{{D}_{\max }}}}({{D} _{\max }}:{{p}_{1}},{{p}_{2}},\ldots ,{{p}_{N}})$ probability mass function, and $P \{{{X}_{1}}={{m}_{1}},{{X}_{2}}={{m}_{2}},\ldots ,{{X}_ {N}}={{m}_{N}}\}\ge 0$, then according to Eq.(\ref{eq:1}), $\tilde{P}\{G\}\ge 0$.

\item[(2)] $\sum\nolimits_{{{D}_{j}}=1}^{{{D}_{\max }}}{\sum\nolimits_{G\in {{H}_{{{D}_{j}}}}}{\tilde{P}\{G\}}}=1$

Because $P\{{{X}_{1}}={{m}_{1}},{{X}_{2}}={{m}_{2}},\ldots ,{ {X}_{N}}={{m}_{N}}\}$ is the probability mass function of multinomial distribution ${{P}_{{{D}_{\max }}}}({{D} _{\max }}:{{p}_{1}},{{p}_{2}},\ldots ,{{p}_{N}})$, and we have:
\begin{equation}\label{eq:2}
\sum\limits_{{{m}_{1}}+{{m}_{2}}+\cdots +{{m}_{N}}={{D}_{\max }}}{P\{{{X}_{1}}={{m}_{1}},{{X}_{2}}={{m}_{2}},\ldots ,{{X}_{N}}={{m}_{N}}\}}=1
\end{equation}
where there are a total of $C_{{{D}_{\max }}+N-1}^{N-1}$ items in Eq.(\ref{eq:2}).

\begin{equation}\label{eq:3}
\begin{aligned}
& \sum\limits_{{{D}_{j}}=1}^{{{D}_{\max }}}{\sum\limits_{G\in {{H}_{{{D}_{j}}}}}{\tilde{P}\{G\}}} \\
& =\sum\limits_{{{D}_{j}}=1}^{{{D}_{\max }}}{\sum\limits_{G\in {{H}_{{{D}_{j}}}}}{\sum\limits_{\{{{m}_{1}},\ldots ,{{m}_{N}}\}\in {{H}_{m}}}{P\{{{X}_{1}}={{m}_{1}},{{X}_{2}}={{m}_{2}},\ldots ,{{X}_{N}}={{m}_{N}}\}}}} \\
\end{aligned}
\end{equation}

Next, we will calculate how many terms there are in equation Eq.(\ref{eq:3}):
\begin{equation}\label{eq:4}
\begin{aligned}
  & \sum\limits_{{{D}_{j}}=1}^{{{D}_{\max }}}{\sum\limits_{G\in {{H}_{{{D}_{j}}}}}{\sum\limits_{\{{{m}_{1}},\ldots ,{{m}_{N}}\}\in {{H}_{m}}}{1}}}=\sum\limits_{{{D}_{j}}=1}^{{{D}_{\max }}}{C_{N}^{{{D}_{j}}}C_{{{D}_{\max }}-1}^{{{D}_{j}}-1}}=\sum\limits_{{{D}_{j}}=1}^{{{D}_{\max }}}{C_{N}^{{{D}_{j}}}C_{{{D}_{\max }}-1}^{{{D}_{\max }}-{{D}_{j}}}} \\ 
 & =\sum\limits_{{{D}_{j}}=0}^{{{D}_{\max }}}{C_{N}^{{{D}_{j}}}C_{{{D}_{\max }}-1}^{{{D}_{\max }}-{{D}_{j}}}}=C_{N+{{D}_{\max }}-1}^{{{D}_{\max }}}=C_{N+{{D}_{\max }}-1}^{N-1} \\ 
\end{aligned}
\end{equation}

According to the definition of ${{H}_{{{D}_{j}}}}$, for different ${{D}_{j}}$, ${{H}_{{{D}_ {j}}}}$ are disjoint with each other; and the composition of ${{H}_{m}}$ depends on $G$ and corresponds to $G$ one-to-one, while $G\in {{ H}_{{{D}_{j}}}}$, so there is no intersection between ${{H}_{m}}$. Then it means that there are no duplicates in the summation of $P\{{{X}_{1}}={{m}_{1}},\ldots ,{{X}_{N}}={{m}_{N}}\}$. Through Eq. (\ref{eq:4}), we know that there are $C_{N+{{D}_{\max }}-1}^{N-1}$ terms in Eq. (\ref{eq:3}), and each term is The probability mass function in the multinomial distribution ${{P}_{{{D}_{\max }}}}({{D}_{\max }}:{{p}_{1}},{{p}_{2}},\ldots ,{{p}_{N}})$, and does not repeat, then you can get $\sum\nolimits_{{{D}_{j}}=1}^{{{D}_{\max }}}{ \sum\nolimits_{G\in {{H}_{{{D}_{j}}}}}{\tilde{P}\{G\}}}=1$.

\subsection{Proof of Proposition 3}
The global function ${{Q}_{tot}}$ can be decomposed into the sum of local value functions ${{Q}_{j}}$:
\begin{equation}
\begin{aligned}
  & \frac{{{\rho }_{\varphi }}\left( {{A}_{t}}\mid{\bm{\tau }_{t}} \right)}{{{\rho }_{{{\varphi }_{old}}}}\left( {{A}_{t}}\mid{\bm{\tau }_{t}} \right)}{{Q}_{tot}}\left( {\bm{\tau }_{t}},{\bm{u}_{t}},{{A}_{t}} \right) \\ 
 & =\frac{{{\rho }_{\varphi }}\left( {{A}_{t}}\mid{\bm{\tau }_{t}} \right)}{{{\rho }_{{{\varphi }_{old}}}}\left( {{A}_{t}}\mid{\bm{\tau }_{t}} \right)}\sum\limits_{j\in \mathcal{J}}{{{Q}_{j}}\left( \bm{\tau}_{t}^{j},\bm{u}_{t}^{j},{{A}_{t}} \right)} \\ 
 & =\sum\limits_{j\in \mathcal{J}}{\frac{{{\rho }_{\varphi }}\left( {{A}_{t}}\mid{{\tau }_{t}} \right)}{{{\rho }_{{{\varphi }_{old}}}}\left( {{A}_{t}}\mid{{\tau }_{t}} \right)}{{Q}_{j}}\left( \bm{\tau}_{t}^{j},\bm{u}_{t}^{j},{{A}_{t}} \right)} \\ 
\end{aligned}
\end{equation}

Because the value of each local value function ${{Q}_{j}}$ is only related to the agent it is connected to, and has nothing to do with the connection relationship of other local value functions, so in the graph policy evaluated by ${{Q}_{j}}$, items unrelated to $j$ can be ignored, that is, ${{Q}_{j}}$ accurately evaluates the "sub-policy" of ${{Q}_{j}}$ connected to the agent. . At the same time, the random variable corresponding to the "sub-policy" is $\rho ({{Q}_{j}})\sim {{\tilde{P}}_{{{D}_{\max }}}}( {{D}_{\max }}:{{p}_{1}},{{p}_{2}},\ldots ,{{p}_{N}})$, then by Definition 1 available:
\begin{equation}
\begin{aligned}
  & \sum\limits_{j\in \mathcal{J}}{\frac{{{\rho }_{\varphi }}\left( {{A}_{t}}\mid{\bm{\tau }_{t}} \right)}{{{\rho }_{{{\varphi }_{old}}}}\left( {{A}_{t}}\mid{\bm{\tau }_{t}} \right)}{{Q}_{j}}\left( \bm{\tau}_{t}^{j},\bm{u}_{t}^{j},{{A}_{t}} \right)} \\ 
 & =\sum\limits_{j\in \mathcal{J}}{\frac{\sum\limits_{\{{{m}_{j,1}},\ldots ,{{m}_{j,N}}\}\in {{H}_{m}}}{{{P}_{\varphi }}\{{{X}_{j,1}}={{m}_{j,1}},\ldots ,{{X}_{j,N}}={{m}_{j,N}}\}}}{\sum\limits_{\{{{m}_{j,1}},\ldots ,{{m}_{j,N}}\}\in {{H}_{m}}}{{{P}_{{{\varphi }_{old}}}}\{{{X}_{j,1}}={{m}_{j,1}},\ldots ,{{X}_{j,N}}={{m}_{j,N}}\}}}{{Q}_{j}}\left(\bm{\tau}_{t}^{j},\bm{u}_{t}^{j},{{A}_{t}} \right)} \\ 
 & =\sum\limits_{j\in \mathcal{J}}{\frac{{{\left( \sum\limits_{\{{{m}_{j,1}},\ldots ,{{m}_{j,N}}\}\in {{H}_{m}}}{\frac{C!}{{{m}_{j,1}}!\cdots {{m}_{j,N}}!}p_{j,1}^{{{m}_{j,1}}}\cdots p_{j,N}^{{{m}_{j,N}}}} \right)}_{\varphi }}}{{{\left( \sum\limits_{\{{{m}_{j,1}},\ldots ,{{m}_{j,N}}\}\in {{H}_{m}}}{\frac{C!}{{{m}_{j,1}}!\cdots {{m}_{j,N}}!}p_{j,1}^{{{m}_{j,1}}}\cdots p_{j,N}^{{{m}_{j,N}}}} \right)}_{{{\varphi }_{old}}}}}{{Q}_{j}}\left( \bm{\tau}_{t}^{j},\bm{u}_{t}^{j},{{A}_{t}} \right)} \\ 
\end{aligned}
\end{equation}

Since the graph policy uses importance sampling, there is ${{A}_{t}}\sim {{\mathcal{T}}_{old}}$, which means that sample the same graph structure under $\varphi $ and ${{\varphi }_{old}}$, that is, ${{\left( {{m}_{j,1}}!\cdots { {m}_{j,N}}! \right)}_{\varphi }}={{\left( {{m}_{j,1}}!\cdots {{m}_{j,N}}! \right)}_{{{\varphi }_{old}}}}$, then we can get:
\begin{equation}
\begin{aligned}
  & \sum\limits_{j\in \mathcal{J}}{\frac{\sum\limits_{\{{{m}_{j,1}},\ldots ,{{m}_{j,N}}\}\in {{H}_{m}}}{{{P}_{\varphi }}\{{{X}_{j,1}}={{m}_{j,1}},\ldots ,{{X}_{j,N}}={{m}_{j,N}}\}}}{\sum\limits_{\{{{m}_{j,1}},\ldots ,{{m}_{j,N}}\}\in {{H}_{m}}}{{{P}_{{{\varphi }_{old}}}}\{{{X}_{j,1}}={{m}_{j,1}},\ldots ,{{X}_{j,N}}={{m}_{j,N}}\}}}{{Q}_{j}}\left( \bm{\tau}_{t}^{j},\bm{u}_{t}^{j},{{A}_{t}} \right)} \\ 
 & =\sum\limits_{j\in \mathcal{J}}{\sum\limits_{\{{{m}_{j,1}},\ldots ,{{m}_{j,N}}\}\in {{H}_{m}}}{\frac{{{\left( p_{j,1}^{{{m}_{j,1}}}\cdots p_{j,N}^{{{m}_{j,N}}} \right)}_{\varphi }}}{{{\left( p_{j,1}^{{{m}_{j,1}}}\cdots p_{j,N}^{{{m}_{j,N}}} \right)}_{{{\varphi }_{old}}}}}{{Q}_{j}}\left( \bm{\tau}_{t}^{j},\bm{u}_{t}^{j},{{A}_{t}} \right)}} \\ 
 & =\sum\limits_{j\in \mathcal{J}}{\frac{{{P}_{\varphi ,j}}\left( {{m}_{j}} \right)}{{{P}_{{{\varphi }_{old}},j}}\left( {{m}_{j}} \right)}{{Q}_{j}}\left( \bm{\tau}_{t}^{j},\bm{u}_{t}^{j},{{A}_{t}} \right)} \\ 
\end{aligned}
\end{equation}

Finally got:
\begin{equation}
\begin{aligned}
\frac{{{\rho }_{\varphi }}\left( {{A}_{t}}\mid{\bm{\tau }_{t}} \right)}{{{\rho }_{{{\varphi }_{old}}}}\left( {{A}_{t}}\mid{\bm{\tau }_{t}} \right)}{{Q}_{tot}}\left( {\bm{\tau }_{t}},{\bm{u}_{t}},{{A}_{t}} \right)=\sum\limits_{j\in \mathcal{J}}{\frac{{{P}_{\varphi ,j}}\left( {{m}_{j}} \right)}{{{P}_{{{\varphi }_{old}},j}}\left( {{m}_{j}} \right)}{{Q}_{j}}\left( \bm{\tau}_{t}^{j},\bm{u}_{t}^{j},{{A}_{t}} \right)}
\end{aligned}
\end{equation}

\end{enumerate}

\section{Supplementary Experiments}

\subsection{HO-Predator-Prey}

The experimental environment is formulated as a complex 10$\times$10 grid, wherein nine agents are tasked with the capture of six prey entities. Agents are endowed with the capacity to navigate in one of the four cardinal directions, remain stationary, or engage in a capture action. Actions deemed unfeasible, such as moving to an already occupied space or attempting a capture action outside the eight adjacent squares surrounding a prey, are systematically restricted. Prey exhibit random movement within the four cardinal directions and will remain stationary should all adjacent positions be occupied. Upon successful capture, prey are removed from the grid, permitting agents to pursue subsequent targets. Each agent's perceptual field encompasses a 5$\times$5 subgrid centered upon its location. Agents and prey that have been removed from play are rendered invisible within the simulation. An episode reaches its conclusion either when all prey have been captured or after the elapse of 200 time steps. 

% \subsubsection{Hyper-parameters}
\textit{Hyper-parameters}: The implementation of all algorithms adheres to the MARLBenchmark framework, with the source code accessible via \url{https://github.com/SICC-Group/DDFG}. 

For all tasks, a discount factor $\gamma = 0.98$ is employed alongside $\epsilon$-greedy exploration strategy. This strategy initiates with $\epsilon = 1$ and linearly decays to $\epsilon = 0.05$ over the first 50,000 time steps. Evaluation occurs every 2,000 steps through the execution of ten greedy test trajectories at $\epsilon = 0$. Each algorithm undergoes three independent trials, with the mean and standard error computed for each trial's outcomes.

A replay buffer, with a capacity for 5,000 entries, is maintained across all algorithms, alongside normalization of rewards contained within. A recurrent neural network (RNN) ${h}_{\psi}$ processes agent observations across all methodologies. The Adam optimization algorithm is uniformly applied. For VDN, QMIX, MADDPG, QTRAN, and QPLEX, a learning rate of $7\times10^{-4}$ is set, while CW-QMIX, OW-QMIX, DCG, SOPCG, and CASEC adopt a learning rate of $1\times10^{-3}$. Within DDFG, the Q-value and V-value function networks utilize a learning rate of $1\times10^{-3}$, whereas the graph policy network employs a learning rate of $1\times10^{-5}$. DCG incorporates a fully connected graph structure $\mathcal{E} := \left\{ \left\{ i,j \right\}\mid1\le i<n,i<j\le n \right\}$. The DDFG constrains the local value function's highest order to 3 and incorporates a distinct replay buffer, with a capacity of 8 entries, for the graph policy (referencing PPO), setting the constant ${{\lambda }_{\mathcal{H}}}$ at 0.01.

\subsection{SMAC}

In our study, we delve into the intricacies of unit micromanagement tasks within the StarCraft II environment, a domain characterized by its strategic complexity and real-time decision-making demands. In these scenarios, enemy units are orchestrated by the game's built-in artificial intelligence, whereas allied units are governed by a reinforcement learning algorithm. The composition of both groups may encompass a variety of soldier types, albeit restricted to a single race per faction. Agents are afforded a discrete set of actionable choices at each timestep, including inaction, directional movement, targeted attacks, and cessation of activity. These actions facilitate agent navigation and engagement on continuous maps. The reinforcement learning framework assigns a global reward to Multi-Agent Reinforcement Learning (MARL) agents, proportional to the cumulative damage inflicted on adversary units. Furthermore, the elimination of enemy units and overall combat victory are incentivized with additional reward bonuses of 10 and 200 points, respectively. A concise overview of the StarCraft Multi-Agent Challenge (SMAC) scenarios investigated in this study is provided in Table \ref{sample-table}, incorporating the inclusion of particularly challenging maps: 1c3s5z (hard), $8m\_vs\_9m$(hard), and MMM2 (superhard).

\begin{table}
  \caption{The StarCraft multi-Agent challenge benchmark}
  \label{sample-table}
  \centering
  \begin{tabular}{lll}
    \toprule
    Map Name     & Ally Units     & Enemy Units \\
    \midrule
    3s5z & 3 Stalkers $\And $ 5 Zealots  & 3 Stalkers $\And $ 5 Zealots     \\
    $5m \_ vs \_ 6m$     & 5 Marines & 6 Marines      \\
    1c3s5z     & 9 Marines & 9 Marines      \\
    $8m \_ vs \_ 9m$     & 8 Marines & 9 Marines      \\
    MMM2     & 1 Medivac, 2 Marauders $\And $ 7 Marines       & 1 Medivac, 2 Marauders $\And $ 8 Marines  \\
    \bottomrule
  \end{tabular}
\end{table}

% \subsubsection{Hyper-parameters}
\textit{Hyper-parameters}: For all conducted SMAC experiments, version 2.4.10 of StarCraft II was utilized. Experimental setups consistently applied a discount factor ($\gamma$) of 0.97 and an $\epsilon$-greedy exploration strategy, which linearly decreased from $\epsilon = 1$ to $\epsilon = 0.05$ across the initial 50,000 timesteps. Evaluations, comprising ten greedy test trajectories, were systematically conducted every 10,000 steps at $\epsilon = 0$. Each algorithm was subjected to three distinct initialization seeds, facilitating the computation of mean performance and standard error metrics.

All implemented algorithms employed a replay buffer capacity of 5,000 entries, with rewards within this buffer normalized for consistency. Observational data from agents were processed through a recurrent neural network (RNN) denoted as ${h}_{\psi}$, with the Adam optimization algorithm uniformly applied across all models. The learning rates for VDN, QMIX, CW-QMIX, OW-QMIX, MADDPG, QTRAN, DCG, SOPCG, and CASEC were set at $1 \times 10^{-3}$, while QPLEX adopted a slightly reduced learning rate of $1 \times 10^{-4}$. Within the Dynamic Deep Factor Graph (DDFG) framework, both the Q-value and V-value function networks (as outlined in section III-A of the main text and Appendix B, respectively) utilized a learning rate of $1 \times 10^{-3}$, with the graph policy network assigned a learning rate of $1 \times 10^{-6}$. The DDFG model imposes a limitation on the highest order of local value function to 2 and incorporates a dedicated replay buffer of eight entries for the graph policy (referencing Proximal Policy Optimization, PPO). Additionally, a constant ${{\lambda}_{\mathcal{H}}}$ value of 0.01 was established.

\section{Code Deployment and Utilization}

\subsection{Setup}
Here we briefly describe the setup of DDFG. More information is given in the README file of the repository.

\begin{enumerate}
\item Install the dependencies.

Here we give an example installation on CUDA == 10.1. For non-GPU and other CUDA version installation, please refer to the PyTorch website.

$\$ \quad$ conda create -n marl python==3.6.1

$\$ \quad$ conda activate marl

$\$ \quad$ pip install torch==1.5.1+cu101 torchvision==0.6.1+cu101 

$\quad$ -f https://download.pytorch.org/whl/torch$\_$stable.html

$\$ \quad$ pip install -r requirements.txt

\item Obtain the source code and install.

$\$ \quad$ git clone https://github.com/SICC-Group/DDFG.git

$\$ \quad$ cd off-policy

$\$ \quad$ pip install -e .

\item Install StarCraftII 4.10

$\$ \quad$ pip install git+https://github.com/oxwhirl/smac.git

$\$ \quad$ echo "export SC2PATH=~/StarCraftII/" > ~/.bashrc

$\#$ Download SMAC Maps, and move it to ~/StarCraftII/Maps/
\end{enumerate}

\subsection{Reproduction}
\begin{enumerate}
\item Train of DDFG

$\$ \quad$  cd offpolicy/scripts

$\$ \quad$  chmod +x ./train$\_$<experiment>$\_$<algorithm>.sh

$\$ \quad$  .train$\_$<experiment>$\_$<algorithm>.sh

$\#$ <experiment> can choose between prey or smac, <algorithm> can choose between rmdfg or other baselines.

$\#$ The trained model and data files can be found under ./results/<experiment$\_$name>/<algorithm$\_$name>/debug/run1/models/.

$\#$ <experiment$\_$name> can choose between Predator$\_$prey or StarCraft2, <algorithm$\_$name> can choose between rddfg$\_$cent$\_$rw or other baselines.

\item Test of DDFG

$\#$ Select the trained model to load and test

$\$ \quad$  .train$\_$<experiment>$\_$<algorithm>.sh 

$ \quad$  --model$\_$dir ./results/<experiment$\_$name>/<algorithm$\_$name>/debug/run1/models/

\item Replot the experimental results of this manuscript

$\$ \quad$ cd off-policy/experiment$\_$eval$\_$data

$\#$ Execute file plot$\_$experiment$\_$pp.py or plot$\_$experiment$\_$smac.py to obtain results. 
    
\end{enumerate}

\bibliographystyle{IEEEtran}
\bibliography{references}